\documentclass[a4paper,12pt]{article}

\newcommand{\longtitle}{
    The regret lower bound for communicating Markov Decision Processes
}
\newcommand{\shorttitle}{
    MDP lower bounds
}

\title{\textsc{\longtitle}}
\date{}
\author{\qquad
	\begin{minipage}{0.45\textwidth}
        \small \textbf{Victor Boone}$^\dagger$\\[-0.1em]
        \footnotesize \texttt{victor.boone@inria.fr}\\[-0.1em]
    \end{minipage}
    \qquad
    \hfill
    \begin{minipage}{0.45\textwidth}
        \small \textbf{Odalric-Ambrym Maillard}$^\ddagger$\\[-0.1em]
        \footnotesize \texttt{odalric.maillard@inria.fr}\\[-0.1em]
    \end{minipage}
    \\
    \footnotesize $^\dagger$Univ.~Grenoble Alpes, Inria, CNRS, Grenoble INP, LIG, 38000 Grenoble, France. 
    \\[-0.5em]
    \footnotesize $^\ddagger$Univ.~Lille, Inria, CNRS, Centrale Lille, UMR 9189 – CRIStAL, F-59000, Lille, France 
}

\usepackage[utf8]{inputenc} 
\usepackage{newcent} 
\usepackage{subfiles} 
\usepackage{comment} 
\usepackage{bbding} 

\usepackage[top=2.5cm, bottom=2cm, left=2cm, right=2cm]{geometry}

\usepackage{fancyhdr}
\usepackage{hyperref}       
\usepackage{url}            
\usepackage{soul}           

\hypersetup
{
	colorlinks=true,
	linkcolor=MidnightBlue,
	filecolor=magenta,
	urlcolor=FireBrick,
	citecolor=MidnightBlue,
}

\usepackage{booktabs}       
\usepackage{multicol}

\usepackage[svgnames,dvipsnames]{xcolor}         
\usepackage{graphicx} 
\usepackage{tikz}
\usetikzlibrary{positioning}
\usetikzlibrary{backgrounds}
\usetikzlibrary{shapes.geometric} 
\graphicspath{{figures/}}

\tikzset{
    state/.style={
        draw,
        circle,
        fill=none,
    },
    action/.style={
        draw,
        circle,
        color=white,
        fill=black,
        inner sep=0pt,
        minimum size=4mm,
        anchor=center,
    },
    reward/.style={
        font={\scriptsize}
    },
    transition/.style={
        ->,
        >=stealth,
    },
}

\colorlet{MyRed}{Crimson!60!DarkRed}
\colorlet{MyBlue}{DodgerBlue!75!black}
\colorlet{MyGreen}{DarkGreen}
\colorlet{MyViolet}{DarkMagenta}

\colorlet{MyLightBlue}{DodgerBlue!20}
\colorlet{MyLightGreen}{MyGreen!20}

\colorlet{PrimalColor}{MyBlue}
\colorlet{PrimalFill}{MyLightBlue}
\colorlet{DualColor}{MyRed}

\colorlet{AlertColor}{MyRed}	
\colorlet{BadColor}{MyRed}	
\colorlet{GoodColor}{MyGreen}	
\colorlet{LinkColor}{MediumBlue}	
\colorlet{MacroColor}{MyViolet}
\colorlet{RevColor}{MediumBlue}	

\usepackage[most]{tcolorbox}

\usepackage[inline,shortlabels]{enumitem}

\newenvironment{enum}{
    \begin{enumerate}[{\upshape (1)}, itemsep=-.33em]
}{
    \end{enumerate}
}

\usepackage{natbib}

\usepackage{mathmacros}
\usepackage{mdpmath}

\usepackage{algorithm,algorithmic}

\usepackage{xargs}
\usepackage[
    colorinlistoftodos,
    prependcaption,
    textsize=scriptsize,
    textwidth=1.25in
]{todonotes}
\newcommandx{\unsure}[2][1=]{\todo[linecolor=red,backgroundcolor=red!25,bordercolor=red,#1]{#2}}
\newcommandx{\change}[2][1=]{\todo[linecolor=blue,backgroundcolor=blue!25,bordercolor=blue,#1]{#2}}
\newcommandx{\info}[2][1=]{\todo[linecolor=OliveGreen,backgroundcolor=OliveGreen!20,bordercolor=OliveGreen,#1]{#2}}
\newcommandx{\improvement}[2][1=]{\todo[linecolor=Plum,backgroundcolor=Plum!25,bordercolor=Plum,#1]{#2}}
\newcommandx{\defquery}[2][1=]{\todo[linecolor=Goldenrod,backgroundcolor=Goldenrod!25,bordercolor=Goldenrod,#1]{#2}}
\newcommand{\oam}[1]{\todo[inline,color=orange!40]{\normalfont\normalsize{\textbf{OM:}~}#1}}

\usepackage[showdeletions]{color-edits}	
\newcommand{\draft}[1]{#1}	

\newcommand{\strong}[1]{\textbf{#1}} 
\newcommand{\hideme}[1]{} 

\newtheorem{property}{Property}

\allowdisplaybreaks 
\begin{document}
	
	\maketitle
	
	\begin{abstract}%
        This paper is devoted to the extension of the regret lower bound beyond ergodic Markov decision processes (MDPs) in the problem dependent setting. 
        While the regret lower bound for ergodic MDPs is well-known and reached by tractable algorithms, we prove that the regret lower bound becomes significatively more complex in communicating MDPs. 
        Our lower bound revisits the necessary explorative behavior of consistent learning agents and further explains that \emph{all} optimal regions of the environment must be overvisited compared to sub-optimal ones, a phenomenon that we refer to as co-exploration.
        In tandem, we show that these two explorative and co-explorative behaviors are intertwined with navigation constraints obtained by scrutinizing the navigation structure at logarithmic scale. 
        The resulting lower bound is expressed as the solution of an optimization problem that, in many standard classes of MDPs, can be specialized to recover existing results.
        From a computational perspective, it is provably $\Sigma_2^\textrm{P}$-hard in general and as a matter of fact, even testing the membership to the feasible region is coNP-hard. 
        We further provide an algorithm to approximate the lower bound in a constructive way. 
        \\[1em]
    \end{abstract}

    \section{Introduction}
    This paper considers the problem of regret minimization in discrete Markov decision processes under the average reward objective with the standard communicating assumption. 
We more specifically focus on the derivation of the instance dependent regret lower bound, that characterizes the performance frontier of what learning algorithms can achieve within the regret minimization framework.

\paragraph{Regret minimization in MDPs.}
Indeed, to quantify the learning performance of a learning agent, a popular approach is to compare the amount of reward that the learning agent and an ominous agent that knows everything in advance are able to collect with the same time budget.  
The difference between the cumulative rewards that the two gather is known as the \strong{regret}; the smaller the regret, the more efficient the learner.
The challenging task of designing learners with theoretical and practical regret guarantees has a long history, that goes back to multi-armed bandits. 
In this work, we follow the path of \cite{lai_asymptotically_1985} by focusing on the \strong{model dependent} setting: Given a Markov decision process, what is the best regret that a learner may hope to achieve? 
It happens that this question is only interesting when the focus is restricted to \strong{consistent} learners, that, roughly speaking, have an asymptotically small regret for every environment in a given class of Markov decision processes.
The seminal paper of \cite{lai_asymptotically_1985} emerges as the first model dependent regret lower bound for multi-armed bandits: for $\model$ a bandit, $\learner$ a consistent learner and $T$ a number of learning steps, the expected regret of $\learner$ when running on $\model$ for $T$ steps, denoted $\Reg(T; \model, \learner)$, is lower-bounded as
\begin{equation}
    \Reg(T; \model, \learner) \ge \regretlb(\model) \log(T) + \oh(\log(T))
    ,
\end{equation}
where $\regretlb(\model)$ is a known and tractable constant. 
\cite{lai_asymptotically_1985} also show that their bound is tight, by providing a family of consistent learners that achieve the above lower bound.
Such learners are said \strong{asymptotically optimal}. 
In the more general setting of Markov decision processes, the lower bound of \cite{lai_asymptotically_1985} has been generalized in various directions.
However, most of the literature seems to plateau at \strong{ergodic} environments \cite{agrawal_asymptotically_1988,burnetas_optimal_1997,graves1997asymptotically}, with very few exceptions such as \cite{tranos_regret_2021}.
Overall, the question of obtaining a tight regret lower bound for general \strong{communicating} Markov decision processes was left open for decades. 

\paragraph{Contribution.}
This article bridges this gap by deriving the regret lower bound $\regretlb(\model)$ for communicating Markov decision processes.
As our first contribution, we unravel the delicate structure of the general regret lower bound by describing it as the solution of an optimization problem that conjugates \emph{three} complementary behaviors:
(1)~consistent learners have to gather information on suboptimal regions (known as mandatory exploration, or information constraints);
(2)~they need to saturate the amount of information gathered on \emph{all} optimal regions (mandatory co-exploration);
(3)~the two previous behaviors must be achieved while preserving a second order structural constraint in the way navigate their environment (second order navigation constraints). 
Second, we show that the solution of this optimization problem is computationally hard in general by reducing it to a Knapsack problem. 
Third, we present multiple specification of the lower bound in several illustrative cases of interest, including classical ones and highlighting key simplifications.
Lastly, we present a constructive algorithm to approximate these bounds, offering both practical utility and further insight in the lower bound. 
Because of its delicacy, this paper is entirely dedicated to the lower bound and discusses it exhaustively. 
The construction of an explicit algorithm that matches our lower bound is deliberately postponed to future work.

\paragraph{Previous work on regret lower bounds.}
The lower bound for ergodic MDPs can be traced back to \cite{agrawal_asymptotically_1988} that generalize the lower bound and the asymptotically optimal algorithm of \cite{lai_asymptotically_1985} to ergodic MDPs.
The approach of \cite{agrawal_asymptotically_1988} is policy-wise and every policy is treated as an arm, each explored in turn until regeneration; Thus their approach is heavily relying on the ergodic nature of the environment and is suffering from a tremendous burn-in phase, due to the immensely large number of policies. 
\cite{burnetas_optimal_1997} circumvent this issue by decomposing the lower bound relatively to state-action pairs and provide a pair-wisely indexed learning algorithm that is asymptotically optimal. 
Specifically, instead of playing policies in a round robin fashion like \cite{agrawal_asymptotically_1988}, the algorithm of \cite{burnetas_optimal_1997} associates an index to every playable action and play the actions with the highest index. 
Their solution was recently improved and mixed to more modern bandit methods, for instance \cite{pirutinsky_asymptotically_2020} with \texttt{DMED} of \cite{honda_bounded_2010} or \cite{pesquerel_imed_rl_2022} with \texttt{IMED} of \cite{honda_non_asymptotic_2015}. 
In parallel, \cite{graves1997asymptotically} generalize \cite{lai_asymptotically_1985} to MDPs with compact state-action spaces but stick to the ergodic assumption. 
We have to wait for \cite{ok_exploration_2018} to find lower bounds again, generalizing \cite{burnetas_optimal_1997} but still keeping the ergodic assumption. 
The first paper to ever escape the world of ergodic MDPs is \cite{tranos_regret_2021}, that provides a regret lower bound for the very specific setting of deterministic transition MDPs together with algorithms that are asymptotically optimal in a few cases. 
To the best of our knowledge, general communicating Markov decision processes remained unsolved ever since. 

\paragraph{Related work on regret minimization.}
Our paper lives in the neighborhood of a large literature of algorithms with theoretical regret guarantees for average reward Markov decision processes. 
These works do not aim, however, at specifically matching the instance dependent regret lower bound. 
In the setting of average-reward regret minimization, model-based methods include optimistic approaches \cite{auer_near_optimal_2009,bartlett2012regal,filippi_optimism_2010,talebi_variance_aware_2018,fruit_efcient_2018,zhang_regret_2019,bourel_tightening_2020,boone_achieving_2024}, inspired from the popular Upper Confidence Bound (\texttt{UCB}) strategy \cite{auer_using_2002} and variants \cite{cappe2013kullback}; Bayesian approaches \cite{osband_more_2013,ouyang_learning_2017,agrawal_optimistic_2023} derived from
posterior sampling \cite{thompson_likelihood_1933}; Information-theoretic methods \cite{saber_logarithmic_2024,pesquerel_imed_rl_2022} derived from \cite{honda_non_asymptotic_2015} or \cite{agrawal1989certainty}; Regret minimization has also been considered in model-free methods \cite{zhang_sharper_2023} and references therein, based on Q-learning \cite{watkins1992q}, or extended e.g.~to smooth continuous MDPs \cite{lakshmanan2015improved, qian2019exploration}.
Quite often, these works actually aim at reaching the celebrated \strong{minimax} (or \strong{model independent}) regret lower bound of \cite{auer_near_optimal_2009} although many of the above mentioned are also provided with model dependent guarantees of order $\OH(\log(T))$ such as \cite{auer_logarithmic_2006,auer_near_optimal_2009,filippi_optimism_2010,pesquerel_imed_rl_2022,saber_logarithmic_2024}. 
Yet, none are proven to be asymptotically optimal in the general setup. 

\paragraph{Outline of the paper.}
The paper begins with the exposition of general background on Markov decision processes, reinforcement learning and regret minimization in \cref{sec:setup}.
\cref{section_main_ideas_and_concepts} serves as an informal and reader-friendly introduction to the main ideas and concepts behind our main result, the regret lower bound of \cref{theorem_lower_bound}.
\cref{sec:lowerbound} is entirely dedicated to proving the lower bound. 
In \cref{section_rejecting_alternative}, we quantify the necessary explorative behavior of consistent learners, showing that sub-optimal regions are explored logarithmically often.
We introduce and detail the concept of \strong{co-exploration} in \cref{section_visit_rates_of_optimal}, where consistent learners are shown to inevitably visit \emph{all} optimal regions of their environment over-logarithmically.
In \cref{section_minors}, we describe the navigation structure of how consistent learners may wander in their environment, showing that what matters is found at higher order (i.e., $\log(T)$) rather than first order (i.e., $T$). 
We rely on the novel notion of \strong{minors}, introduced in \cref{section_navigation_informal} and formalized in \cref{section_minors_formal_definition}.
Everything is combined together in \cref{section_lower_bound_contracted} for an intermediate regret lower bound, then adapted in \cref{section_lower_bound_without_contracted} into a simpler and final bound. 
In \cref{section_complexity}, we prove that the regret lower bound is computationally hard in general.
As a matter of fact, even the related information tests are provably NP-hard, see \cref{section_confusing_np_hard}. 
In \cref{section_examples}, we discuss many examples while drawing links with the existing literature. 
Our bound covers multi-armed bandits (\cref{section_example_bandits}), ergodic and recurrent Markov decision processes (\cref{section_example_recurrent}) that are shown to be special instances of the new class of \strong{optimally recurrent} models (\Cref{definition_optimally_recurrent}), bandits with switching costs (\cref{section_example_bandits_switching_costs}) as well as fixed kernel spaces (\cref{section_example_fixed_kernel_spaces}) with deterministic kernel spaces as special cases. 
We make our last stop at a policy-wise decomposition of the regret lower bound in \cref{section_example_policy_wise}.
In our final \cref{sec:contruction}, we discuss a special case of MDPs for which the optimality of optimal policies can be broken with only a few local modifications of the original model (\cref{section_locally_modifying}).
In this scenario, which extends constructions of the ergodic setting, the regret lower bound can be approximated using an algorithmic scheme (\cref{section_construction_of_confusing}).

    \tableofcontents
    
    \section{Preliminaries: Formal setup and notations}\label{sec:setup}
    
\allowdisplaybreaks

	\paragraph{General notations.} 
    For each $k\in\Nat$, $[k]$ denotes the set $\{1,\dots, k\}$.
    For a topological space $(\states, \mathcal{T})$, $\cP(\states)$ denotes the set of Borel probability measures on $\states$.
    Provided that $\cS, \pairs$ are discrete, a kernel $\bfp: \pairs \to \cP(\cS)$ is seen as a stochastic operator acting on functions $f:\cS\to\Real$. 
    Following folklore notations, $\bfp(s|\pair)$ denotes the mass of $\bfp(\pair)$ at point $s$.
We further use operator notation $\bfp f$ to denote $\pair \mapsto (\bfp f)(\pair) =\sum_{s\in\cS} \bfp(s|\pair)f(s)$.
    Given a probability distribution $\kerrew\in\cP(\cS)$, we denote $\Esp_\kerrew[f]$ or sometimes $\langle \kerrew, f\rangle$ the expected value of $f$ under $\kerrew$.
    We denote $\KL(\kernel||\kernel')$ the Kullback-Laibler divergence between $\kernel$ and $\kernel'$.
    For $u \in \R_+^n$ a non-negative vector, we write $\dmin(u) := \min\set{u_i : u_i > 0}$ its definite minimum, which is its smallest positive entry.

    For reference and reader-friendliness, an index of notations is provided in Appendix. 

    \subsection{Markov decision processes in average reward}

    A \strong{Markov decision process} (MDP) $\model \equiv (\states, \actions, \kerneld, \rewardd)$ consists in a four-tuple formed by a finite set of \strong{states} $\states$, finite sets of \strong{actions} indexed by states $\actions \equiv (\actions(\state))_{\state \in \states}$, together forming the \strong{pair space} $\pairs := \bigcup_{\state \in \states} \set{\state} \times \actions(\state)$; $\kerneld: \pairs \to \probabilities(\states)$ is the \strong{transition kernel}; and $\rewardd: \pairs \to \probabilities(\R)$ is the \strong{reward function}. 
    We denote $\reward: \pairs \to \R$ the mean reward vector, i.e., $\reward(\pair)$ is the mean of $\rewardd(\pair)$ for each $\pair \in \pairs$. 

    Managing a Markov decision process $\model$ by picking dynamically actions generates a trajectory in discrete time as follows: 
    At time $t \ge 1$ from the current state $\State_t$, playing an action $\Action_t$ chosen respective to the past of the trajectory makes the system switch to the new state $\State_{t+1} \sim \kerneld(\State_t, \Action_t)$ and produce a reward $\Reward_t \sim \rewardd(\State_t, \Action_t)$ independently of the past. 
    For short, $\Pair_t \equiv (\State_t, \Action_t)$ denotes the current pair.

    In the \strong{average reward setting}, the controlling agent aims at picking actions to maximize
    the average of all produced rewards $\Reward_1,\ldots, \Reward_T$ when $T$ goes to infinity. 

    \paragraph{Stationary policies.}
    A \strong{stationary policy}, or \strong{policy}, is a map $\policy : \states \to \probabilities(\actions)$. Its set is denoted $\randomizedpolicies$.
    Every policy $\policy$ induces a Markov reward process on $\states$ by acting on $\model$, of which the transition kernel is denoted $\kernel_\pi(\state) := \sum_{\action \in \actions(\state)} \policy(\action|\state) \kernel(\state, \action)$ and the mean reward function $\reward_\policy(\state) := \sum_{\action \in \actions(\state)} \policy(\action|\state) \reward(\state, \action)$. 
    The associated probability and expectation operators induced by iterating $\policy$ on $\model$ starting from $\state_0 \in \states_0$ are respectively $\Pr_{\state_0}^{\model, \policy} \set{-}$ and $\E_{\state_0}^{\model, \policy} [-]$. 
    It is well-known \cite[Chapters~8,9]{puterman2014markov} that every policy $\policy \in \randomizedpolicies$ has well-defined \strong{gain} and \strong{bias} functions, $\gain_\policy$ and $\bias_\policy$, given by:
    \begin{equation*}
    \begin{aligned}
        \gain_{\policy}(\state_0; \model)
        & :=
        \lim_{T \to \infty} 
        \frac 1T \E_{\state_0}^{\model, \pi} 
        \brackets*{
            \sum_{t=1}^T \Reward_t
        }
        =
        \lim_{T \to \infty} 
        \parens*{
            \frac 1T \sum_{t=1}^T \kernel_\pi^{t-1} \reward_\pi
        }(\state_0)
        ;
        \\
        \bias_{\policy}(\state_0; \model) 
        & :=
        \Clim_{T \to \infty}
        \E_{\state_0}^{\model, \pi} \brackets*{
            \sum_{t=1}^T \parens*{\Reward_t - \gain_{\pi}(\State_t; \model)}
        }
        =
        \Clim_{T \to \infty}
        \parens*{ 
            \sum_{t=1}^T \kernel_\pi^{t-1} (\reward_\pi - \gain_\pi)
        }(\state_0)
    \end{aligned}
    \end{equation*}
    where $\Clim$ denotes the Cesàro-limit. 
    A policy $\pi$ is said \strong{ergodic}, \strong{recurrent} or \strong{unichain} if the Markov chain of kernel $\kernel_\pi$ is respectively ergodic, recurrent or unichain, see \cite{levin_markov_2017,puterman2014markov}. 
    A state $\state \in \states$ is recurrent under a policy $\policy$ if it is visited infinitely often starting from itself, i.e., $\Pr_{\state}^{\model, \policy}\set{\forall n, \exists m \ge n: \State_m = \state} = 1$; or equivalently, if its return time has bounded expectation with $\E_{\state}^{\model, \policy}[\inf\set{t \ge 2: \State_t = \state}] < \infty$. 
    Likewise, a pair $\pair \equiv (\state, \action) \in \pairs$ is said recurrent under $\policy$ if it satisfies $\Pr_{\state}^{\model, \policy}\set{\forall n, \exists m \ge n: \Pair_m = \pair} = 1$.
    We write $\states_{\policy}^\opt(\model)$ and $\pairs_{\policy}^\opt(\model)$ the recurrent states and pairs of a policy. 

    In the remaining of our work, we make an important structural assumption on MDPs. 

    \begin{assumption}
    \label{assumption_communicating}
        All considered MDPs are \strong{communicating}, i.e., every $\model \in \models$ has finite \strong{diameter}:
        \begin{equation}
        \label{equation_diameter}
            \diameter(\model) 
            := 
            \max_{\state \ne \state'} 
            \min_{\policy \in \policies} 
            \E_{\state}^{\model, \policy}\brackets*{\inf\set{t \ge 1 : \State_t = \state'}} 
            < \infty
            .
        \end{equation}
    \end{assumption}

    Equivalently, a Markov decision process is communicating if every fully supported policy is recurrent, i.e., every policy satisfying $\policy(\action|\state) > 0$ for all $(\state, \action) \in \pairs$ is recurrent.

    \paragraph{Optimal play and Bellman equations.}
    The \strong{optimal gain} $\optgain(\model)$ and \strong{bias} $\optbias(\model)$ of $\model$ are given by:
    \begin{equation*}
        \optgain(\state_0; \model)
        :=
        \max_{\policy \in \randomizedpolicies} \gain_{\policy}(\state_0; \model)
        \text{\quad and \quad}
        \optbias(\state_0; \model)
        :=
        \max_{\policy \in \optpolicies(\model)} 
        \bias_{\policy}(\state_0; \model)\,,
    \end{equation*}
    and every policy achieving the maximum is said \strong{gain optimal}, written $\policy \in \optpolicies(\model)$. 
    When $\model$ is communicating, the optimal gain function does not depend on $\state_0$ and we will sometimes write $\optgain(\model)$ instead of $\optgain(\state_0; \model)$. 
    A gain optimal policy of which the bias reaches $\optbias(\state_0; \model)$ from every initial state is further said \strong{bias optimal}. 

    At the interplay of gain and bias optimalities stand the famous \strong{Bellman optimality equations}:
    \begin{equation}\label{eqn:Bellman}
        \forall \state \in \states,
        \quad
        \optgain(\state; \model) + \optbias(\state; \model)
        =
        \max_{\action \in \actions(\state)}
        \set*{
            \reward(\state, \action) + \kernel(\state, \action) \optbias(\model)
        }
    \end{equation}
    that can also be written as $\optgain(\model) + \optbias(\model) = \max_{\policy \in \randomizedpolicies} \set{\reward_\policy + \kernel_\policy \optbias(\model)}$.
    Every bias optimal policy satisfies the Bellman equations;
    Every policy satisfying the Bellman equations is gain optimal;
    Both reverse statements are wrong in general. 
    We introduce the \strong{Bellman gaps} $\ogaps(\state, \action; \model) := \optgain(\state; \model) + \optbias(\state; \model) - \reward(\state,\action) - \kernel(\state,\action) \optbias(\model)$. 
    The Bellman equations equivalently rewrite as $\ogaps(\model) \ge 0$. 

    \subsection{Reinforcement learning and regret}

    A \strong{learning agent}, or \strong{learner}, is formally a measurable map $\learner: \histories \to \probabilities(\actions)$, associating histories $\History_t \equiv (\State_1, \Action_1, \Reward_1, \ldots, \State_t) \in \histories$ to probabilities over actions.
    Given a MDP $\model$ and an initial state $\state_0$, a choice of learner completely determines the law of $\Action_t \in \actions(\State_t)$ conditioned on $\History_t$ for all $t \ge 1$, hence of $\State_t$ and $\Reward_t$ as well.
    We write $\Pr_{\state_0}^{\model, \learner}\set{-}$ and $\E_{\state_0}^{\model, \learner}[-]$ the associated probability and expectation operators. 

    A common benchmark to evaluate the performance of a learning agent is the \strong{regret}, that compares the amount of reward that the learning agent gathers to the amount of reward that an \emph{ominous} agent, that knows everything in advance and plays accordingly, is able to score within the same time budget. 
    The standard theory of Markov decision processes with average rewards \cite{puterman2014markov} explains that $\Reward_1 + \ldots + \Reward_T$ is equal to $T \optgain(\model) + \OH(1)$ in expectation, when actions are picked optimally with respect to time, the initial state $\state_0 \in \states$ and $\model$. 
    Following this observation, the regret compares the optimal score proxy $T \optgain(\state_0; \model)$ to what the learning agent actually collects, i.e., $\Reward_1 + \ldots + \Reward_T$.

    \begin{definition}[Regret, \cite{auer_near_optimal_2009}]
        The \strong{expected regret} of a learning agent $\learner$ in $\model$ at horizon $T \in \N$ and from the initial state $\state_0 \in \states$ is given by:
        \begin{equation}
            \Reg(T; \learner, \model, \state_0)
            :=
            \E_{\state_0}^{\model, \learner} \brackets*{
                T \optgain(\model)
                - 
                \sum_{t=1}^T \Reward_t
            }
        \end{equation}
    \end{definition}

    The smaller the regret, the better the learning agent. 
    In particular, there is convergence to optimal play if, and only if the regret grows sublinearly. 

    The quantity $T \optgain(\model) - \sum_{t=1}^T \Reward_t$, although easy to interpret, is a bit inconvenient because it is subject to various noises over which no learner has any control on.
    The so-called noise includes the stochasticity on observed rewards (i.e., $\Reward_t$ versus $\reward(\Pair_t)$) as well as the stochasticity on state transitions (i.e., $\State_{t+1}$ versus $\kernel(\Pair_t)$).
    Several variants of \strong{pseudo-regret} exist in the literature to circumvent this inconvenience, easing the empirical study and providing finer insight into a learner's quality of play.\footnote{
        There is no universally recognized definition of what a pseudo-regret actually is.
        In this work, we mean \strong{pseudo-regret} for any quantity of the form $\sum_{t=1}^T C_t$ with $C_t$ a $\sigma(\History_t, \Action_t)$-measurable quantity and $\E_{\state_0}^{\model, \learner}[\sum_{t=1}^T C_t] = \Reg(T; \learner, \model, \state_0) + \OH(1)$. 
    }
    In this work, we use  $\sum_{t=1}^T \ogaps(\Pair_t)$, as motivated by \Cref{proposition_pseudo_regret}.

    \begin{proposition}
    \label{proposition_pseudo_regret}
        For every communicating model $\model$, learner $\learner$ and initial state $\state_0 \in \states$, we have:
        \begin{equation}
            \abs*{
                \Reg(T; \learner, \model, \state_0)
                -
                \E_{\state_0}^{\model, \learner} \brackets*{
                    \sum_{t=1}^T \ogaps(\Pair_t)
                }
            }
            \le
            \mathrm{sp}(\optbias(\model))
        \end{equation}
        where $\mathrm{sp}(\optbias(\model)) = \max(\optbias(\model)) - \min(\optbias(\model))$ is the span of the optimal bias function.
    \end{proposition}
    \begin{proof}
        Use that $\E_{\state_0}^{\model, \learner}[\optgain(\model) - \Reward_t|\History_t] = \E_{\state_0}^{\model, \learner}[\ogaps(\Pair_t) + \optbias(\State_t) - \optbias(\State_{t+1})|\History_t]$ and sum over $t \le T$.
    \end{proof}

    Although the optimal policy depends on $\model$, a learning agent $\learner$ is only properly learning if there is convergence to optimal play \emph{regardless} of $\model$, that remains hidden. 
    Instead, a learning algorithm is tuned to work for a \strong{class} $\models$ of Markov decision processes and therefore is a function of that class.
    The class $\models$ is viewed as the assumptions that the learning agent makes on the hidden environment.
    Claiming that the learner eventually converges to optimal play for every $\model \in \models$ is informally referred to as \strong{consistency}, and can be stated at various degrees of strength. 
    For $\log(T)$-rate lower bounds to be even possible, we make the strong consistency assumption of \cite{lai_asymptotically_1985}, also known as uniformly fast convergence \cite{burnetas_optimal_1996,garivier_explore_2018,agrawal_asymptotically_1988} or even simply consistency \cite{lattimore_bandit_2020} in the literature. 
    This assumption can hardly be weakened without degrading the lower bound, see \cite{salomon2013lower}.

    \begin{definition}[Consistency]
    \label{definition_consistency}
        A learning agent $\learner$ is said \strong{consistent} on a space of Markov decision processes $\models$ if for all $\eta > 0$, all $\model \in \models$ and $\state_0 \in \states$, $\Reg(T; \learner, \model, \state_0) = \oh(T^\eta)$.
    \end{definition}

    Note that this consistency assumption is more concisely written $\log(\Reg(T; \learner, \model, \state_0)) = \oh(\log(T))$. 
    
    \section{Main ideas and concepts behind the lower bound}
    \label{section_main_ideas_and_concepts}

    We start by providing some general intuition on our regret lower bound and what are the main concepts behind our general regret lower bound. 

    \subsection{Exploration: Consistent learners must fetch information}
    \label{section_fetching_information_informal}

    \begin{figure}[h]
        \centering
        \begin{tikzpicture}
            \node at (0, 0) {$\model$};

            \node[draw, circle, inner sep=0cm, minimum size=0.7cm] (s1) at (55+0*360/5: 1.5cm) {$s_1$};
            \node[draw, circle, inner sep=0cm, minimum size=0.7cm] (s2) at (55+1*360/5: 1.5cm) {$s_2$};
            \node[draw, circle, inner sep=0cm, minimum size=0.7cm] (s3) at (55+2*360/5: 1.5cm) {$s_3$};
            \node[draw, circle, inner sep=0cm, minimum size=0.7cm] (s4) at (55+3*360/5: 1.5cm) {$s_4$};
            \node[draw, circle, inner sep=0cm, minimum size=0.7cm] (s5) at (55+4*360/5: 1.5cm) {$s_5$};

            \node[draw, circle, inner sep=0cm, minimum size=0.7cm] (s1') at (55+0*360/5: 2.5cm) {$s'_1$};
            \node[draw, circle, inner sep=0cm, minimum size=0.7cm] (s2') at (55+1*360/5: 2.5cm) {$s'_2$};
            \node[draw, circle, inner sep=0cm, minimum size=0.7cm] (s3') at (55+2*360/5: 2.5cm) {$s'_3$};
            \node[draw, circle, inner sep=0cm, minimum size=0.7cm] (s4') at (55+3*360/5: 2.5cm) {$s'_4$};
            \node[draw, circle, inner sep=0cm, minimum size=0.7cm] (s5') at (55+4*360/5: 2.5cm) {$s'_5$};

            \draw[color=Crimson, ->, >=stealth] (s1) to[bend left=0.7cm] node[pos=0.33, right] {$2$} (s5);
            \draw[color=Crimson, ->, >=stealth] (s5) to[bend left=0.7cm] node[pos=0.33, below] {$2$} (s4);
            \draw[color=Crimson, ->, >=stealth] (s4) to[bend left=0.7cm] node[pos=0.66, below] {$2$} (s3);
            \draw[color=Crimson, ->, >=stealth] (s3) to[bend left=0.7cm] node[pos=0.66, left] {$2$} (s2);
            \draw[color=Crimson, ->, >=stealth] (s2) to[bend left=0.7cm] node[midway, above] {$2$} (s1);

            \draw[<-, >=stealth, dash pattern=on 3pt off 2pt] (s1') to[bend left=0.8cm] node[pos=0.33, right] {$1$} (s5');
            \draw[<-, >=stealth, dash pattern=on 3pt off 2pt] (s5') to[bend left=0.8cm] node[pos=0.33, below] {$1$} (s4');
            \draw[<-, >=stealth, dash pattern=on 3pt off 2pt] (s4') to[bend left=0.8cm] node[pos=0.66, below] {$1$} (s3');
            \draw[<-, >=stealth, dash pattern=on 3pt off 2pt] (s3') to[bend left=0.8cm] node[pos=0.66, left] {$1$} (s2');
            \draw[<-, >=stealth, dash pattern=on 3pt off 2pt] (s2') to[bend left=0.8cm] node[midway, above] {$1$} (s1');

            \draw[->, >=stealth] (s1) to node[pos=-.1, right] {$0$} (s1');
            \draw[->, >=stealth] (s2') to node[pos=1.1, left] {$0$} (s2);
        \end{tikzpicture}
        \hspace{5em}
        \begin{tikzpicture}
            \node at (0, 0) {$\model^\dagger$};

            \node[draw, circle, inner sep=0cm, minimum size=0.7cm] (s1) at (53+0*360/5: 1.5cm) {$s_1$};
            \node[draw, circle, inner sep=0cm, minimum size=0.7cm] (s2) at (53+1*360/5: 1.5cm) {$s_2$};
            \node[draw, circle, inner sep=0cm, minimum size=0.7cm] (s3) at (53+2*360/5: 1.5cm) {$s_3$};
            \node[draw, circle, inner sep=0cm, minimum size=0.7cm] (s4) at (53+3*360/5: 1.5cm) {$s_4$};
            \node[draw, circle, inner sep=0cm, minimum size=0.7cm] (s5) at (53+4*360/5: 1.5cm) {$s_5$};

            \node[draw, circle, inner sep=0cm, minimum size=0.7cm] (s1') at (53+0*360/5: 2.5cm) {$s'_1$};
            \node[draw, circle, inner sep=0cm, minimum size=0.7cm] (s2') at (53+1*360/5: 2.5cm) {$s'_2$};
            \node[draw, circle, inner sep=0cm, minimum size=0.7cm] (s3') at (53+2*360/5: 2.5cm) {$s'_3$};
            \node[draw, circle, inner sep=0cm, minimum size=0.7cm] (s4') at (53+3*360/5: 2.5cm) {$s'_4$};
            \node[draw, circle, inner sep=0cm, minimum size=0.7cm] (s5') at (53+4*360/5: 2.5cm) {$s'_5$};

            \draw[color=Crimson, ->, >=stealth] (s1) to[bend left=0.7cm] node[pos=0.33, right] {$2$} (s5);
            \draw[color=Crimson, ->, >=stealth] (s5) to[bend left=0.7cm] node[pos=0.33, below] {$2$} (s4);
            \draw[color=Crimson, ->, >=stealth] (s4) to[bend left=0.7cm] node[pos=0.66, below] {$2$} (s3);
            \draw[color=Crimson, ->, >=stealth] (s3) to[bend left=0.7cm] node[pos=0.66, left] {$2$} (s2);
            \draw[color=Crimson, ->, >=stealth] (s2) to[bend left=0.7cm] node[midway, above] {$2$} (s1);

            \draw[color=blue, <-, >=stealth, dash pattern=on 3pt off 2pt] (s1') to[bend left=0.8cm] node[pos=0.33, right] {$2.1$} (s5');
            \draw[color=blue, <-, >=stealth, dash pattern=on 3pt off 2pt] (s5') to[bend left=0.8cm] node[pos=0.33, below] {$2.1$} (s4');
            \draw[color=blue, <-, >=stealth, dash pattern=on 3pt off 2pt] (s4') to[bend left=0.8cm] node[pos=0.66, below] {$2.1$} (s3');
            \draw[color=blue, <-, >=stealth, dash pattern=on 3pt off 2pt] (s3') to[bend left=0.8cm] node[pos=0.66, left] {$2.1$} (s2');
            \draw[color=blue, <-, >=stealth, dash pattern=on 3pt off 2pt] (s2') to[bend left=0.8cm] node[midway, above] {$2.1$} (s1');

            \draw[->, >=stealth] (s1) to node[pos=-.1, right] {$0$} (s1');
            \draw[->, >=stealth] (s2') to node[pos=1.1, left] {$0$} (s2);
        \end{tikzpicture}
        \caption{
            \label{figure_todo}
            Two 10 state Markov decision processes $\model$ and $\model^\dagger$ with deterministic transitions.
            Every arrow is a choice of action that leads to the same pointed state with probability one and with reward $\mathrm{N}(x, 1)$ where $x \in \set*{0, 1, 2}$ is the label of the arrow. 
        }
    \end{figure}
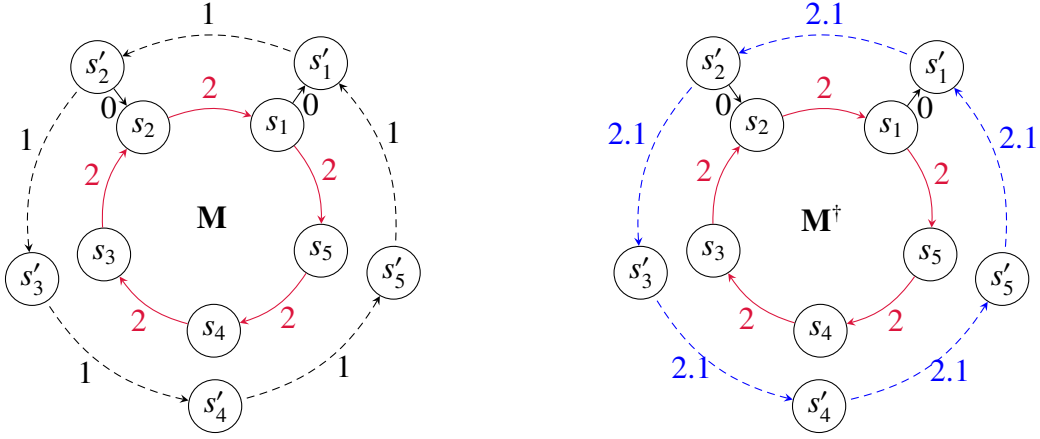

    Consider running a consistent learning agent $\learner$. 
    On both of the environments $\model$ and $\model^\dagger$ displayed in \cref{figure_todo}, such a learning algorithm is supposed to converge to optimal play; Specifically, it must eventually discover how to drive and navigate the environment in order to collect as much reward as possible and its efficiency is measured by $\Reg(T; \learner, \model, \state_0)$. 
    Every regret lower bound essentially says that, provided that an algorithm is consistent, it cannot learn too fast. 
    Accordingly, that every of such algorithm has some mandatory behavior.

    Looking at $\model$ in more detail, it is made of two circles of length five; the outer and the inner circle, scoring $1$ and $2$ in average respectively. 
    Especially, every consistent learner should quickly discover that the inner circle is the best. 
    However, such a learner can only prioritize playing this inner optimal circle up to a limited extent. 
    To understand why this is exactly the case, we consider the alternative model $\model^\dagger$. 
    On $\model^\dagger$, the inner circle (in red) is a copy of the one of $\model$.
    In particular, $\model$ and $\model^\dagger$ are indistinguishable if one only observes what happens on the red circle.
    However, on $\model^\dagger$, the average reward on the outer circle is increased to $2.1$, so that the outer circle is actually better than the inner one. 
    This means that on $\model^\dagger$, efficient learners should spend most of their time on the outer circle instead. 
    Yet, if on $\model$ a learner spends too little time on the outer circle, then there are decent chances that when playing on $\model^\dagger$, the empirically observed average reward of the outer circle is close to $1$ (because rewards are random), hence this learner is equally likely to mistake $\model^\dagger$ for $\model$, and to commit to the red inner circle that is actually suboptimal in $\model^\dagger$. 
    Overall, on $\model$, and because the learner is able to learn $\model^\dagger$ correctly, the outer circle cannot be too little explored because the learner must \strong{gather information} in order to reject the plausibility that the actual environment is $\model^\dagger$ rather than $\model$. 

    This idea goes back to \cite{lai_asymptotically_1985} in multi-armed bandits:
    Consistent learners must, on $\model$, fetch information in order to reject environments such as $\model^\dagger$.
    Such environments are somehow indistinguishable from $\model$ if one only plays optimal actions of $\model$ yet optimal actions of $\model$ are not guaranteed to be optimal anymore; They are called \strong{confusing models}. 
    More precisely, $\model$ and $\model^\dagger$ are indistinguishable in that they are identical at every pair $\pair \in \pairs$ such that there exists an optimal policy $\policy \in \optpolicies(\model)$ under which $\pair$ is recurrent, i.e., $\pair \in \pairs_{\policy}^\opt(\model)$. 
    Such pairs are said \strong{optimal}, written $\pair \in \optpairs(\model)$, and the first requirement for a confusing model is that $\model^\dagger(\pair) = \model(\pair)$ for all $\pair \in \optpairs(\model)$.

    Following this intuition, pairs are classified as follows.

    \begin{definition}[Classification of pairs]
        The pairs of a Markov decision process $\model$ are split as follows:
        \begin{align*}
            (\strong{Optimal pairs})
            \quad
            \optpairs(\model) 
            & := \bigcup \set*{
                \pairs^{\opt}_\policy(\model)
                :
                \policy \in \optpolicies(\model)
            }
            ~;
            \\
            (\strong{Weakly optimal pairs})
            \quad
            \wkoptpairs(\model)
            & := \set*{\pair \in \pairs : \ogaps(\pair; \model) = 0}
            ~;
            \\
            (\strong{Suboptimal pairs})
            \quad
            \suboptimalpairs(\model)
            & := \pairs \setminus \wkoptpairs(\model)
            ~.
        \end{align*}
    \end{definition}

    The key differences between optimal and weakly optimal pairs, and their different roles regarding the description of a consistent learner's behavior, will be discussed when time comes, see \cref{section_visit_rates_of_optimal}.
    For now, we will stick to the observation that optimal pairs are indeed weakly optimal (\cref{lemma_optimal_are_weakly_optimal}).

    \begin{lemma}
    \label{lem:opt_are_wkopt}
    \label{lemma_optimal_are_weakly_optimal}
        Assuming $\model$ is communicating, we have the inclusion $\optpairs(\model) \subseteq \wkoptpairs(\model)$. 
    \end{lemma}
    \begin{proof}
        Let $\pair \equiv (\state, \action) \in \optpairs(\model)$.
        We have $\pair \in \pairs_{\policy}^\opt(\model)$ for some $\policy \in \optpolicies(\model)$. 
        Starting from $\state$ and upon iterating $\policy$, $\pair$ is visited infinitely many times, so because the number of states is finite, $\E_{\state}^{\model, \policy}[\sum_{t=1}^T \indicator{\Pair_t = \pair}] = \Omega(T)$.
        Combined with \cref{proposition_pseudo_regret}, we deduce that $\Reg(T; \policy, \model, \state) = \Omega(\ogaps(\pair; \model) T)$. 
        Since $\policy \in \optpolicies(\model)$, we have $\Reg(T; \policy, \model, \state) = \oh(T)$ hence necessarily $\ogaps(\pair; \model) = 0$, i.e., $\pair \in \wkoptpairs(\model)$.
    \end{proof}

    The second requirement for a confusing model $\model^\dagger$ is that $\optpolicies(\model^\dagger) \cap \optpolicies(\model) = \emptyset$. 

    \begin{definition}[Confusing models]
        \label{definition_confusing_model}
        Fix $\model \in \models$.
        The \strong{confusing set} of $\model$ is given by:
        \begin{align*}
            \confusing(\model)
            & :=
            \set*{
                \model^\dagger \in \models
                :
                \model^\dagger \gg \model
                \text{~and~}
                \model^\dagger = \model \text{~on~} \optpairs(\model)
                \text{~and~}
                \optpolicies(\model^\dagger) \cap \optpolicies(\model) = \emptyset
            }
            .
        \end{align*}
    \end{definition}

    In \cref{definition_confusing_model} and in the sequel, we write $\model^\dagger \gg \model$ to mean that $\kerneld(\pair) \ll \kerneld^\dagger(\pair)$ and $\rewardd(\pair) \ll \rewardd^\dagger(\pair)$ for all $\pair \in \pairs$. 
    This absolute continuity assumption is linked to changes of measures and is crucial to the underlying proof techniques. 

    \subsection{Navigation: The higher order navigational structure}
    \label{section_navigation_informal}
    
    Yet, if fetching information is the first ingredient of our regret lower bound, it is insufficient to obtain a tight one.
    This makes quite a big difference with multi-armed bandits \cite{lai_asymptotically_1985} and ergodic Markov decision processes \cite{agrawal_asymptotically_1988,burnetas_optimal_1997,graves1997asymptotically}.
    In general, every MDP has a dynamical structure, so it cannot be explored freely and every learner is confined within the exploration possibilities allowed by the transition kernel. 
    On $\model$ in \cref{figure_todo}, it is rather obvious that the number of times $s'_4$ is visited should be about equal to the number of times $s'_3$ is visited. 
    Formally, the \strong{visit vector} $\visits_T \in \R^\pairs$ given by $\visits_T(\pair) := \sum_{t=1}^{T-1} \indicator{\Pair_t = \pair}$, mapping state-action pairs to their current number of visits, is nearly a flow of the MDP by satisfying an ``entering mass equals outgoing mass'' property. 

    In this work, we instead talk of \strong{invariant measures} (see \cref{definition_invariant_measures}).

    \begin{definition}
    \label{definition_invariant_measures}
        An \strong{invariant measure} $\imeasure \in \imeasures(\model)$ of a Markov decision process $\model$ is a measure on $\pairs$ such that:
        \begin{equation*}
            \forall \state \in \states,
            \quad
            \sum_{\pair \in \pairs}
            \kernel(\state|\pair) \imeasure(\pair)
            =
            \sum_{\action \in \actions(\state)}
            \imeasure(\state, \action)
            .
        \end{equation*}
    \end{definition}

    Unfortunately, we are not interested in the first order structure of the visit vector nor of the associated invariant measure, contrasting with the literature on provably approximately correct (PAC) learning, see \cite{marjani_navigating_2021,tirinzoni_near_2022,russo_model_free_2024,marjani_active_2023}. 
    For every consistent learner, as the regret is growing sublinearly, all the mass of $\visits_T$ provably concentrates on optimal pairs and optimal pairs do not account for the regret. 
    Thankfully, this dominant part can be discarded with a structural operation on $\model$ that we call \strong{pair space contraction} which is close in spirit to the state aggregation operation of \cite{ortner2013adaptive}: Every communicating component of $\model|_{\optpairs(\model)}$ (the original model constrained to optimal pairs) is contracted into a single state, providing a new contracted Markov decision process denoted $\model/\optpairs(\model)$. 

\begin{figure}[ht]
        \centering
        \begin{tikzpicture}
            \node at (0, 0) {$\model$};

            \node[draw, circle, inner sep=0cm, minimum size=0.7cm] (s1) at (53+0*360/5: 1.5cm) {$s_1$};
            \node[draw, circle, inner sep=0cm, minimum size=0.7cm] (s2) at (53+1*360/5: 1.5cm) {$s_2$};
            \node[draw, circle, inner sep=0cm, minimum size=0.7cm] (s3) at (53+2*360/5: 1.5cm) {$s_3$};
            \node[draw, circle, inner sep=0cm, minimum size=0.7cm] (s4) at (53+3*360/5: 1.5cm) {$s_4$};
            \node[draw, circle, inner sep=0cm, minimum size=0.7cm] (s5) at (53+4*360/5: 1.5cm) {$s_5$};

            \node[draw, circle, inner sep=0cm, minimum size=0.7cm] (s1') at (53+0*360/5: 2.5cm) {$s'_1$};
            \node[draw, circle, inner sep=0cm, minimum size=0.7cm] (s2') at (53+1*360/5: 2.5cm) {$s'_2$};
            \node[draw, circle, inner sep=0cm, minimum size=0.7cm] (s3') at (53+2*360/5: 2.5cm) {$s'_3$};
            \node[draw, circle, inner sep=0cm, minimum size=0.7cm] (s4') at (53+3*360/5: 2.5cm) {$s'_4$};
            \node[draw, circle, inner sep=0cm, minimum size=0.7cm] (s5') at (53+4*360/5: 2.5cm) {$s'_5$};

            \draw[color=Crimson, ->, >=stealth] (s1) to[bend left=0.7cm] node[pos=0.33, right] {$2$} (s5);
            \draw[color=Crimson, ->, >=stealth] (s5) to[bend left=0.7cm] node[pos=0.33, below] {$2$} (s4);
            \draw[color=Crimson, ->, >=stealth] (s4) to[bend left=0.7cm] node[pos=0.66, below] {$2$} (s3);
            \draw[color=Crimson, ->, >=stealth] (s3) to[bend left=0.7cm] node[pos=0.66, left] {$2$} (s2);
            \draw[color=Crimson, ->, >=stealth] (s2) to[bend left=0.7cm] node[midway, above] {$2$} (s1);

            \draw[<-, >=stealth, dash pattern=on 3pt off 2pt] (s1') to[bend left=0.8cm] node[pos=0.33, right] {$1$} (s5');
            \draw[<-, >=stealth, dash pattern=on 3pt off 2pt] (s5') to[bend left=0.8cm] node[pos=0.33, below] {$1$} (s4');
            \draw[<-, >=stealth, dash pattern=on 3pt off 2pt] (s4') to[bend left=0.8cm] node[pos=0.66, below] {$1$} (s3');
            \draw[<-, >=stealth, dash pattern=on 3pt off 2pt] (s3') to[bend left=0.8cm] node[pos=0.66, left] {$1$} (s2');
            \draw[<-, >=stealth, dash pattern=on 3pt off 2pt] (s2') to[bend left=0.8cm] node[midway, above] {$1$} (s1');

            \draw[->, >=stealth] (s1) to node[pos=-.1, right] {$0$} (s1');
            \draw[->, >=stealth] (s2') to node[pos=1.1, left] {$0$} (s2);
        \end{tikzpicture}
        \hfill
        \begin{tikzpicture}
            \node at (0, 0) {$\model|_{\optpairs(\model)}$};

            \node[draw, circle, inner sep=0cm, minimum size=0.7cm] (s1) at (53+0*360/5: 1.5cm) {$s_1$};
            \node[draw, circle, inner sep=0cm, minimum size=0.7cm] (s2) at (53+1*360/5: 1.5cm) {$s_2$};
            \node[draw, circle, inner sep=0cm, minimum size=0.7cm] (s3) at (53+2*360/5: 1.5cm) {$s_3$};
            \node[draw, circle, inner sep=0cm, minimum size=0.7cm] (s4) at (53+3*360/5: 1.5cm) {$s_4$};
            \node[draw, circle, inner sep=0cm, minimum size=0.7cm] (s5) at (53+4*360/5: 1.5cm) {$s_5$};

            \node[color=white, draw, circle, inner sep=0cm, minimum size=0.7cm] (s1') at (53+0*360/5: 2.5cm) {$s'_1$};
            \node[color=white, draw, circle, inner sep=0cm, minimum size=0.7cm] (s2') at (53+1*360/5: 2.5cm) {$s'_2$};
            \node[color=white, draw, circle, inner sep=0cm, minimum size=0.7cm] (s3') at (53+2*360/5: 2.5cm) {$s'_3$};
            \node[color=white, draw, circle, inner sep=0cm, minimum size=0.7cm] (s4') at (53+3*360/5: 2.5cm) {$s'_4$};
            \node[color=white, draw, circle, inner sep=0cm, minimum size=0.7cm] (s5') at (53+4*360/5: 2.5cm) {$s'_5$};

            \draw[color=Crimson, ->, >=stealth] (s1) to[bend left=0.7cm] node[pos=0.33, right] {$2$} (s5);
            \draw[color=Crimson, ->, >=stealth] (s5) to[bend left=0.7cm] node[pos=0.33, below] {$2$} (s4);
            \draw[color=Crimson, ->, >=stealth] (s4) to[bend left=0.7cm] node[pos=0.66, below] {$2$} (s3);
            \draw[color=Crimson, ->, >=stealth] (s3) to[bend left=0.7cm] node[pos=0.66, left] {$2$} (s2);
            \draw[color=Crimson, ->, >=stealth] (s2) to[bend left=0.7cm] node[midway, above] {$2$} (s1);

            \draw[color=white, <-, >=stealth, dash pattern=on 3pt off 2pt] (s1') to[bend left=0.8cm] node[pos=0.33, right] {$1$} (s5');
            \draw[color=white, <-, >=stealth, dash pattern=on 3pt off 2pt] (s5') to[bend left=0.8cm] node[pos=0.33, below] {$1$} (s4');
            \draw[color=white, <-, >=stealth, dash pattern=on 3pt off 2pt] (s4') to[bend left=0.8cm] node[pos=0.66, below] {$1$} (s3');
            \draw[color=white, <-, >=stealth, dash pattern=on 3pt off 2pt] (s3') to[bend left=0.8cm] node[pos=0.66, left] {$1$} (s2');
            \draw[color=white, <-, >=stealth, dash pattern=on 3pt off 2pt] (s2') to[bend left=0.8cm] node[midway, above] {$1$} (s1');

            \draw[color=white, ->, >=stealth] (s1) to node[pos=-.1, right] {$0$} (s1');
            \draw[color=white, ->, >=stealth] (s2') to node[pos=1.1, left] {$0$} (s2);
        \end{tikzpicture}
        \hfill
        \begin{tikzpicture}
            \node at (0, -1) {$\model/\optpairs(\model)$};

            \node[draw, ellipse, inner sep=0.2em] (quotient) at (0, 0) {$[\state_1, \state_2, \state_3, \state_4, \state_5]$};

            \node[draw, circle, inner sep=0cm, minimum size=0.7cm] (s1') at (55+0*360/5: 2.5cm) {$s'_1$};
            \node[draw, circle, inner sep=0cm, minimum size=0.7cm] (s2') at (55+1*360/5: 2.5cm) {$s'_2$};
            \node[draw, circle, inner sep=0cm, minimum size=0.7cm] (s3') at (55+2*360/5: 2.5cm) {$s'_3$};
            \node[draw, circle, inner sep=0cm, minimum size=0.7cm] (s4') at (55+3*360/5: 2.5cm) {$s'_4$};
            \node[draw, circle, inner sep=0cm, minimum size=0.7cm] (s5') at (55+4*360/5: 2.5cm) {$s'_5$};

            \draw[<-, >=stealth, dash pattern=on 3pt off 2pt] (s1') to[bend left=0.8cm] node[pos=0.33, right] {$1$} (s5');
            \draw[<-, >=stealth, dash pattern=on 3pt off 2pt] (s5') to[bend left=0.8cm] node[pos=0.33, below] {$1$} (s4');
            \draw[<-, >=stealth, dash pattern=on 3pt off 2pt] (s4') to[bend left=0.8cm] node[pos=0.66, below] {$1$} (s3');
            \draw[<-, >=stealth, dash pattern=on 3pt off 2pt] (s3') to[bend left=0.8cm] node[pos=0.66, left] {$1$} (s2');
            \draw[<-, >=stealth, dash pattern=on 3pt off 2pt] (s2') to[bend left=0.8cm] node[midway, above] {$1$} (s1');

            \draw[->, >=stealth] (s2') to node[midway, left] {$0$} (quotient.north west);
            \draw[<-, >=stealth] (s1') to node[midway, right] {$0$} (quotient.north east);
            \draw[->, >=stealth, loop, looseness=7, color=Crimson] (quotient) to node[midway, above] {$2$} (quotient);
        \end{tikzpicture}
        \caption{
            \label{figure_contraction_discussion}
            The contraction of $\model$ by $\optpairs(\model)$.
            The states spawning $\optpairs(\model)$ are merged together into a single state, forming a new Markov decision process $\model/\optpairs(\model)$, where the navigational behavior of a consistent learner is better understood. 
        }
    \end{figure}
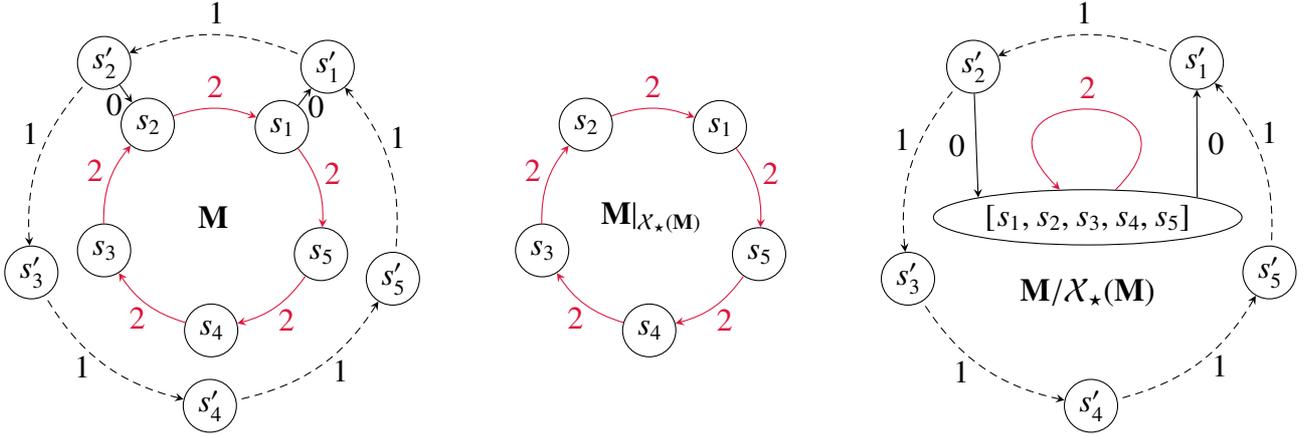

    This operation is illustrated in \cref{figure_contraction_discussion} and described formally in \cref{definition_minors}, see \cref{section_minors}.

    On \cref{figure_contraction_discussion}, the states spawned by $\optpairs(\model)$ are merged into a single state.
    The visit vector $\visits_T$ satisfies a near-flow property on $\model/\optpairs(\model)$ as well, but this time all pairs of $\optpairs(\model)$ are loops; And loops are trivial regarding flows hence are utterly discarded. 
    What remains is the second order structure of $\visits_T$, which is nothing less than a near-flow of $\model/\optpairs(\model)$. 
    It follows that the second order structure of $\visits_T$ converges to an invariant measure of $\model/\optpairs(\model)$ up to rescaling, as formally described by \cref{corollary_navigation_constraints}.

    A third, slightly more subtile ingredient will come refine what happens regarding exploration.
    This one, that we refer to as \strong{co-exploration}, will be discussed more carefully in \cref{section_visit_rates_of_optimal}. 
    Combining all these ideas, we derive that consistent learners must fetch information to ``reject'' confusing models while preserving the structure of a contracted version of $\model$ at second order; These ingredients are enough to derive a complete regret lower bound in \cref{theorem_lower_bound}, hence they completely describe the mandatory asymptotic behavior of consistent learners.

    \section{An asymptotic regret lower bound}\label{sec:lowerbound}
    \allowdisplaybreaks

This section is dedicated to the main contribution of this paper in the form of \cref{theorem_lower_bound}. 
We provide an asymptotic lower bound on the cumulative regret of strong consistent learning agents under a communicating assumption.

\begin{theorem}[Regret lower bound]
	\label{theorem_lower_bound}
	Let $\mdp \in \models$ communicating.
    The regret of every consistent learning agent on $\models$ satisfies $\liminf\limits_{T \to \infty} \Reg(T;\alg,\mdp) \log^{-1}(T) \ge \regretlb_{\jopt}(\mdp)$ where $\regretlb_{\jopt}(\mdp) \in [0, \infty]$ is:
	\begin{equation}
		\label{equation_lower_bound}
        \tag{LB}
        \regretlb_{\jopt}(\model) =
		\inf \set*{
            \sum_{\pair \in \pairs} \imeasure(\pair) \ogaps(\pair)
			:
            \imeasure \in \imeasures(\model)
			\text{~and~}
            \inf_{\model^\dagger \in \confusing(\model)}
            \set*{
			    \sum_{\pair \in \pairs} \imeasure(\pair) \KL_\pair (\mdp||\mdp^\dagger) 
            } \ge 1
		}
		.
	\end{equation}
\end{theorem}

In the definition of $\regretlb_{\jopt}(\model)$,
we introduced 
$\KL_x(\mdp||\mdp^\dagger) := \KL(\kernel(\pair)||\kernel^\dagger(\pair))+\KL(\rewardd(\pair)||\rewardd^\dagger(\pair))$.

The measure $\imeasure \in \R^\pairs$ tracks the number of times pairs are being \strong{explored} (in logarithmic scale), and an element $\imeasure^*$ that achieves the minimum of \eqref{equation_lower_bound} can be interpreted as an optimal way to explore the environment and gather information.
In that regard, such a $\imeasure^*$ will be referred to as an \strong{optimal exploration measure} and indirectly defines a randomized policy given by $\policy(\action|\state) \propto \imeasure^*(\state, \action)$ that can be used to explore optimally.
In light of this terminology, the result of \cref{theorem_lower_bound} can be read as follows.
It claims that 
(1) any optimal exploration measure $\imeasure^*$ must respect the structure of the Markov decision process, with $\imeasure^* \in \imeasures(\model)$; 
and 
(2) any optimal exploration measure must be informative enough, with $\inf \braces[\big]{\sum_{\pair \in \pairs} \imeasure(\pair) \KL_\pair(\model||\model^\dagger) : {\model^\dagger \in \confusing(\model)}} \ge 1$.
The second condition means that, when explorating according to $\imeasure^*$, one visits all the pairs in such a way that the learning agent can assess with sufficiently high confidence that seemingly optimal pairs are indeed optimal.

\paragraph{Outline of the section.}
The purpose of this section is to prove \cref{theorem_lower_bound}.
We start by following the ideas motivated in \cref{section_main_ideas_and_concepts}, starting with information constraints in \cref{section_rejecting_alternative}. 
These information constraints are given relatively to a super-set of the confusing set (\cref{definition_confusing_model}), called the alternative set (\cref{definition_alternative_model}). 
The resulting information constraints embody the idea that consistent learners must \strong{explore}, by wandering the suboptimal regions to make sure that they do not miss the optimal policies.
In \cref{section_visit_rates_of_optimal}, we show that \strong{every} optimal region should be explored way more than sub-optimal regions, a phenomenon that we refer to as \strong{co-exploration}.
This leads to the simplification of the alternative set to the confusing set. 
In \cref{section_minors}, we dive into navigation constraints by properly defining minors and showing that the second order structure of the visit vector converges in expectation to an invariant measure of a well-chosen minor. 
In \cref{section_lower_bound_contracted}, we mix these three ingredients together to provide a first regret lower bound with \cref{theorem_intermediate_lower_bound}, given in terms of the confusing set $\confusing(\model)$ and contracted invariant measures $\imeasures(\model/\optpairs(\model))$.
We finally prove \cref{theorem_lower_bound} from \cref{theorem_intermediate_lower_bound} in \cref{section_lower_bound_without_contracted} by showing that elements of $\imeasures(\model/\optpairs(\model))$ can be \strong{represented} by elements of $\imeasures(\model)$ without changing the lower bound. 
We conclude by showing the importance of navigation constraints in \cref{section_mandatory_navigation_constraints}.

\subsection{Exploration: Fetching information to reject alternative models}
\label{section_rejecting_alternative}

In \cref{section_rejecting_alternative,section_visit_rates_of_optimal}, we formalize the informal idea of \cref{section_fetching_information_informal}: that consistent learners must gather information to make sure that they do not miss the optimal policy.
This is achieved in a two-phase process. 
We start by proving that consistent learning agent must explore their environments $\model$ so that $\E_{\state_0}^{\model, \learner}\brackets[\big]{\sum_{\pair \in \pairs} \visits_{T+1}(\pair) \KL_\pair(\model||\model^\dagger)} \gtrsim \log(T)$ for every \strong{alternative} model $\model^\dagger \in \alternative(\model)$. 
Alternative models (\cref{definition_alternative_model}) are all the environments $\model^\dagger$ such that $\optpolicies(\model^\dagger) \cap \optpolicies(\model) = \emptyset$, and confusing models (\cref{definition_confusing_model}) are therefore special cases of alternative models.
We will show that confusing models can replace alternative ones without degrading the lower bound in \cref{section_visit_rates_of_optimal}.

The purpose of this paragraph is to provide a complete overview of the techniques and intuitions behind the main result of this paragraph, see \cref{proposition_rejecting_alternative} below.

We motivate the introduction of alternative models to begin with. 
In \cref{section_fetching_information_informal}, the motivational example makes obvious that any $\model^\dagger$ such that ``$\model^\dagger = \model$ on $\optpairs(\model)$'' doesn't hold will be trivially rejected by consistent learners. 
This is properly shown in \cref{section_visit_rates_of_optimal} under mild assumptions, and this behavior is to be distinguished with the fetching of information displayed in this section, usually known as \strong{exploration}.
For the time being, the assumption ``$\model^\dagger = \model$ on $\optpairs(\model)$'' in the definition of confusing models is dropped, resulting in the broader notion of alternative models, named accordingly to the works of \cite{marjani_adaptive_2021,marjani_navigating_2021,marjani_active_2023} in the PAC setting. 

\begin{definition}[Alternative models]
    \label{definition_alternative_model}
    Fix $\model \in \models$.
    The \strong{alternative set} of $\model$ is given by:
    \begin{align*}
        \alternative(\model)
        & :=
        \set*{
            \model^\dagger \in \models
            :
            \model^\dagger \gg \model
            \text{~and~}
            \optpolicies(\model^\dagger) \cap \optpolicies(\model) = \emptyset
        }
        .
    \end{align*}
\end{definition}

It is obvious that $\confusing(\model) \subseteq \alternative(\model)$.
The informal argument of information gathering is formalized by the following result.

\begin{proposition}[Information constraint]
\label{proposition_rejecting_alternative}
    Let $\model \in \models$ and let $\model^\dagger \in \alternative(\model)$.
    Whatever the initial state $\state_0 \in \states$, every consistent learner $\learner$ satisfies:
    \begin{equation*}
        \E^{\model, \learner}_{\state_0}\brackets*{
            \sum_{\pair \in \pairs}
            \visits_{T+1}(\pair)
            \KL_\pair (\model||\model^\dagger)
        }
        \ge 
        \log(T) + \oh\parens*{\log(T)}.
    \end{equation*}
\end{proposition}

\paragraph{Changes of measures.}
The core of the proof is to show that a consistent learner $\learner$ cannot behave similarly on $\model$ and $\model^\dagger$ if $\model^\dagger \in \alternative(\model)$.
The learner $\learner$ explores \emph{because} it cannot perform too poorly on $\model^\dagger$, even though there is a positive probability that when navigating $\model^\dagger$, the observed data that resembles $\model$ much more than $\model^\dagger$. 
This informal argument is concretized using {log-likelihood ratios} to relate the behavior of the learning agent on $\model^\dagger$ to its behavior on $\model$.
By doing a change of measure, it happens (see \cref{theorem_change_measure}) that for all $\sigma(\History_T)$-measurable random variable $U \ge 0$ and all $\model^\dagger \gg \model$, we have:
\begin{equation}
\label{equation_change_of_measure}
    \E^{\model^\dagger, \learner}_{\state_0}[U]
    \ge
    \E^{\model, \learner}_{\state_0}\brackets*{
        U \exp\braces*{
            - \sum_{t=1}^T \log\parens*{
                \frac{\kerneld(\State_{t+1}|\State_t,\Action_t)}{\kerneld^\dagger(\State_{t+1}|\State_t, \Action_t)}
                \frac{\rewardd(\Reward_t|\State_t,\Action_t)}{\rewardd^\dagger(\Reward_t|\State_t, \Action_t)}
            }
        }
    }
\end{equation}
with equality when $\model^\dagger \ll \model$ as well. 
The quantity $L(\History_T; \model, \model^\dagger) := \sum_{t=1}^T \log\parens*{ \frac{\kerneld(\State_{t+1}|\State_t,\Action_t)}{\kerneld^\dagger(\State_{t+1}|\State_t, \Action_t)} \frac{\rewardd(\Reward_t|\State_t,\Action_t)}{\rewardd^\dagger(\Reward_t|\State_t, \Action_t)} }$ is known as the \strong{log-likelihood ratio} of observations. 

If the equation \eqref{equation_change_of_measure} can be used as is, like in \cite{lai_asymptotically_1985}, it is often transformed \cite{wald1945sequential, kaufmann2014analyse, maillard2019mathematics} to obtain an inequality that are more easy to work with. 
The key observation is that the expected log-likelihood ratio takes the simpler form
$
    \E_{\state_0}^{\model, \learner}\brackets{
        L(\History_T; \model, \model^\dagger)
    }
    = 
    \E_{\state_0}^{\model, \learner}\brackets{
        \sum_{\pair \in \pairs} \visits_T(\pair) \KL_\pair (\model||\model^\dagger)
    }
    .
$
In tandem with an application of Jensen's inequality in \eqref{equation_change_of_measure}, we obtain (see \cref{corollary_likelihood_jensen}), for $U > 0$, 
\begin{equation}
\label{equation_change_of_measure_log}
    \E_{\state_0}^{\model, \learner}\brackets*{
        \sum_{\pair \in \pairs} \visits_T(\pair) \KL_\pair (\model||\model^\dagger)
    }
    \ge
    \E_{\state_0}^{\model, \learner} \brackets*{\log(U)}
    - \log\parens*{\E_{\state_0}^{\model^\dagger, \learner} \brackets*{U}}
    .
\end{equation}
Accordingly, we are looking for a $U > 0$ such that $\E_{\state_0}^{\model, \learner} \brackets{\log(U)}$ is large while $\log\parens{\E_{\state_0}^{\model^\dagger, \learner} \brackets{U}}$ is small; Or, such that $\E_{\state_0}^{\model, \learner} \brackets{\log(U)}$ is small while $- \log (\E_{\state_0}^{\model^\dagger, \learner} [U])$ is large.
In other words, we are looking for a random variable that discriminate how the learner behaves on $\model$ and $\model^\dagger$.

\paragraph{A simple example of discriminating random variable.}
In a few cases where $\model^\dagger$ satisfies more properties than merely $\model^\dagger \in \alternative(\model)$, the design of a discriminating random variable may be greatly simplified, so that results \emph{à-la} \cref{proposition_rejecting_alternative} can be derived within a few lines from the change of measure inequality \eqref{equation_change_of_measure_log}. 
One such example is when $\weakoptimalpairs(\model^\dagger) \cap \weakoptimalpairs(\model) = \emptyset$, in which case the set of sub-optimal pairs of $\model^\dagger$ is a superset of $\weakoptimalpairs(\model)$.
Following this, letting $U := 1 + \sum_{\pair \in \weakoptimalpairs(\model)} \visits_{T+1}(\pair)$ and rewriting the expected regret as a sum of gaps with \cref{proposition_pseudo_regret}, we readily obtain:
\begin{equation*}
\begin{aligned}
    \Reg(T; \model, \learner, \state_0) 
    & \le
    \max(\ogaps(\model)) \E_{\state_0}^{\model, \learner} \brackets*{T - U} + \OH(1)
    \text{; and}
    \\
    \Reg(T; \model^\dagger, \learner, \state_0)
    & \ge 
    \dmin(\ogaps(\model^\dagger))
    \E_{\state_0}^{\model, \learner} \brackets{U}
    + \OH(1)
    .
\end{aligned}
\end{equation*}
What is important in the above, is to convert the consistency properties that control the expected regret to controlled inequalities on visit counts; The very value of $\max(\ogaps(\model))$ and $\dmin(\ogaps(\model^\dagger))$ have very little importance asymptotically. 
By consistency, it follows that $\E_{\state_0}^{\model, \learner}[T-U] = \oh(T^\eta)$ and $\E_{\state_0}^{\model^\dagger, \learner}[U] = \oh(T^\eta)$ where $\eta > 0$ is arbitrarily small.
So, by Markov's inequality, $\Pr_{\state_0}^{\model, \learner} \parens*{U \le T - T^\eta} = \oh(1)$ and $\E_{\state_0}^{\model, \learner}[\log(U)] \ge \log(T) + \oh(\log(T))$. 
Together with \eqref{equation_change_of_measure_log}, we obtain:
\begin{equation*}
\begin{aligned}
    \E_{\state_0}^{\model, \learner}\brackets*{
        \sum_{\pair \in \pairs} \visits_{T+1}(\pair) \KL_\pair (\model||\model^\dagger)
    }
    & \ge
    (1 - \eta) \log(T) + \oh(\log(T))
    .
\end{aligned}
\end{equation*}
As $\eta > 0$ can be chosen arbitrarily small, we conclude that \cref{proposition_rejecting_alternative} holds in particular for $\model^\dagger \in \models$ such that $\weakoptimalpairs(\model^\dagger) \cap \weakoptimalpairs(\model) = \emptyset$. 

\paragraph{A general discriminating random variable.}
For the general setting, where $\model^\dagger \in \alternative(\model)$ without any additional assumption, the random variable $U$ that we choose is provided by \cref{lemma_discriminating} below.
We track with $U$ how many times the algorithm plays sub-optimal pairs of $\model$ that are optimal on $\model^\dagger$. 
Intuitively, $\E^{\model, \learner}_{\state_0}[U]$ is roughly bounded by the regret $\Reg(T; \model, \learner)$ while $\E^{\model^\dagger, \learner}_{\state_0}[U]$ grows at the same rate as $T - \OH(\Reg(T; \model^\dagger, \learner))$. 
A fine control of the tails of $U$ is however required to properly spawn the $\log(T)$ factor that appears in \cref{proposition_rejecting_alternative}.
This is made possible by the key \cref{lemma_discriminating} below.

\begin{lemma}
\label{lemma_discriminating}
    Let $\model \in \models$, pick $\model^\dagger \in \alternative(\model)$
    and let $\pairs_c := \set{\pair \in \optpairs(\model^\dagger) : \ogaps(\pair; \model) > 0}$. 
    There exist constants $\epsilon^\dagger_c, \diameter^\dagger_c > 0$ such that, whatever the learning agent $\learner$ and the initial state $\state_0$, we have:
    \begin{equation*}
        \forall u \ge 0,
        \quad
        \Pr_{\state_0}^{\model^\dagger, \learner} \parens*{
            \sum_{x\in\pairs_c} N_{T+1}(\pair)
            +
            \diameter^\dagger_c \sum_{x \in \suboptimalpairs(\model)} \visits_{T+1}(\pair)
            \le
            \epsilon^\dagger_c T - u
        }
        \le 
        \exp \parens*{
            - \frac{2 u^2}{T \diameter^{\dagger 2}_c}
        }
        .
    \end{equation*}
\end{lemma}

The proof of \cref{lemma_discriminating} is provided in \cref{appendix_inevitable}.
We can now prove the proposition of interest. 

\begin{proof}[Proof of \cref{proposition_rejecting_alternative}]
    Let $\model \in \models$ and pick $\model^\dagger \in \alternative(\model)$. 
    Let $\pairs_c := \set{\pair \in \optpairs(\model^\dagger) : \ogaps(\pair; \model) > 0}$ and fix $U := \parens{1 + \sum_{\pair \in \pairs_c} \visits_{T+1}(\pair)}^{-1}$. 
    The goal is to invoke \eqref{equation_change_of_measure_log} for $U$. 

    Fix $\eta > 0$ arbitrary.

    We start by upper-bounding $\E_{\state_0}^{\model^\dagger, \learner}[U]$.
    Let $\epsilon_c^\dagger, \diameter_c^\dagger > 0$ the constants provided by \cref{lemma_discriminating}.
    By \cref{proposition_pseudo_regret}, we know that for $\alpha := \vecspan{\optbias(\model^\dagger)}$ and for all $T \ge 1$, we have:
    \begin{equation*}
        \alpha + \Reg(T; \model^\dagger, \learner, \state_0)
        \ge
        \E_{\state_0}^{\model^\dagger, \learner} \brackets*{
            \sum_{\pair \in \pairs} \visits_{T+1}(\pair) \ogaps(\pair; \model^\dagger)
        }
        \ge
        \dmin(\ogaps(\model^\dagger))
        ~\E_{\state_0}^{\model^\dagger, \learner} \brackets*{
            \sum_{\pair \in \pairs_-(\model^\dagger)} \visits_{T+1}(\pair) 
        }
        .
    \end{equation*}
    Observe that with this kind of reasoning, we also show that $\E_{\state_0}^{\model, \learner}[\sum_{\pair \in \pairs_c} \visits_{T+1}(\pair)] \le \frac{\Reg(T; \model, \learner, \state_0) + \vecspan{\optbias(\model)}}{\dmin(\ogaps(\model))} = \oh(T^\eta)$. 
    Now, combining \cref{lemma_discriminating} with Markov's inequality, we obtain:
    \begin{equation*}
        \Pr_{\state_0}^{\model^\dagger, \learner} \parens*{
            U \le \frac 12 \epsilon^\dagger_c T - u
        }
        \le \frac{
            2 \Reg(T; \model^\dagger, \learner, \state_0) T^{-1}
        }{\dmin(\ogaps(\model^\dagger)) \epsilon_c^\dagger}
        + \exp\parens*{ - \frac{2u}{T\diameter_c^{\dagger2}} }
        =
        \oh\parens*{T^{\eta - 1}}
        + \exp\parens*{ - \frac{2u}{T\diameter_c^{\dagger2}} }
    \end{equation*}
    where the equality follows by strong consistency of $\learner$ on $\model^\dagger$.
    So, we get:
    \begin{align*}
        \E_{\state_0}^{\model^\dagger, \learner}[U]
        \equiv
        \E_{\state_0}^{\model^\dagger, \learner}\brackets*{
            \frac1{1 + \sum_{\pair \in \pairs_c} \visits_{T+1}(\pair)}
        }
        & \le
        \inf_{v \ge 0} \braces*{
            \frac 1{1+v}
            +
            \Pr_{\state_0}^{\model^\dagger, \learner} \parens*{\sum_{\pair \in \pairs_c} \visits_{T+1}(\pair) \le v}
        }
        \\
        & \le
        \inf_{v \ge 0} \braces*{
            \frac 1{1+v}
            + \oh\parens*{T^{\eta - 1}}
            + \exp\parens*{ - \frac{2\parens*{\frac 12 \epsilon_c^\dagger T - v}^2}{T\diameter_c^{\dagger2}} }
        }
        \\ 
        & \overset{(\dagger)}\le
        \frac 1{1+\frac 14 \epsilon^\dagger_c T}
        + \oh\parens*{T^{\eta - 1}}
        + \exp \parens*{
            - \frac{\epsilon^\dagger_c T}{8\diameter^{\dagger 2}_c}
        }
    \end{align*}
    where $(\dagger)$ follows by setting $v := \frac 14 \epsilon_c^\dagger T$.
    We conclude that $\log(\E_{\state_0}^{\model^\dagger, \learner}[U]) \le (\eta - 1) \log(T) + \oh(\log(T))$. 

    We conclude as follows.
    By \eqref{equation_change_of_measure_log}, we have:
    \begin{equation*}
    \begin{aligned}
        \E_{\state_0}^{\model, \learner} \brackets*{
            \sum_{\pair \in \pairs} \visits_{T+1}(\pair) \KL_\pair(\model||\model^\dagger)
        }
        & \ge 
        \E_{\state_0}^{\model, \learner} \brackets*{\log(U)}
        - 
        \log \parens*{\E_{\state_0}^{\model^\dagger, \learner} \brackets*{U}}
        \\
        & =
        - \E_{\state_0}^{\model, \learner} \brackets*{\log\parens*{1 + \sum_{\pair \in \pairs_c} \visits_{T+1}(\pair))}}
        - 
        \log \parens*{
            \E_{\state_0}^{\model^\dagger, \learner} \brackets*{
                \frac 1{1 + \sum_{\pair \in \pairs_c} \visits_{T+1}(\pair)}
            }
        }
        \\
        & \overset{(\dagger)}\ge 
        - \log \parens*{
            1 + \E_{\state_0}^{\model, \learner} \brackets*{\sum_{\pair \in \pairs_c} \visits_{T+1}(\pair)}
        }
        + (1 - \eta) \log \parens*{ T }
        + \oh\parens*{\log(T)}
        \\
        & \overset{(\ddagger)}\ge
        (1 - 2 \eta) \log \parens*{T} + \oh \parens*{\log(T)}.
    \end{aligned}
    \end{equation*}
    where $(\dagger)$ is obtained by applying Jensen's inequality to bound the first term and the previous computation to bound the second
    and $(\ddagger)$ uses that $\E_{\state_0}^{\model, \learner} [\sum_{\pair \in \pairs_c} \visits_{T+1}(\pair)] = \oh(T^\eta)$. 
    As this holds for arbitrary $\eta > 0$, we conclude accordingly. 
\end{proof}

\subsection{Co-exploration: From alternative to confusing sets}
\label{section_visit_rates_of_optimal}

\cref{proposition_rejecting_alternative} provides our strongest result about how much information consistent learners must collect during play, by essentially showing that consistent learners have to reject every alternative model. 
In this paragraph, we couple this fact with an orthogonal result, stating that under a mild assumptions, every alternative model $\model^\dagger$ that differ from $\model$ on optimal pairs is trivially ``ruled out'' by a consistent learner. 
If, in fact, $\alternative(\model)$ was introduced in \cref{definition_alternative_model} to account for the fact that generic learners are not guaranteed to gather a lot of information of $\optpairs(\model)$, we can actually show that pairs of $\optpairs(\model)$ are all visited overlogarithmically by consistent learners. 
This behavior is to be distinguished from exploration (playing sub-optimal pairs to make sure that they are sub-optimal) because it underlines a need to play optimal (hence zero-cost) pairs in order to gather information on them rather than gathering rewards alone; This is referred to as \strong{co-exploration}.

\begin{proposition}[Co-exploration constraint]
\label{proposition_forced_optimal_pairs}
    Let $\model \in \models$ and fix $\pair_\opt \in \optpairs(\model)$. 
    Assume that, for all $\epsilon > 0$, there exists $\model' \in \models$ such that (1) $\model'$ and $\model$ only differ with $\rewardd(\pair_\opt) \ne \rewardd'(\pair_\opt)$, (2) $\reward(\pair_\opt) < \reward'(\pair_\opt)$ and (3) $\KL(\rewardd(\pair_\opt)||\rewardd'(\pair_\opt)) < \epsilon$.
    Then,\footnote{We recall that $f(T)=\omega(g(T))$ means that $f$ grows strictly faster than $g$ as $T\to\infty$ that is, $\liminf_{T\to\infty} |f(T)/g(T)|= \infty$.} for every consistent learner $\learner$ and regardless of the initial state $\state_0 \in \states$, 
    \begin{equation*}
        \E_{\state_0}^{\model, \learner}\brackets*{\visits_{T+1}(\pair_\opt)}
        =
        \omega \parens*{\log(T)}
        .
    \end{equation*}
\end{proposition}

The proof of \cref{proposition_forced_optimal_pairs} is deferred to \cref{appendix_inevitable} --- it again relies on change of measures, but in a quite different way than \cref{proposition_rejecting_alternative}.
In \cref{proposition_forced_optimal_pairs}, the three conditions (1-3) imply that there exists models that are arbitrarily close to $\model$ for which every gain optimal policy must use $\pair_{\opt}$ infinitely often. 
These conditions are for instance automatically satisfied when $\models \equiv \prod_{\pair \in \pairs} (\probabilities(\states) \otimes \probabilities([0,1]))$ is the set of all Markov decision processes with pair space $\pairs$ and $\model \in \models$ satisfies $\max(\reward) < 1$.
That scenario is quite similar to the one under which \cite{lai_asymptotically_1985} prove the tightness of their lower bound in the first place.
And in that scenario, the information constraints of \cref{proposition_rejecting_alternative} can be dropped to confusing models instead of alternative ones.
Indeed, if $\model^\dagger \in \alternative(\model)$ is such that there exists $\pair \in \optpairs(\model)$ with $\KL_\pair (\model||\model^\dagger) > 0$, then every consistent learner is such that:
\begin{equation*}
    \liminf_{T \to \infty}
    \frac 1{\log(T)}
    \E_{\state_0}^{\model, \learner} \brackets*{
        \sum_{\pair \in \pairs} 
        \visits_{T+1}(\pair) 
        \KL_\pair(\model||\model^\dagger)
    }
    = 
    \infty.
\end{equation*}
Therefore, the information constraint associated to $\model^\dagger$ in \cref{proposition_rejecting_alternative} is trivially met.  
Accordingly, the immediate consequence of \cref{proposition_rejecting_alternative}, \cref{corollary_rejecting_confusing} below, \strong{does not} lose any information relatively to \cref{proposition_rejecting_alternative} when completed with \cref{proposition_forced_optimal_pairs}.

\begin{corollary}
\label{corollary_rejecting_confusing}
    Let $\model \in \models$ and let $\model^\dagger \in \confusing(\model)$. 
    Then, every consistent learner $\learner$ satisfies:
    \begin{equation*}
        \E_{\state_0}^{\model, \learner} \brackets*{
            \sum_{\pair \in \pairs} 
            \visits_{T+1}(\pair)
            \KL_\pair \parens{\model||\model^\dagger}
        }
        \ge
        \log(T)
        + \oh\parens*{\log(T)}
        .
    \end{equation*}
\end{corollary}

\paragraph{Beyond optimal pairs.}
The alternative set can be provably simplified to the confusing set via \cref{proposition_forced_optimal_pairs}, because consistent learners are forced to gather much more information on $\optpairs(\model)$ than on $\pairs \setminus \optpairs(\model)$.
However, it is impossible to further simplify $\confusing(\model)$ with more results in the style of \cref{proposition_forced_optimal_pairs}.
For instance, $\confusing(\model)$ cannot be strengthened to
\begin{equation*}
    \wkconf(\model) := \set*{
        \model^\dagger \in \models
        :
        \model^\dagger \gg \model
        \text{~and~}
        \model^\dagger = \model \text{~on~} \weakoptimalpairs(\model)
        \text{~and~}
        \optpolicies(\model^\dagger) \cap \optpolicies(\model) = \emptyset
    }
\end{equation*}
in \cref{corollary_rejecting_confusing}.
The reason for this is that the amount of time spent outside of $\optpairs(\model)$ is proportional to the regret, underlying a key difference between $\optpairs(\model)$ and $\weakoptimalpairs(\model)$, see \cref{lemma_time_outside_optimal_pairs} below, that illustrates the transient nature of $\weakoptimalpairs(\model) \setminus \optpairs(\model)$.

\begin{lemma}
\label{lemma_time_outside_optimal_pairs}
    Let $\model \in \models$. 
    There exist constants $\alpha, \beta > 0$ such that, for every learner $\learner$ and $T \ge 1$,
    \begin{equation*}
        \E_{\state_0}^{\model, \learner}\brackets*{
            \sum_{t=1}^T \indicator{\Pair_t \notin \optpairs(\model)}
        }
        \le \alpha~\Reg(T; \model, \learner, \state_0) + \beta.
    \end{equation*}
\end{lemma}
\begin{proof}
    Consider the revised version $\model_f$ of $\model$ with revised reward vector $f(\pair) := \indicator{\pair \in \optpairs(\model)}$. 
    Let $\policy$ the policy such that $\policy(\state)$ is the uniform probability distribution over $\set{\action \in \actions(\state): (\state, \action) \in \weakoptimalpairs(\model)}$. 
    This policy is gain-optimal, and we denote $\gain_f, \bias_f$ and $\gaps_f$ the gain, bias and gap functions of $\policy$ under the reward function $f$. 
    We have:
    \begin{align*}
        \sum_{\pair \notin \optpairs(\model)}
        \visits_{T+1}(\pair)
        & = 
        T - \sum_{t=1}^{T}
        f(\Pair_t)
        \overset{(\dagger)}=
        T - \sum_{t=1}^T \parens*{\gain_f(\State_t) + \parens*{\unit_{\State_t} - \kernel(\Pair_t)} \bias_f - \gaps_f(\Pair_t)}
        \\
        & \overset{(\ddagger)}\le 
        \vecspan{\bias_f} + \norm{\gaps_f}_\infty \sum_{\pair \notin \weakoptimalpairs(\model)} \visits_{T+1}(\pair)
        \\
        & \le 
        \vecspan{\bias_f} + \frac{\norm{\gaps_f}_\infty}{\dmin(\ogaps(\model))} \sum_{\pair \notin \weakoptimalpairs(\model)} \visits_{T+1}(\pair) \ogaps(\pair;\model) 
    \end{align*}
    where $(\dagger)$ unfolds the definition of $\gaps_f$,
    $(\ddagger)$ uses the remark that since $\policy$ is gain optimal, it converges to $\optpairs(\model)$ so $\gain_f = 1$; and that we have $\gaps_f(\pair) = 0$ for $\pair \notin \weakoptimalpairs(\model)$.
    Take the expectation.
\end{proof}

If the learner cannot spend too much time outside of $\optpairs(\model)$ without degrading the regret, it means that it cannot gather a lot of information on any pair outside of $\optpairs(\model)$ without increasing the regret.
In that aspect, \cref{proposition_forced_optimal_pairs} and \cref{lemma_time_outside_optimal_pairs} are the two sides of the same coin: efficient consistent learners will gather enormous amount of information \emph{precisely} on all optimal pairs. 
In other words, $\optpairs(\model)$ and $\pairs \setminus \optpairs(\model)$ are treated differently by such learners. 

\paragraph{Regret bounds obtained from information constraints.} 
As a matter of fact, a regret lower bound can immediately be derived from the information constraints described by \cref{proposition_rejecting_alternative}. 
By \cref{proposition_pseudo_regret}, the expected regret can be written as $\sum_{\pair \in \pairs} \E_{\state_0}^{\model, \learner}[\visits_{T+1}(\pair)] \ogaps(\pair; \model) + \OH(1)$, while \cref{proposition_rejecting_alternative} states that $\sum_{\pair \in \pairs}\E_{\state_0}^{\model, \learner}[\visits_{T+1}(\pair)] \KL_\pair(\model||\model^\dagger) \ge \log(T) + \oh(\log(T))$ for every $\model^\dagger \in \alternative(\model)$. 
Normalizing by $\log(T)$ and going to the asymptotic, we see that a consistent learner satisfies:
\begin{equation*}
    \liminf_{T \to \infty}
    \frac{\Reg(T; \model, \learner, \state_0)}{\log(T)}
    \ge
    \inf \set*{
        \sum_{\pair \in \pairs}
        \imeasure(\pair) \ogaps(\pair; \model)
        :
        \imeasure \in [0, \infty]^\pairs
        \text{~and~}
        \inf_{\model^\dagger \in \alternative(\model)}
        \sum_{\pair \in \pairs}
        \imeasure(\pair) \KL_\pair (\model||\model^\dagger)
        \ge 
        1
    }
    .
\end{equation*}
What \cref{proposition_forced_optimal_pairs} states is that in the above and under mild assumptions, $\alternative(\model)$ can be changed to $\confusing(\model)$ without changing the lower bound.
The resulting lower bound is of the same kind as \cite{lai_asymptotically_1985,graves1997asymptotically,burnetas_optimal_1997,agrawal_asymptotically_1988} and can be seen as a generalization of their results to the more general setting of communicating Markov decision processes. 
It appears clearly that in such bounds however, the structure of $\visits_{T+1}/\log(T)$ is disregarded.
In the respective settings of these works, it is disregarded because the obtained lower bound is provably reached by a learning agent. 
This is only possible in their particular settings, and the lower bound that we provide downstream will provide new light into why, and when, the structure of $\visits_{T+1}/\log(T)$ can be ``ignored'', see \cref{section_example_bandits,section_example_recurrent}.

In general and as motivated in \cref{section_navigation_informal}, the structure of $\visits_{T+1}/\log(T)$ is far from arbitrary for communicating Markov decision processes, and has to be taken into account.

\subsection{Decomposition of the navigation of consistent learners and minors} 
\label{section_minors}

With \cref{proposition_rejecting_alternative}, consistent learning agents have been shown to wander their environment in order to implicitely collect enough information so that alternative models are rejected. 
This environment, in opposition to multi-armed bandits, usually has structure. 
The critical lack of information on a pair $\pair \in \pairs$ requires the agent to compute an efficient way to reach that pair, hence the visit vector $\visits_{T+1}$ is not just any element of $\N^\pairs$. 
This all has already been discussed in \cref{section_navigation_informal}, where we have pinpointed the importance of minors, that we shall formally define now. 

\subsubsection{Formal definition of minors}
\label{section_minors_formal_definition}

Minors generalize edge contraction on graphs to Markov decision processes and are close in spirit to the state reduction/aggregation of \cite{ortner2013adaptive}.
They are obtained by contracting subsets of pairs of the initial model/
In opposition to classical graph theory, we won't allow for the contraction of arbitrary subsets of pairs.
The contracted subset $\pairs'$ must be \strong{closed} (\cref{definition_closed}), meaning that (1) one remains in the states spawned by $\pairs'$ by playing pairs of $\pairs'$, and (2) it does not contain transient states. 
The property (1) implies that $\model$ can be restricted to $\pairs'$ and the states it spawns, and (2) that the obtained model is a union of disjoint communicating Markov decision processes.

\begin{definition}[Closed pair subspace of $\mdp$]
\label{definition_closed}
    A subset $\pairs_0 \subseteq \pairs$ with corresponding states $\states(\pairs_0)$ is a \strong{closed}  set of $\mdp$ if  it is (1) \strong{forward closed} meaning $\kerneld(-|\pair)$ is supported in $\states(\pairs_0)$ for all $\pair \in \pairs_0$, and (2) \strong{backward closed}, meaning the model $\mdp$ constrained to $\pairs_0$\footnote{It is well-defined by forward closeness (1).} is a union of communicating components (no transient states).
\end{definition}

    A simple, yet illuminating observation, is that a subset $\pairs_0$ is closed if, and only if it is the set of recurrent pairs under some randomized stationary policy.
    In this paper, the most important example of closed set is the set of optimal pairs $\optpairs(\model)$, which is obtained as the recurrent pairs of the policy $\policy^\opt$ given by $\policy^\opt(\state)$ as the uniform distribution on $\set{\action \in \actions(\state): (\state, \action) \in \weakoptimalpairs(\model)}$.

\begin{definition}[Minors/Contractions]
\label{definition_minors}
    Up to re-labeling actions, assume that $\actions(\state) \cap \actions(\state') = \emptyset$ for $\state \ne \state'$.
    Let $\model \in \models$ a model and fix $\pairs_0 \subseteq \pairs$ a closed set of $\model$. 
    The \strong{contraction of $\model$ by $\pairs_0$} is the model $\model/\pairs_0$ obtained by merging every communicating component of $\pairs_0$ into single states.
    More formally, letting $\states_1,\ldots,\states_k$ the communicating components of $\pairs_0$, we have:
    \begin{enum}
        \item
            The state space is $\states(\model/\pairs_0) := \set[\big]{\states_1, \ldots, \states_k} \cup \set[\big]{\set*{\state}: \state \in \states \text{~and~}\forall i, \state \notin \states_i}$ and contracted states are denoted $[\state]$;
        \item
            The action space is, for $[\state] \in \states(\model/\pairs_0)$, $\actions(\model/\pairs_0)[s] := \bigcup_{\state' \in [\state]} \actions(s')$; Because in $\model$, the choice of an action uniquely determines a state, the state-action space $\pairs(\model/\pairs_0)$ is canonically isomorphic to $\pairs(\model)$, by associating $([\pair], \action)$ to $(\state', \action)$ where $\state'$ is the unique state such that $\action \in \actions(\state')$.
        \item 
            The kernel is $[\kerneld]([\state_1]|[\state_0], \action) := \sum_{\state'_1 \in [\state_1]} \kerneld(\state'_1|\state'_0, \action)$;
        \item 
            The reward is $[\rewardd]([\state_0], \action) := \rewardd(\state'_0, \action)$.
    \end{enum}
    We also say that $\model/\pairs_0$ is a \strong{minor} of $\models$.
\end{definition}

\subsubsection{Decomposing executions using invariant measures of minors}

Minors provide a descriptive decomposition of how a Markov decision process can be explored, following a general principle. 
The collection of optimal pairs $\optpairs$ is an example of closed set (\cref{definition_closed}), that themselves are recurrent pairs of randomized policies. 
Independently of the way a planner explores the model, the normalized ratio of visits outside of a closed set converges to an invariant measure of the minor induced by that closed set, provided that the outside is visited at least logarithmically often. 

\begin{proposition}
\label{proposition_expected_minor}
    Let $\policy$ a randomized policy and let $\pairs_\policy$ its recurrent pairs.
    Let $\alg$ any learning agent such that $\E_{\state_0}^{\model, \alg}[\sum_{\pair \notin \pairs_\policy} N_T (\pair)] = \Omega(\log(T))$.
    The vector given by
    \begin{equation*}
        \mu_t (\pair) 
        :=
        \frac{
            \E^{\model, \learner}_{\state_0}[\visits_t(\pair)] \indicator{\pair \notin \pairs_\policy}
        }{
            \E^{\model, \learner}_{\state_0}\brackets*{\sum_{\pair' \notin \pairs_\policy} N_t(\pair')}
        }
    \end{equation*}
    converges to the space of invariant measures of $\model / \pairs_\policy$, i.e., every limit point of $(\imeasure_t)$ is an invariant measure of $\model / \pairs_\policy$.
\end{proposition}

This result is coupled with a second observation.
It can be shown that if the expected visits counts can be written as $\E^{\alg}[\visits_T(\pair)] = \alpha(\pair) T + \oh(T)$, then $\pairs_{\alg} := \set{\pair : \alpha(\pair) > 0}$ is a closed set that, if the planner is consistent, is a subset of $\optpairs(\model)$.
\cref{proposition_expected_minor} supports the previously motivated idea that the ``sublinear'' part of visit counts is easier to understand after contracting the model by $\optpairs(\model)$: Up to normalization, the truncation of the visit vector to $\pairs \setminus \optpairs(\model)$, i.e., $\visits_T \indicator{\optpairs(\model)^c}$, of a consistent learning agent converges in expectation to a probability invariant measure of $\model / \optpairs(\model)$.

Specifically, \cref{proposition_expected_minor} will be used in the following form.

\begin{corollary}[Navigation constraints]
\label{corollary_navigation_constraints}
    Consider a consistent algorithm $\alg$ and let $\model \in \mathcal{\model}$ such that $\confusing(\model) \ne \emptyset$. 
    Then the vector given by
    \begin{equation*}
        \mu_T(\pair) := 
        \frac{
            \E_{\state_0}^{\model,\alg}[\visits_T(\pair)] \indicator{\pair \notin \optpairs(\model)}
        }{
                \E_{\state_0}^{\model,\alg}\brackets*{\sum_{\pair' \notin \optpairs(\model)} N_T(\pair')}
        }
    \end{equation*}
    converges to $\imeasures(\model/\optpairs(\model)) \cap \mathcal{P}(\pairs)$ when $T \to \infty$.
\end{corollary}
\begin{proof}[Proof of \cref{corollary_navigation_constraints}]
    Since $\confusing(\model) \ne \emptyset$, there exists $\model^\dagger \in \confusing(\model)$ and $\max_{\pair \in \pairs} \KL_\pair(\model||\model^\dagger) < \infty$.
    By \cref{corollary_rejecting_confusing}, we have 
    \begin{equation*}
        \log(T)
        \lesssim
        \E_{\state_0}^{\model, \learner} \brackets*{
            \sum_{\pair \in \pairs}
            \visits_{T+1}(\pair) \KL_{\pair}(\model||\model^\dagger)
        }
        \le 
        \max_{\pair \in \pairs} \set*{\KL_{\pair}(\model||\model^\dagger)}
        \E_{\state_0}^{\model, \learner} \brackets*{
            \sum_{\pair \notin \optpairs(\model)}
            \visits_{T+1}(\pair)
        }
    \end{equation*}
    where the second inequality follows from $\model^\dagger = \model$ on $\optpairs(\model)$.
    So $\E_{\state_0}^{\model, \learner} \brackets{ \sum_{\pair \notin \optpairs(\model)} \visits_{T}(\pair) } = \Omega(\log(T))$. 
    Moreover, remark that $\optpairs(\model)$ is closed as the set of recurrent pairs of the policy $\policy^*$ such that $\policy^*(-|s)$ is uniform on $\set{a: (s,a) \in \optpairs(\model)}$ if $s \in \mathcal{S}(\optpairs(\model))$ and uniform on $\mathcal{A}(s)$ otherwise.
    Accordingly, we can apply \cref{proposition_expected_minor}, to see that 
    \begin{equation*}
        \mu'_T (\pair) 
        :=
        \frac{
            \E^{\model, \learner}_{\state_0}[\visits_T(\pair)] \indicator{\pair \notin \optpairs(\model)}
        }{
            \E^{\model, \learner}_{\state_0}\brackets*{\sum_{\pair' \notin \optpairs(\model)} N_T(\pair')}
        }
    \end{equation*}
    converges to $\imeasures(\model/\optpairs(\model))$ as $T \to \infty$. 
    The fact that it is a probability vector is obvious.
\end{proof}

We now move on to the proof of \cref{proposition_expected_minor}.

\begin{proof}[Proof of \cref{proposition_expected_minor}]
    For conciseness, we write $\E[-]$ in place of $\E_{\state_0}^{\model, \learner}[-]$.
    Let $f(t) := \E[\sum_{\pair' \notin \pairs_\policy} N_t(\pair')]$ for short.
    By definition, the states of $[\model] := \model/\pairs_\policy$ are subsets of states of the original model $\model$ and its pair-space is canonically identical to the one of $\model$.
    Introduce the $\pairs_\policy^c$-truncated visit counts:
    \begin{equation*}
        N'_t (\pair) := \indicator{\pair \notin \pairs_\policy} N_t(\pair)
    \end{equation*}
    with the induced state-wise visits $N'_t(s) := \sum_a N'_t(s,a)$, and in the minor $N'_t[s] := \sum_{s' \in [s]} N'_t(s')$.
    By definition, we have $\mu_t(s,a) \equiv \E[N_t'(s, a)] f(t)^{-1}$.
    Let $\mu_t(s) := \sum_{a \in \mathcal{A}(s)} \mu_t(s,a) = \E[N'_t(s)]$ and, for $[s] \in \mathcal{S}[\model]$ a state of the contraction, $\mu_t[s] := \sum_{s' \in \mathcal{S}} \mu_t(s') = \E[N'_t[s]]$.
    Viewing $N'_t$ as a vector indexed by states of $[\model]$, a remarkable property of $N'_t$ is that it satisfies the quasi-flow property (what goes in is what goes out):
    \begin{equation}
    \label{equation:quasi-flow}
        N'_t[s]
        =
        \sum_{s' \in [s]} \sum_\pair N'_{t}(\pair; s')
        + \indicator{S_0 \in [s]}
        - \indicator{\Pair_t \in \pairs_\policy, S_t \in [s]}
        =
        \sum_{s' \in [s]} \sum_{a \in \mathcal{A}(s')} N'_t(s', a)
    \end{equation}
    where $N'_{t+1}(\pair; s') := \indicator{\pair \notin \pairs_\policy} N_{t+1}(\pair; s')$.
    This is established by induction on $t \ge 0$. 
    This is obvious for $t = 0$, and for $t \ge 1$, we have:
    \begin{align*}
        (-)
        & :=
        \sum_{s' \in [s]} \sum_{a \in \mathcal{A}(s')} N'_{t}(s',a)
        \\
        & \equiv N'_t[s]
        = 
        N'_{t-1}[s] 
        + \indicator{S_t \in [s], X_{t} \notin \pairs_\policy}
        \\
        & \overset{(*)}=
        \indicator{S_0 \in [s]}
        - \indicator{X_{t-1} \in \pairs_\policy, S_{t-1} \in [s]}
        + \sum_{s' \in [s]} \sum_\pair N'_{t-1}(\pair; s')
        + \indicator{S_t \in [s], X_{t} \notin \pairs_\policy}
        \\
        & = 
        \indicator{S_0 \in [s]}
        + \sum_{s' \in [s]} \sum_\pair N'_{t}(\pair; s')
        \\ & \phantom{{}={}}
        - \indicator{X_{t-1} \in \pairs_\policy, S_{t-1} \in [s]}
        + \indicator{S_t \in [s], X_{t} \notin \pairs_\policy}
        - \indicator{S_t \in [s], X_{t-1} \notin \pairs_\policy}
    \end{align*}
    where $(*)$ is obtained by induction.
    We focus on the RHS:
    \begin{align*}
        \alpha 
        & =
        - \indicator{X_{t-1} \in \pairs_\policy, S_{t-1} \in [s]}
        + \indicator{S_t \in [s], X_{t} \notin \pairs_\policy}
        - \indicator{S_t \in [s], X_{t-1} \notin \pairs_\policy}
        \\
        & \overset{(\dagger)}= 
        - \indicator{\Pair_t \in \pairs_\policy, S_t \in [s]}
        .
    \end{align*}
    The equality $(\dagger)$ is shown by distinguishing cases.
    \begin{itemize}
        \item If $X_{t-1} \in \pairs_\policy$ and $\Pair_t \in \pairs_\policy$, then because the states of $\pairs_\policy$ are closed by playing pairs of $\pairs_\policy$, it follows that $[S_t] = [S_{t-1}]$ correspond to the same recurrent class of $\policy$.
            If $[S_t] = [S_{t-1}] \ne [s]$, then we get $(\dagger): 0 = 0$ and if $[S_t] = [S_{t-1}] = [s]$, then $(\dagger): -1 = -1$.
        
        \item If $X_{t-1} \notin \pairs_\policy$ and $\Pair_t \notin \pairs_\policy$, then $(\dagger): 0 = 0$.

        \item If $X_{t-1} \in \pairs_\policy$ and $\Pair_t \notin \pairs_\policy$, then similarly $[S_{t-1}] = [S_t]$.
            If equal to $[s]$ then $(\dagger): 0 = 0$ and otherwise $(\dagger): 0 = 0$.

        \item If $X_{t-1} \notin \pairs_\policy$ and $\Pair_t \in \pairs_\policy$, then $[S_t]$ and $[S_{t-1}]$ can be equal or different. 
            If (1) $[S_t] = [S_{t-1}] = [s]$, we have $(\dagger): -1 = -1$; 
            If (2) $[S_t] = [S_{t-1}] \ne [s]$, we have $(\dagger): 0 = 0$;
            If (3) $[S_t] = [s] \ne [S_{t-1}]$, we have $(\dagger): -1 = -1$;
            And if (4) $[S_t] \ne [s] = [S_{t-1}]$, we have $(\dagger): 0 = 0$.
    \end{itemize}
    So $(\dagger)$ is established the quasi-flow property follows immediately. 

    Introduce $\policy'_t$ the policy of $[\model]$ as any state-wise probability distribution with $\E[N'_t[s]] \policy'_t(a|[s]) = \E[N'_t([s], a)]$, which is uniquely defined when $\E[N'_t[s]] > 0$.
    For all $[s'] \in \mathcal{S}[\model]$, we have:
    \begin{align*}
        \sum_{\pair \equiv ([s], a) \in \pairs[M]}
        \mu_t[s] \policy'_t(x|[s]) p([s']|x)
        & =
        \sum_{\pair \equiv ([s], a) \in \pairs[M]}
        \frac{\E[N'_t[s]] \policy'_t(x|[s])}{f(t)} p([s']|x)
        \\
        & =
        \sum_{\pair \notin \pairs_\policy} 
        \frac{\E[N_t(\pair)]}{f(t)} p([s']|x)
        \\
        & = 
        \sum_{\pair \notin \pairs_\policy} \sum_{s'' \in [s']}
        \frac{\E[N_t(\pair)]}{f(t)} p(s''|x)
        \\
        & = 
        \sum_{\pair \notin \pairs_\policy} \sum_{s'' \in [s']}
        \frac{
            \E\brackets*{N_t(\pair)\parens*{p(s''|x) - \hat{p}_t(s''|x)}}
            +
            \E\brackets*{N_t(\pair) \hat{p}_t(s''|x)}
        }{f(t)}
        \\
        \text{(\cref{lemma_self_normalized_expected_deviations})} & =
        \sum_{\pair \notin \pairs_\policy} \sum_{s'' \in [s']} 
        \parens*{
            \frac{\E[N_{t+1}(\pair; s'))]}{f(t)}
            + \oh \parens*{1+\frac{\E[N_t(\pair)]}{f(t)}}
        }
        \\
        & = 
        \sum_{\pair \notin \pairs_\policy} \sum_{s'' \in [s']}
        \parens*{
            \frac{\E[N_{t+1}(\pair;s'')]}{f(t)}
            + 
            \oh(1)
        }
        \\
        & \equiv
        \sum_{\pair \notin \pairs_\policy} \sum_{s'' \in [s']}
        \parens*{
            \frac{\E[N'_{t+1}(\pair;s'')]}{f(t)}
            + 
            \oh(1)
        }
        \\
        \text{(quasi-flow property \eqref{equation:quasi-flow})} & =
        \sum_{s'' \in [s']} 
        \parens*{
            \frac{\E[N'_{t+1}(s'')]}{f(t)}
            +
            \oh(1)
        }
        = \mu[s'] + \oh(1).
    \end{align*}
    We have shown that for all $[\state] \in \states[\model]$, $\sum_\pair \imeasure_t(\pair) \kernel([\state]|\pair) = \imeasure_t[\state] + \oh(1)$. 
    Written differently, introducting $\phi(\imeasure) := \max_{[\state] \in \states[\model]} \abs{\imeasure[\state] - \sum_\pair \imeasure(\pair) \kernel([\state]|\pair)}$, we have shown that $\phi(\imeasure_t) \to 0$.
    The function $\varphi$ is continuous, $\probabilities(\pairs)$ is compact and $\varphi(\imeasure) = 0$ is equivalent to $\imeasure \in \imeasures(\model/\pairs_\pi)$.
    Because $\imeasures(\model/\pairs_\policy)$ is closed in $\probabilities(\pairs)$, we deduce that $\varphi(\imeasure_t) \to 0$ implies that $\imeasure_t$ converges to $\imeasures(\model/\pairs_\policy)$.

    This concludes the proof.
\end{proof}

\subsection{A regret lower bound with contracted invariant measures}
\label{section_lower_bound_contracted}

We start by proving the following version of the lower bound.
In the next section, we will show that it is equivalent to \cref{theorem_lower_bound}.

\begin{theorem}
    \label{theorem_intermediate_lower_bound}
    Let $\mdp \in \models$.
    Regardless of the initial state $\state_0 \in \states$, the regret of every consistent algorithm $\alg$ satisfies $\liminf_{T \to \infty} \Reg(T; \alg, \model, \state_0) \log^{-1}(T) \ge \regretlb(\mdp)$ where $\regretlb(\mdp) \in [0, \infty]$ is equal to:
    \begin{equation}
    \label{equation_lower_bound_minor}
    \tag{LB${}'_{\jopt}$}
        \inf \set*{
            \sum_{\pair \in \pairs} \imeasure(\pair) \ogaps(\pair)
            :
            \imeasure \in \imeasures(\mdp/\optpairs(\model))
            \text{~and~}
            \inf_{\model^\dagger \in \confusing(\model)}
            \braces*{
                \sum_{\pair \in \pairs} \imeasure(\pair) \KL_\pair (\mdp||\mdp^\dagger) 
            }
            \ge 1
        }
    \end{equation}
\end{theorem}

The only difference with \cref{theorem_lower_bound} is that the condition ``$\imeasure \in \imeasures(\model)$'' of \cref{theorem_lower_bound} is changed to the less restrictive ``$\imeasure \in \imeasures(\mdp/\optpairs(\model))$'' in \cref{theorem_intermediate_lower_bound}.

\begin{proof}[Proof of \cref{theorem_intermediate_lower_bound}]
    Consider a consistent algorithm $\learner$ and pick $\model \in \mathcal{M}$.
    Introduce the quantities:
    \begin{equation*}
        \imeasure_{T}(\pair) := \frac{
            \indicator{\pair \notin \optpairs(\model)} \E_{\state_0}^{\model,\alg}[\visits_{T+1}(\pair)]
        }{
            \E_{\state_0}^{\model,\alg}\brackets*{\sum_{\pair'\notin\optpairs(\model)} \visits_{T+1}(\pair')}
        }
        \quad\text{and}\quad
        \lambda_T := \frac{
            \E_{\state_0}^{\model, \learner}\brackets*{
                \sum_{\pair \notin \optpairs(\model)} 
                \visits_{T+1}(\pair)
            }
        }{\log(T)}
    \end{equation*}
    and further introduce $\psi(T) := \Reg(T; \model, \learner, \state_0) \log(T)^{-1}$.
    Note that $\psi(T) = \lambda_T \sum_{\pair \notin \optpairs(\model)} \imeasure_T(\pair) \ogaps(\pair; \model)$. 

    Let $L := \liminf_{T \to \infty} \psi(T)$ and let $(T_n)$ a sequence with $\lim \psi(T_n) = L$. 
    Remark that $\limsup \lambda_{T_n} \le \alpha L$ as a consequence of \cref{lemma_time_outside_optimal_pairs}, where $\alpha > 0$ is the constant provided by \cref{lemma_time_outside_optimal_pairs}.
    Up to extracting a subsequence of $(T_n)$, we can assume that $\lambda_{T_n}$ converges to some $\Lambda \in [0, \alpha L]$. 
    Similarly, since $(\imeasure_T)$ evolves within the compact space $[0, 1]^\pairs$, we can assume that $\imeasure_{T_n}$ converges to some $\imeasure \in [0, 1]^\pairs$ up to extracting a subsequence of $(T_n)$. 
    By \cref{corollary_navigation_constraints}, every limit point of $(\imeasure_T)$ belongs to $\imeasures(\model/\optpairs(\model))$ which is closed, so $\imeasure \in \imeasures(\model/\optpairs(\model))$. 
    By \cref{proposition_rejecting_alternative}, we further have
    \begin{equation}
    \label{equation_theorem_lower_bound_proof_1}
        \forall T \ge 1,
        \quad
        \inf_{\model^\dagger \in \confusing(\model)}
        \braces*{
            \lambda_T
            \sum_{\pair \in \pairs} 
            \imeasure_T(\pair)
            \KL_\pair(\model||\model^\dagger)
        }
        \ge
        1
    \end{equation}
    so the limit $\Lambda \imeasure \in \imeasures(\model/\optpairs(\model))$ must satisfy \eqref{equation_theorem_lower_bound_proof_1} as well. 
    In tandem with $L = \Lambda \sum_{\pair \in \pairs} \imeasure(\pair) \ogaps(\pair; \model)$, we conclude that $L \ge \regretlb(\model)$.
\end{proof}

\begin{remark}
    The quantity $\regretlb(\model)$ actually depends on $\models$. 
    In general, $\models$ is obvious in the context but whenever it is not, we write $\regretlb(\model; \models)$ to make the dependency clear. 
\end{remark}

\subsection{A regret lower bound without contracted invariant measures}
\label{section_lower_bound_without_contracted}

To finally prove \cref{theorem_lower_bound}, we show that the contraction $\imeasure \in \imeasures(\model/\optpairs(\model))$ can be dropped to $\imeasure \in \imeasures(\model)$ in the lower bound. 
While minors play a central role in the decomposition of the execution consistent learning agents, they are eventually dropped in the final result, hence slightly simplifying the lower bound.
More precisely, by using that $\ogaps(\pair)$ is null for $\pair \in \optpairs(\model)$, invariant measures of the minor $\model/\optpairs(\model)$ can be \strong{represented} by invariant measures of the initial model $\model$, leading to:

\begin{proposition}[Removing the contraction]
\label{proposition_removing_contraction}
    The regret lower bound $\regretlb(\model)$ is equal to $\regretlb_{\jopt}(\model)$:
    \begin{equation}
    \label{equation_lower_bound_simplification}
        \regretlb(\model) =
        \regretlb_{\jopt}(\model) =
        \inf \set*{
            \sum_{\pair \in \pairs} \imeasure(\pair) \ogaps(\pair)
            :
            \imeasure \in \imeasures(\model)
            \text{~and~}
            \inf_{\model^\dagger \in \confusing(\model)}
            \braces*{\sum_{\pair \in \pairs} \imeasure(\pair) \KL_\pair (\mdp||\mdp^\dagger)}
            \ge 1
        }
        .
    \end{equation}
\end{proposition}

This is a direct consequence of the following remarkable result.

\begin{lemma}
\label{lemma_removing_contraction}
    Denote $u|_{\pairs \setminus \optpairs(\model)}$ the truncation of $u \in \R^\pairs$ to pairs of $\pairs \setminus \optpairs(\model)$ and extend the notations to subsets of $\R^\pairs$. 
    Then $\imeasures(\model)|_{\pairs \setminus \optpairs(\model)} = \imeasures(\model/\optpairs(\model))|_{\pairs \setminus \optpairs(\model)}$.
\end{lemma}
\begin{proof}
    Since $\imeasures(\model/\optpairs(\model)) \supseteq \imeasures(\model)$, one inclusion is obvious and we focus on the other.
    Up to iterating the process on the communicating components of $\optpairs(\model)$, we assume that $\pairs_0 := \optpairs(\model)$ induces a communicating model $\model|_{\pairs_0}$.
    Pick $[\imeasure] \in \imeasures(\model/\pairs_0)$. 
    We show that there $\imeasure$ such that $[\imeasure](\pair) = \imeasure(\pair)$ for $\pair \notin \pairs_0$.
    Denote $\states_0 := \set*{\state: \exists \action, (\state,\action) \in \pairs_0}$ the states in which $\pairs_0$ is rooted.
    In $[\model] := \model/\pairs_0$, we have $[\kernel]([\states_0]|\pair_0) = 1$ for all $\pair_0 \in \pairs_0$ so that we can assume that $[\imeasure](\pair_0) = 0$ without loss of generality.
    For $\state_0 \in \states_0$, introduce 
    \begin{equation}
        \alpha(\state) := \sum_{\action \in \actions(\state_0)} [\imeasure](\state_0,\action) - \sum_{\pair \in \pairs} [\imeasure](\pair) \kernel(\state_0|\pair)
        .
    \end{equation}
    Observe that $\sum_{\state_0 \in \states_0} \alpha(\state_0) = 0$ since $[\imeasure]$ is an invariant measure of $[\model] \equiv \model/\pairs_0$.
    It is enough to find $\imeasure_0 \in \R^{\pairs_0}$ such that
    \begin{equation}
    \label{equation_removing_contraction_2}
        \imeasure_0 \ge 0
        \text{\quad and \quad}
        \forall \state_0 \in \states_0,
        ~
        \sum_{\pair_0 \in \pairs_0} \imeasure_0(\pair_0) \kernel(\state_0|\pair_0) = \sum_{\action_0: (\state_0, \action_0) \in \pairs_0} \imeasure(\state_0, \action_0) + \alpha(\state_0);
    \end{equation}
    then $\imeasure \in \R^\pairs$ given by $\imeasure(\pair) = \imeasure_0(\pair)$ if $\pair \in \pairs_0$ and $[\imeasure](\pair)$ if $\pair \notin \pairs_0$ will be solution.
    Assume that \eqref{equation_removing_contraction_2} has no solution. 
    By Farkas' Lemma, there exists $\nu_0 \in \R^{\states_0}$ such that:
    \begin{equation}
    \label{equation_removing_contraction_3}
        \sum_{\state_0 \in \states_0}
        \nu_0(\state_0) \alpha(\state_0) 
        < 0
        \quad\text{and}\quad
        \forall (\state_0,\action_0) \in \pairs_0, ~
        \sum_{\state'_0 \in \states_0}
        \nu_0(\state_0) \parens*{
            \kernel(\state'_0|\state_0,\action_0)
            -
            \indicator{\state'_0 = \state_0}
        }
        \ge
        0
        .
    \end{equation}
    Let $\policy_0$ the policy picking its actions uniformly in $\pairs_0$ from $\states_0$.
    The second condition can be rewritten as $\kernel_{\policy_0}(\model|_{\pairs_0}) \nu_0 \ge \nu_0$ so by induction, $\akernel_{\policy_0}(\model|_{\pairs_0}) \nu_0 \ge \nu_0$ where $\akernel_{\policy_0} := \lim \frac 1T \sum_{t=1}^T \kernel_{\policy_0}^{t-1}$.
    Because $\pairs_0$ is communicating, $\akernel_{\policy_0}(\model|_{\pairs_0})$ has full support and $\nu_0 \in \R  e$. 
    But $\sum_{\state_0 \in \states_0} \alpha(\state_0) = 0$ so $\sum_{\state_0 \in \states_0} \nu_0(\state_0) \alpha(\state_0) = 0$, contradicting \eqref{equation_removing_contraction_3}.
\end{proof}

Combining \cref{proposition_removing_contraction} and \cref{theorem_intermediate_lower_bound}, \cref{theorem_lower_bound} follows immediately.

\subsection{Why navigation constraints are mandatory}
\label{section_mandatory_navigation_constraints}

To conclude this section, we discuss a simple example that illustrates the importance of navigation constraints. 

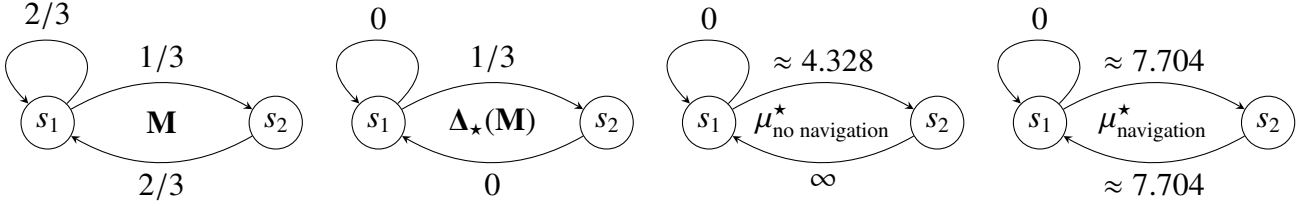
\begin{figure}[h]
    \centering
    \begin{tikzpicture}
        \clip (-.6, -1.5) rectangle (3.75, 2);
        \node at (1.5, 0) {$\model$};

        \node[draw, circle, inner sep=0cm, minimum size=0.7cm] (s1) at (0, 0) {$s_1$};
        \node[draw, circle, inner sep=0cm, minimum size=0.7cm] (s2) at (3, 0) {$s_2$};

        \draw[->, >=stealth] (s1) to[bend left] node[midway, above] {$1/3$} (s2);
        \draw[->, >=stealth] (s2) to[bend left] node[midway, below] {$2/3$} (s1);
        \draw[->, >=stealth] (s1) to[loop] node[midway, above] {$2/3$} (s1);
    \end{tikzpicture}%
    \begin{tikzpicture}
        \clip (-.6, -1.5) rectangle (3.75, 2);
        \node at (1.5, 0) {$\ogaps(\model)$};
        
        \node[draw, circle, inner sep=0cm, minimum size=0.7cm] (s1) at (0, 0) {$s_1$};
        \node[draw, circle, inner sep=0cm, minimum size=0.7cm] (s2) at (3, 0) {$s_2$};

        \draw[->, >=stealth] (s1) to[bend left] node[midway, above] {$1/3$} (s2);
        \draw[->, >=stealth] (s2) to[bend left] node[midway, below] {$0$} (s1);
        \draw[->, >=stealth] (s1) to[loop] node[midway, above] {$0$} (s1);
    \end{tikzpicture}%
    \begin{tikzpicture}
        \clip (-.6, -1.5) rectangle (3.75, 2);
        \node at (1.5, 0) {$\imeasure^\opt_\text{no navigation}$};

        \node[draw, circle, inner sep=0cm, minimum size=0.7cm] (s1) at (0, 0) {$s_1$};
        \node[draw, circle, inner sep=0cm, minimum size=0.7cm] (s2) at (3, 0) {$s_2$};

        \draw[->, >=stealth] (s1) to[bend left] node[midway, above] {$\approx 4.328$} (s2);
        \draw[->, >=stealth] (s2) to[bend left] node[midway, below] {$\infty$} (s1);
        \draw[->, >=stealth] (s1) to[loop] node[midway, above] {$0$} (s1);
    \end{tikzpicture}%
    \begin{tikzpicture}
        \clip (-.6, -1.5) rectangle (3.75, 2);
        \node at (1.5, 0) {$\imeasure^\opt_\text{navigation}$};

        \node[draw, circle, inner sep=0cm, minimum size=0.7cm] (s1) at (0, 0) {$s_1$};
        \node[draw, circle, inner sep=0cm, minimum size=0.7cm] (s2) at (3, 0) {$s_2$};

        \draw[->, >=stealth] (s1) to[bend left] node[midway, above] {$\approx 7.704$} (s2);
        \draw[->, >=stealth] (s2) to[bend left] node[midway, below] {$\approx 7.704$} (s1);
        \draw[->, >=stealth] (s1) to[loop] node[midway, above] {$0$} (s1);
    \end{tikzpicture}%
    \caption{
        \label{figure_example_navigation_importance}
        An example of a two-state deterministic Markov decision process with Bernoulli rewards where the navigation constraints are necessary to obtain the right regret lower bound. 
    }
\end{figure}

On \cref{figure_example_navigation_importance}, we present a two-state Markov decision process $\model$ with deterministic transitions and Bernoulli rewards. 
$\optpairs(\model)$ consists in the loop from $s_1$ to itself, scoring $\frac 23$ in average. 
The other circle, between $\state_1$ and $\state_2$, only scores $\frac 12$ in average. 
With navigation constraints, we see that every $\imeasure \in \imeasures(\model)$ satisfies $\imeasure(\state_1 \to \state_2) = \imeasure(\state_2 \to \state_1)$; We can numerically evaluate $\regretlb(\model)$ and obtain $\imeasure^\opt(\state_1 \to \state_2) \approx 7.704$ and $\regretlb(\model) \approx 2.6$. 
Meanwhile, if one ignores navigation constraints (refer to the last paragraph of \cref{section_visit_rates_of_optimal}), then the transition $\state_2 \to \state_1$ can be explored arbitrarily more than $\state_1 \to \state_2$, while it is free to pick because it scores $\frac 23 = \optgain$.
Therefore, one puts infinitely much mass on $\state_2 \to \state_1$ and has less information to gather on $\state_1 \to \state_2$, leading to the navigation-less lower bound of $\approx 1.4$, which is significatively far from $\regretlb(\model)$.

    \section{Computational intractability of the regret lower bound}\label{sec:hardness}
    \allowdisplaybreaks
\label{section_complexity}

In this section, we discuss the computational difficulties related to the regret lower bound $\regretlb_{\jopt}(\model; \models)$ of \cref{theorem_lower_bound}, given by:
\begin{equation}
    \nonumber
    \inf \set*{
        \sum_{\pair \in \pairs} \imeasure(\pair) \ogaps(\pair)
        :
        \imeasure \in \imeasures(\model)
        \text{~and~}
        \forall \model^\dagger \in \conf(\model),
        \sum_{\pair \in \pairs} \imeasure(\pair) \KL_\pair (\mdp||\mdp^\dagger) \ge 1
    }
    .
\end{equation}
While the objective function is linear, the unknown $\imeasure$ must satisfy a continuum of constraints that are non-convex in general. 
Behind the computation of $\regretlb_{\jopt}(\model; \models)$ hides a question that is even simpler:
Given an exploration measure $\imeasure \in \imeasures(\model)$, is it informative enough?
Does it satisfy the information constraint $\inf_{\model^\dagger \in \confusing(\model)} \sum_{\pair \in \pairs} \imeasure(\pair) \KL_\pair (\mdp||\mdp^\dagger) \ge 1$, i.e., is it a point of the feasible region of \eqref{equation_lower_bound}?
If the constraint $\imeasure \in \imeasures(\model)$ oppose no difficulty in general because $\imeasures(\model)$ is a polytope, the second constraint is more troublesome.
In the sequel, we show that if $\models$ is discrete, then checking that an exploration measure is informative enough is computationally difficult (\cref{section_confusing_np_hard}) and that the estimation of $\regretlb_{\jopt}(\model; \models)$ is even harder (\cref{section_estimation_lower_bound}). 

\begin{remark}
    The lower bound $\regretlb_{\jopt}(\model; \models)$ depends on the ambient space $\models$, and the computational hardness of estimating $\regretlb_{\jopt}(\model; \models)$ depends both on $\model$ and $\models$. 
    The regret lower bound $\regretlb_{\jopt}(\model; \models)$ may actually be easy to estimate if $\model$ and/or $\models$ have a special structure, see \cref{section_examples}.
\end{remark}

\subsection{\texttt{CONFUSING-MODEL}: Information tests are NP-complete}
\label{section_confusing_np_hard}

As motivated earlier, a simple question that arises from \eqref{equation_lower_bound} is, given a measure $\imeasure$, how easy it is to check the condition $\forall {\model^\dagger \in\confusing(\model)}, \sum_\pair \imeasure(\pair) \KL_\pair(\model||\model^\dagger) \ge 1$. 
In this paragraph, the problem is shown to be coNP difficult, see the problem \texttt{CONFUSING-MODEL} thereafter. 

\begin{quote}
    \texttt{CONFUSING-MODEL}:
        \textit{
        Given a space of Markov decision processes $\models$, a reference model $\model \in \models$ and a pair of scalars $\alpha, \beta \ge 0$, is there a $\model^\dagger \in \models$ such that:
        \begin{equation}
            \sum_\pair \imeasure(\pair) \KL_\pair(\model||\model^\dagger) < \alpha
            \text{~and~}
            \optgain(\model^\dagger) > \beta,
        \end{equation}
        where $\imeasure$ is an invariant probability measure of $\model$?
    }
\end{quote}

\noindent
Out of simplicity and to essentialize the problem, \texttt{CONFUSING-MODEL} does not require that $\model^\dagger \in \confusing(\model; \models)$. 
Imposing that $\model^\dagger \in \confusing(\model; \models)$ does not change the computational hardness of the problem, see \cref{remark_confusing}.

To be formal, the way the space of MDPs may be fed to an algorithm must be specified.
If $\models$ contains polynomially many models and if every element of it is encoded in polynomial size, then the enumeration of $\models$ is polysized --- In this case, it is clear that \texttt{CONFUSING-MODEL} is polynomial.
Therefore, we focus on spaces $\models$ that cannot be enumerated in polynomial time.
We assume that $\models$ is given by its state-action space together with polynomially many polynomial constraints on its kernel and reward functions. 
Then, we have the following result:

\begin{theorem}
\label{theorem_critical}
    \texttt{CONFUSING-MODEL} is NP-complete.
\end{theorem}

We provide a reduction from the Knapsack Problem (\texttt{KP}), where the choice of $\model^\dagger$ modelizes a choice of items, $\optgain(\model^\dagger)$ corresponds to the aggregate value of that choice and $\sum_\pair \imeasure(\pair) \KL_\pair(\model||\model^\dagger)$ accounts for the required capacity to carry the chosen items. 
We provide a high level overview of the construction and leave the careful tuning of constants to \cref{appendix_complexity}.

\begin{proof}[Proof sketch of \cref{theorem_critical}]
    Proving that it is NP is immediate, because the optimal gain of a MDP is the solution of a LP \cite[§8.8]{puterman2014markov} hence given $\model^\dagger$, checking $\sum_\pair \imeasure(\pair) \KL_\pair(\model||\model^\dagger) < \alpha$ and $\optgain(\model^\dagger) > \beta$ is done in polynomial time.
    The point is to show the NP-hardness.

    To prove that the problem is NP-hard, it is reduced from the Knapsack Problem (\texttt{KP}). 
    Recall that an instance of \texttt{KP} is given by a collection of $n$ items of integer values $\set{v_1, \ldots, v_n}$ and integer weights $\set{w_1, \ldots w_n}$, as well as a capacity $W$ and a value threshold $V$, both integers.
    The problem is to determine whether there exists ${\mathcal{K}} \subseteq [n]$ such that $\sum_{k \in {\mathcal{K}}} w_k \le W$ and $\sum_{k \in {\mathcal{K}}} v_k \ge V$. 

    Fix $\epsilon, \sigma, \delta > 0$ to be tuned later on.
    Given an instance of \texttt{KP}, consider the Markov decision process $\models$ whose structure is given by $n$ ({\sc choose $k$}) $3$-state widgets connected in a ring fashion. 

    \begin{figure}[h]
        \centering
        \begin{tikzpicture}
            \node (0) at (0, 0) {\sc Choose $k$};
            \node (1) at (4, 1) {\sc Pick $k$};
            \node (2) at (4, -1) {\sc Skip $k$};
            \node (3) at (8, 0) {\begin{tabular}{c}\sc Choose\\ $k + 1 \textrm{ mod } n$\end{tabular}};

                \draw[fill=black] (2, .5) circle(.5mm);
                \draw (0) to node[midway, above, sloped] {\scriptsize $\mathrm{N}(\delta, \sigma_k^2)\$$} (2, .5);
                \draw[->,>=stealth] (0) to (2, .5) to node[midway, above, sloped] {\footnotesize $1-\epsilon$} (1);
                \draw[->,>=stealth] (0) to (2, .5) to[out=20,in=180-45] node[pos=.5, above] {\footnotesize $\epsilon$} (2);

                \draw[fill=black] (2,-.5) circle(.5mm);
                \draw (0) to node[midway, below, sloped] {\scriptsize $\mathrm{N}(0, \sigma_k^2)\$$} (2,-.5);
                \draw[->,>=stealth] (0) to (2,-.5) to node[midway, below, sloped] {\footnotesize $\frac 12$} (2);
                \draw[->,>=stealth] (0) to (2,-.5) to[out=-20,in=180+45] node[pos=.25, above] {\footnotesize $\frac 12$} (1);

                \draw[fill=black] (6, 0) circle(.5mm);
                \draw (1) to[in=90,out=0] node[pos=.3, above] {\scriptsize $\mathrm{N}(v_k, 0)\$$} (6, 0); 
                \draw (2) to[in=-90,out=0] node[pos=.3, below] {\scriptsize $\mathrm{N}(0, 0)\$$} (6, 0); 
                \draw[->,>=stealth] (6, 0) to (3);
        \end{tikzpicture}
        \caption{
            \label{figure_widget_main}
            The ({\sc Choose $k$}) widget, where
            $\sigma_k^2 := \frac{\sigma^2}{w_k}$.
            The labels $\mathrm{N}(x, y)\$$ are the (Gaussian) reward distributions.
        }
    \end{figure}
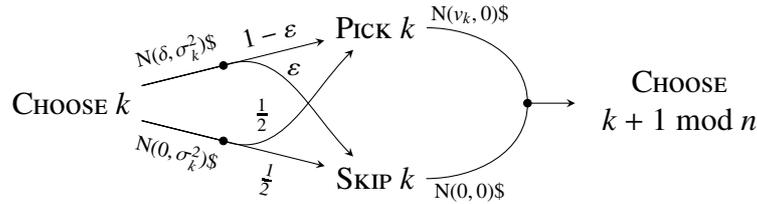

    From the state ({\sc Choose $k$}) are two actions: The top action that is likely to go to ({\sc Pick $k$}) that shall be referred to as action {\sc Pick}, and the bottom action called {\sc Skip}. 
    From every over state, there is a single action that denoted $*$. 
    A (deterministic) policy of $\models$ is analogue to a subset ${\mathcal{K}} \subseteq \set{1, ..., n}$, written $\policy_{\mathcal{K}}$, which is given by $\policy_{{\mathcal{K}}}(\textsc{Pick}|\textsc{Choose}~k) := \indicator{k \in {\mathcal{K}}}$. 
    We get:
    \begin{equation}
    \nonumber
        \gain(\policy_{\mathcal{K}}) 
        = \frac 1{2n} \parens*{
            \frac 12 \sum_{k=1}^n v_k
            +
            \sum_{k \in {\mathcal{K}}} \parens*{\parens*{\tfrac 12-\epsilon} v_k + \delta}
        }
        =
        \frac{\norm{v}_1}{4n}
        + 
        \frac 1{2n}
        \sum_{k \in {\mathcal{K}}} \parens*{\parens*{\tfrac 12-\epsilon} v_k + \delta}.
    \end{equation}

    The \strong{key idea} is that every policy $\policy_{{\mathcal{K}}}$ can equivalently be seen as a \strong{single-action} Markov decision process $\model_{{\mathcal{K}}}$, i.e., the model of a policy over the state-space 
    \begin{equation}
    \nonumber
        \states :=
        \set{(\textsc{Choose }k), (\textsc{Pick }k), (\textsc{Skip }k) : k=1, \ldots, n}.
    \end{equation}
    The choice of an action is equivalently the choice of a kernel distribution. 
    The set of stationary deterministic policies of $\models$, denoted $\policies^\text{SD}(\models)$, can therefore be seen as the set of Markov reward processes $\models^{\text{SD}} := \set{\model_{\mathcal{K}} : {\mathcal{K}} \subseteq \set{1, \ldots, n}}$. 
    Provided that the parameters $\epsilon, \sigma, \delta$ are polynomial in $n, v, w$, this (structured) set of models is described in polynomial size in $n, v, w$.
    Then, carefully choosing $\epsilon, \sigma$ and $\delta$ relatively to $V$ and $W$, we can make it so that the gain of a model $\model_\mathcal{K}$ is directly related to the collected value $\sum_{k \in \mathcal{K}} v_k$, while the divergence $\sum_{\pair} \imeasure(\pair) \KL_\pair (\model||\model^\dagger)$ is directly related to the weight $\sum_{k \in \mathcal{K}} w_k$. 
    Designing $\alpha$ and $\beta$ relatively to $W$ and $V$ respectively, we manage to encode the Knapsack instance with 
    \begin{equation}
    \notag
        \exists? \mathcal{K} \subseteq [n],
        \quad
        \sum_\pair \imeasure(\pair) \KL_\pair(\model||\model_{\mathcal{K}}) \le \alpha
        \text{~and~}
        \optgain(\model_\mathcal{K}) \ge \beta
        .
    \end{equation}
    Details are found in \cref{appendix_complexity}.
\end{proof}

\begin{remark}
\label{remark_confusing}
    In \texttt{CONFUSING-MODEL}, $\model^\dagger$ is not required to belong to $\confusing(\model; \models)$.
    We may consider the problem \texttt{CONFUSING-MODEL*} that adds the constraint that $\model^\dagger \in \confusing(\model; \models)$. 
    The reduction from \texttt{KP} provided in the proof of \cref{theorem_lower_bound} would work similarly, by adding a loop on $\model_\varnothing$ with mean reward $\beta$.
    More specifically, pick the state (\textsc{Choose $0$}) and add a special action $*$ from (\textsc{Choose $0$}) with $\reward(\textsc{Choose }0; *) = \beta$ and $\kernel(\textsc{Choose }0|\textsc{Choose }0, *) = 1$ to all $\model_\mathcal{K}$. 
    Then, all elements $\model^\dagger \in \models$ coincide to each over on $\optpairs(\model_\varnothing) = \set{(\textsc{Choose }0, *)}$.
\end{remark}

\subsection{\texttt{REGRET}: Checking the regret lower bound is co-NP-hard}
\label{section_estimation_lower_bound}

The computation of the regret lower bound seems essentially harder than \texttt{CONFUSING-MODEL}.
However, the computational hardness of \texttt{CONFUSING-MODEL} does not directly propagate to the estimation of $\regretlb_{\jopt}(\model; \models)$.
If $\imeasure, \alpha, \beta$ are close to the values corresponding to the solution of regret optimization problem, is the problem still difficult?
Then answer remains positive so that the computation of the regret lower bound is computationally hard.

\begin{quote}
    \texttt{REGRET}: 
    \itshape
    Given a space of MDPs $\models$, a reference model $\model \in \models$ and a scalar $\rho \ge 0$, does there exists a $\imeasure \in \imeasures(\model/\optpairs(\model))$ such that
    \begin{equation}
        \sum_{\pair \in \pairs} \imeasure(\pair) \ogaps(\pair; \model) \le \rho
        \text{~and~}
        \inf_{\model^\dagger \in \confusing(\model; \models)}
        \braces*{
            \sum_{\pair \in \pairs} \imeasure(\pair) \KL(\model(\pair)||\model^\dagger(\pair))
        } \ge 1
        ?
    \end{equation}
\end{quote}

\noindent
We have the following result.

\begin{theorem}
\label{theorem_conp_hard}
    Checking a solution of \texttt{REGRET} is co-NP complete.
\end{theorem}

We suggest a reduction from the co-knapsack problem that refines the construction of the proof of \cref{theorem_critical}.
Again, we focus here on the construction and leave the details to \cref{appendix_complexity}.

\begin{proof}
    We provide a reduction from the co-knapsack problem (co-\texttt{KP}), which is coNP-complete because \texttt{KP} is NP-complete.
    An instance of co-\texttt{KP} is given by a collection of $n$ items of integer values $\set{v_1, \ldots, v_n}$ and integer weights $\set{w_1,\ldots, w_n}$, as well as a capacity and a value threshold $V$, both integers.
    The problem is to determine if, for all ${\mathcal{K}} \subseteq [n]$, we either have $\sum_{k \in {\mathcal{K}}} w_k \ge W$ or $\sum_{k \in {\mathcal{K}}} v_k \le V$.

    The reduction is very similar to \texttt{CONFUSING-MODEL}'s. 
    Fix $\epsilon, \sigma, \delta, \theta$ to be tuned later on and consider an instance of co-\texttt{KP}.
    Consider the MDP $\models$ whose structure is as given by \cref{figure_knapsack_widget_main}.
    
    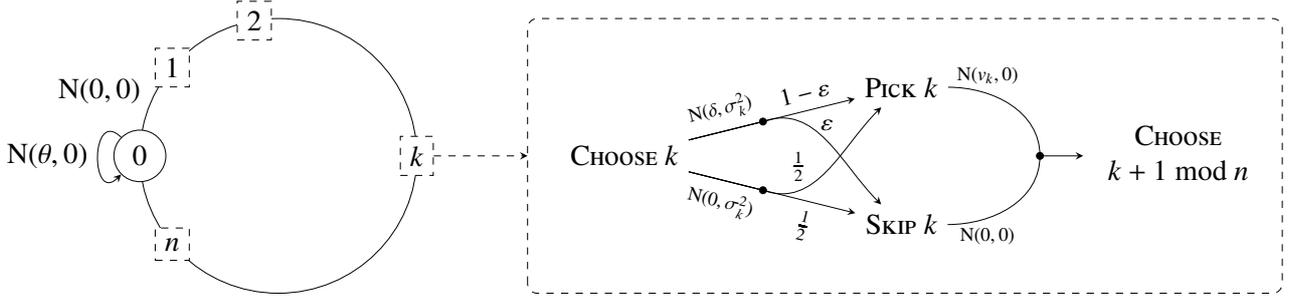
\begin{figure}[h]
        \centering
        \resizebox{\linewidth}{!}{
        \begin{tikzpicture}
            \draw (-2, 0) arc (180:180+360:2);
            \node[fill=white, draw, circle] (0) at (180:2) {$0$};
            \node[fill=white, draw, dashed] (1) at (140:2) {$1$};
            \node[fill=white, draw, dashed] (2) at (100:2) {$2$};
            \node[fill=white, draw, dashed] (n) at (220:2) {$n$};
            \node[fill=white, draw, dashed] (k) at (0:2) {$k$};

            \draw[->, >=stealth, dashed] (k) to (3.6, 0);
            \draw[->, >=stealth] (0) to[in=180+45,out=180-45,looseness=3] node[left,pos=0.5] {$\mathrm{N}(\theta, 0)$} (0);
            \node at (160:2.75) {$\mathrm{N}(0, 0)$};

            \begin{scope}[shift={(5, 0)}]
                \draw[dashed, rounded corners] (-1.4, -2) rectangle (9.5, 2);
                \node (0) at (0, 0) {\sc Choose $k$};
                \node (1) at (4, 1) {\sc Pick $k$};
                \node (2) at (4, -1) {\sc Skip $k$};
                \node (3) at (8, 0) {\begin{tabular}{c}\sc Choose\\ $k + 1 \mathrm{~mod~} n$\end{tabular}};

                    \draw[fill=black] (2, .5) circle(.5mm);
                    \draw (0) to node[midway, above, sloped] {\scriptsize $\mathrm{N}(\delta, \sigma_k^2)$} (2, .5);
                    \draw[->,>=stealth] (0) to (2, .5) to node[midway, above, sloped] {\footnotesize $1-\epsilon$} (1);
                    \draw[->,>=stealth] (0) to (2, .5) to[out=20,in=180-45] node[pos=.5, above] {\footnotesize $\epsilon$} (2);

                    \draw[fill=black] (2,-.5) circle(.5mm);
                    \draw (0) to node[midway, below, sloped] {\scriptsize $\mathrm{N}(0, \sigma_k^2)$} (2,-.5);
                    \draw[->,>=stealth] (0) to (2,-.5) to node[midway, below, sloped] {\footnotesize $\frac 12$} (2);
                    \draw[->,>=stealth] (0) to (2,-.5) to[out=-20,in=180+45] node[pos=.25, above] {\footnotesize $\frac 12$} (1);

                    \draw[fill=black] (6, 0) circle(.5mm);
                    \draw (1) to[in=90,out=0] node[pos=.3, above] {\scriptsize $\mathrm{N}(v_k, 0)$} (6, 0); 
                    \draw (2) to[in=-90,out=0] node[pos=.3, below] {\scriptsize $\mathrm{N}(0, 0)$} (6, 0); 
                    \draw[->,>=stealth] (6, 0) to (3);
            \end{scope}
        \end{tikzpicture}}
        \caption{
        \label{figure_knapsack_widget_main}
            Embedding a knapsack problem in a Markov decision process.
        }
    \end{figure}

    The change regarding the reduction of \texttt{CONFUSING-MODEL} is the state $(0)$, in between ({\sc Choose $n$}) and ({\sc Choose $1$}).
    From $(0)$ you can either loop with the action \textsc{Loop} scoring $\theta$, or go to ({\sc Choose $1$}) with the action {\sc Cycle} scoring $0$, hence entering the big cycle. 
    The state $(0)$ is a special state.
    From the state ({\sc Choose $k$}) are two actions: The top action that is likely to go to ({\sc Pick $k$}) that we shall refer to as action {\sc Pick}, and the bottom action that we shall call {\sc Skip}. 
    From every over state, there is a single action that denoted $*$. 
    The special policy looping on $(0)$ is denoted $\policy^\opt$ and will model the optimal policy later on. 
    The other (deterministic) policies of $\models$ are analogue to a subset ${\mathcal{K}} \subseteq \set{1, ..., n}$, written $\policy_{\mathcal{K}}$, and are given $\policy_{{\mathcal{K}}}(\textsc{Pick}|\textsc{Choose}~k) := \indicator{k \in {\mathcal{K}}}$ with $\policy(\textsc{Cycle}|0) = 1$. 
    We get:
    \begin{equation}
    \nonumber
        \gain(\policy_{\mathcal{K}}) 
        = \frac 1{2(n+1)} \parens*{
            \frac 12 \sum_{k=1}^n v_k
            +
            \sum_{k \in {\mathcal{K}}} \parens*{\parens*{\tfrac 12-\epsilon} v_k + \delta}
        }
        =
        \frac{\norm{v}_1}{4(n+1)}
        + 
        \frac 1{2(n+1)}
        \sum_{k \in {\mathcal{K}}} \parens*{\parens*{\tfrac 12-\epsilon} v_k + \delta}.
    \end{equation}

    Every policy $\policy_{\mathcal{K}}$ can equivalently be seen as a single-action Markov decision process $\model_{\mathcal{K}}$. 
    The choice of an action is equivalently the choice of a kernel distribution. 
    The set of stationary deterministic policies of $\models$, denoted $\policies^\text{SD}(\models)$, can therefore be seen as the set of Markov reward processes $\models^\text{SD} := \set{\model_{\mathcal{K}} : {\mathcal{K}}\subseteq \set{1, \ldots, n}}$.
    The Markov decision process of reference is $\models_\varnothing$ (where the unique optimal policy is looping at the special state $0$) of which every invariant measure is uniform, making the lower bound easier to write. 
    Following that observation, the remaining of the proof is essentially similar to \cref{theorem_critical}. 
    Details are found in \cref{appendix_complexity}.
\end{proof}

\paragraph{Discussion.}

If we focus on convex model classes, the proof of intractability fails. 
The previous reduction doesn't work anymore, because by taking the convex hull of $\models^\text{SD}$, we obtain a space that is very close to the space of randomized policies of $\models$ instead of deterministic ones, which is essential in the proof.
Also, the rational relaxed Knapsack problem is a linear program, so is solvable in polynomial time --- but the optimization problem related to $\models^\text{SR}$ is not exactly the relaxation of the reduced \texttt{KP}, hence we cannot easily claim that $\regretlb(\model; \mathrm{Conv}(\models^\mathrm{SD}))$ is tractable. 
It is only natural to raise the following question.

\paragraph{Open problem.}
\textit{
    If $\models$ is convex,\footnote{e.g., a polyhedron.} does \texttt{REGRET} remain computationally difficult?
}

    \section{Instanciation of the lower bound in classical settings}\label{sec:examples}
    \label{section_examples}

According to \cref{theorem_conp_hard}, $\regretlb_{\jopt}(\model; \models)$ is computationallly difficult to estimate in general.
However, in many scenarios, the environment of interest $\model$ or the considered class of environments $\models$ have special structures that make the regret lower bound easier to describe. 
Alternative descriptions of $\regretlb_{\jopt}(\model; \models)$ can ease the design of asymptotically optimal learning agents.

In this section, we discuss such examples, such as multi-armed bandits, ergodic models, known dynamics problems and beyond.
Our bound is compared to existing results in the literature, and we discuss how the optimization problems \eqref{equation_lower_bound} and \eqref{equation_lower_bound_minor} can be decomposed both in their navigation constraint ``$\imeasure \in \imeasures(\model)$'' or ``$\imeasure \in \imeasures(\model/\optpairs(\model))$'' as well as in their information constraint ``$\forall \model^\dagger \in \confusing(\model)$, $\sum_\pair \imeasure(\pair) \KL_\pair(\model||\model^\dagger) \ge 1$''.

\subsection{Example: Multi-armed bandits}
\label{section_example_bandits}

Instances of \strong{multi-armed bandits} are equivalent to state-less Markov decision processes, i.e., with pair space $\pairs = \set{1} \times \set{1, \ldots, \Action}$. 
In this paragraph, we fix $\pairs = \set{1} \times \set{1, \ldots, \Action}$ and let $\models$ the space of MDPs with Bernoulli rewards and pair space $\pairs$, i.e., $\models \cong \bigcup_{\reward \in [0,1]^\pairs} \bigotimes_{\pair} \mathrm{Ber}(\reward_\pair)$.

Given $\model \in \models$, a policy is a choice of action $\action$ (or \strong{arm}) and its gain $\reward(1, \action)$, so the gap associated to the pair $\pair \equiv (1, \action)$ is $\ogaps(\pair) = \max(\reward) - \reward(\pair)$. 
We see that $\imeasures(\model) = \R_+^\pairs$ hence multi-armed bandits have trivial navigational structure. 
Moreover, confusing models $\model^\dagger \in \confusing(\model)$ are precisely models such that $\exists \pair \notin \optpairs(\model)$ such that $\reward^\dagger(\pair) > \max(\reward) = \optgain(\model)$. 
Combined, \cref{theorem_lower_bound} can be rewritten as:
\begin{equation}
\nonumber
    \regretlb_{\jopt}(\model; \models)
    =
    \inf \set*{
        \sum_{\pair \in \pairs}
        \imeasure(\pair) \ogaps(\pair; \model)
        :
        \imeasure \in \R_+^\pairs
        \text{~and~}
        \forall \model^\dagger \in \confusing(\model),
        \sum_{\pair \in \pairs} \imeasure(\pair) \kl(\reward(\pair)||\reward^\dagger(\pair)) \ge 1
    }.
\end{equation}
Remark that any $\model^\dagger \in \confusing(\model)$ can be simplified into a $\model^\ddagger \in \confusing(\model)$ where $\model$ and $\model^\ddagger$ only differ at one pair $\pair$ (the suboptimal pair of $\model$ that is made optimal in $\model^\ddagger$), consisting in any pair such that $\reward^\dagger(\pair) > \max(\reward)$.
Also, it follows $\confusing(\model) \ne \varnothing$ if, and only if $\max(\reward) < 1$ leading to the well-known interior condition ``$\max(\reward) < 1$''. 
Using the continuity of the function $\reward^\dagger(\pair) \mapsto \kl(\reward(\pair)||\reward^\dagger(\pair))$ on $(0, 1)$, we retrieve the celebrated regret lower bound of \cite{lai_asymptotically_1985}.

\begin{corollary}[\cite{lai_asymptotically_1985}]
\label{corollary_bandit_bernoulli}
    Let $\models$ the space of MDPs with Bernoulli rewards and pair space $\pair$. 
    Let $\model \in \models$ such that $\max(\reward) < 1$.
    Then:
    \begin{equation}
    \label{equation_bandits}
        \regretlb_{\jopt}(\model; \models)
        =
        \sum_{\pair \in \pairs}
        \frac{\ogaps(\pair; \model)}{\kl(\reward(\pair)||\max(\reward))}
        .
    \end{equation}
\end{corollary}

\paragraph{Discussion.}
While \cref{corollary_bandit_bernoulli} is specific to bandits with Bernoulli rewards, it can be straightfully generalized to larger class of bandits (single parameter exponential families or semi-bounded rewards) in which case the denominator $\kl(\reward(\pair)||\max(\reward))$ is changed to $\inf \set{\KL(\rewardd(\pair)||\rewardd^\dagger(\pair)) : \reward^\dagger(\pair) > \max(\reward)}$ where $\rewardd^\dagger(\pair)$ goes other the possible reward distribution on the pair $\pair$.

Several observations can be made regarding \eqref{equation_bandits}.
\begin{enum}
    \item The regret lower bound has a closed-form expression that can easily be evaluated.

    \item Exploration constraints are trivial in multi-armed bandits $\imeasures(\model) = \R_+^\pairs$.

    \item Information constraints are pair-wisely decoupled (or \strong{indexed}) and are equivalent to $\abs{\pairs}$ linear constraints on $\imeasure$, all of the form $\imeasure(\pair) \ge (\kl(\reward(\pair)||\max(\reward)))^{-1}$. 
\end{enum}
The properties (1-3) are not always satisfied. 
As a matter fact, (1) and (3) may fail when considering $\regretlb_{\jopt}(\model; \models')$ for $\models' \subsetneq \models$ (\strong{structured} bandit problems).
Meanwhile, (2) only depends on $\model$ rather than $\models$. 
All together, these properties are arguably what makes multi-armed bandits significatively easier to learn than Markov decision processes. 
However, multi-armed bandits are not the only classes of model spaces with such properties. 

\subsection{Example: (Optimally) Recurrent models, or navigation-free models}
\label{section_example_recurrent}

The property (2) in \cref{section_example_bandits} is written as $\imeasures(\model) = \R_+^\pairs$, but according to \eqref{equation_lower_bound} and \eqref{equation_lower_bound_minor}, we could also see it as $\imeasures(\model/\optpairs(\model)) = \R_+^\pairs$. 
This property only depends on $\model$ and will be referred to as the \strong{navigation-free} property, because the minor $\model/\optpairs(\model)$ is a bandit.
When the environment is navigation-free, then one can commute between any pair with zero cost (and conversely).
This motivates the definition below.

\begin{definition}
\label{definition_optimally_recurrent}
    We say that a model $\model$ is \strong{optimally recurrent} if there exists a gain optimal policy $\policy^*$ whose recurrent states are $\states$; or equivalently, if $\optpairs(\model)$ covers all the states of $\model$; or equivalently again, if $\optpairs(\model) = \weakoptimalpairs(\model)$.
\end{definition}

By definition, $\model$ is optimally recurrent if, and only if $\model/\optpairs$ is state-less.
For instance, ergodic models are optimally recurrent.
The navigation and information constraints can be heavily simplified when the model is optimally recurrent.
Given $\model \in \models$ an optimally recurrent model and $(\state, \action) \in \suboptimalpairs(\model)$, denote:
\begin{equation}
\label{equation_information_recurrent}
    C(\model, \state, \action) 
    := 
    \inf_{\tilde{\rewardd}_{\state,\action}, \tilde{\kerneld}_{\state,\action}}
    \set*{
        \KL\parens*{\rewardd_{\state,\action}\|\tilde{\rewardd}_{\state,\action}}
        +
        \KL\parens*{\kerneld_{\state,\action}\|\tilde{\kerneld}_{\state,\action}}
        :
        \tilde{\reward}(\state,\action) + \tilde{\kernel}(\state,\action) \optbias
        >
        \optgain(\state) + \optbias(\state)
    }
\end{equation}
where $\optgain$ and $\optbias$ are respectively the optimal gain and bias vectors of $\model$.

\begin{proposition}[Lower bound for optimally recurrent models]
\label{proposition_lower_bound_recurrent}
    Assume that $\models$ the space of all models with state-action space $\pairs$.
    If $\model \in \models$ is an optimally recurrent model, then 
    \begin{equation}
    \label{equation_optimally_recurrent}
        \regretlb_{\jopt}(\model)
        =
        \sum_{\pair \notin \optpairs(\model)}
        \frac{\ogaps(\pair)}{C(\model, \pair)}
        .
    \end{equation}
\end{proposition}

\begin{proof}[Proof of \cref{proposition_lower_bound_recurrent}]
    When $\model$ is optimally recurrent, $\model / \optpairs(\model)$ is a single-state Markov decision process, hence $\imeasures(\model/\optpairs(\model)) = \R_+^\pairs$ meaning that the navigation constraints are trivial.
    We now simplify the information constraints using a policy improvement argument which is similar to \cite{burnetas_optimal_1997}.
    Let $\policy^\opt \in \optpolicies(\model)$ with recurrent class $\states$ and pick a confusing model $\model^\dagger \in \confusing(\model)$.
    We have $(\state, \policy^\opt(\state)) \in \optpairs(\model)$ for all $\state \in \states$, and because $\model$ and $\model^\dagger$ coincide on $\optpairs(\model)$, it follows that the gain, bias, reward and kernel of $\policy^\opt$ are preserved in $\model^\dagger$. 
    Yet $\policy^\opt$ is not gain-optimal in $\model^\dagger$, hence is not Bellman optimal in $\model^\dagger$ (\ref{eqn:Bellman}). 
    Accordingly, there must exists $(\state, \action) \in \pairs$ such that:
    \begin{equation*}
        \reward^\dagger(\state, \action) + \kernel^\dagger(\state,\action) \optbias(\model)
        >
        \optgain(\state, \model) + \optbias(\state, \model)
        .
    \end{equation*}
    Meanwhile, all states are recurrent under $\policy^\opt$ which is optimal in $\model$, hence (1) $\optpairs(\model) = \weakoptimalpairs(\model)$ and (2) it is a fixpoint of the Bellman operator of $\model$, in particular $\reward(\state,\action) + \kernel(\state,\action) \optbias(\model) \le \optgain(\state,\model) + \optbias(\state,\model)$.
    By (1), $\model$ and $\model^\dagger$ coincide on $\weakoptimalpairs(\model)$ so invoking (2), we must have $(s, a) \notin \weakoptimalpairs(\model)$.
    In the end, we obtain:
    \begin{equation*}
        \regretlb(\model)
        =
        \inf\set*{
            \sum_{\pair \in \pairs} \imeasure(\pair) \ogaps(\pair)
            :
            \imeasure \in \R_+^\pairs 
            \text{~and~}
            \forall (\state,\action) \in \suboptimalpairs(\model),
            ~
            \imeasure(\state,\action) C(\model, \state,\action)
            \ge 
            1
        }\,,
    \end{equation*}
    of which the solution is obvious.
\end{proof}

\paragraph{Discussion.}
In particular, when $\model \in \models$ is an optimally recurrent model, then the navigation constraints become trivial ($\imeasure \in \R_+^\pairs$) and the information constraints are decoupled along sub-optimal pairs, just like for multi-armed bandits. 
We recover a regret lower-bound that is in closed form, navigation free and with pair-wisely decoupled information constraints.
In other words, the regret lower bound of optimally recurrent models is morally the same than than in multi-armed bandits. 
The obtain bound is a generalization of the well-known results of \cite{burnetas_optimal_1997}, that provides a lower bound which is  similar to \cref{proposition_lower_bound_recurrent}, excepted that it only holds for $\model$ \strong{irreducible} (or \textbf{recurrent}), meaning that all the states are recurrent under every policy --- which is much stronger than the optimal recurrence assumption of \cref{definition_optimally_recurrent}.

\subsection{Example: Bandits with switching costs}
\label{section_example_bandits_switching_costs}

We now go back to multi-armed bandits by adding switching costs.
Multi-armed bandits with switching costs \cite{agrawal_switching_1988,jun_survey_2004,brezzi_optimal_2002,ortner_online_2010} consist in multi-armed bandits where a cost in induced every time the learning agent changes arm.
Formally, a bandit with switching costs can be seen as a Markov decision process $\model$ with pair space $\pairs = \states \times \states$ where each state corresponds to an arm, and actions from a state are another choice of state.
By choosing $\state$ from $\state$, one remains in $\state$ and scores according to $\rewardd(\state, \state')$ of positive mean; And by choosing $\state' \ne \state$ from $\state$, one goes to $\state'$ and pays a fixed switching cost $\lambda > 0$.
That is,
\begin{eqnarray}
    \nonumber
    \kernel(\state'|\state, \state) = \state
    & \text{\quad and \quad}
    & \reward(\state, \state') \ge 0
    ,
    \\
    \nonumber
    \kernel(\state'|\state, \state) = \state'
    & \text{\quad and \quad}
    & \reward(\state, \state') = - \lambda
    .
\end{eqnarray}
\noindent
For a fixed switching cost $\lambda > 0$ and a fixed set of arms $\states$, we let $\models_\lambda$ the collection of all multi-armed bandits with switching cost $\lambda$ and Bernoulli rewards, consisting in the set of Markov decision processes described above. 
With a similar rationale than in \cref{section_example_bandits}, we show that the regret lower bound can be put in closed form, meeting the known result of \cite{agrawal_switching_1988}.

Indeed, just as for bandits, we can show that any confusing model $\model^\dagger \in \confusing(\model; \models_\lambda)$ can be simplified into $\model^\ddagger$ where $\model$ and $\model^\ddagger$ only differ at an optimal pair of $\model^\dagger$. 
When rewards are Bernoulli, information constraints are therefore decoupled along states and can be written as:
\begin{equation}
\label{equation_switching_bandits_1}
    \forall \state \in \states,
    \quad
    \imeasure(\state) 
    \kl(\reward(\state,\state)||\max(\reward)) 
    \ge 1
    .
\end{equation}
Then, it is easy to see that the structure of bandits with switching cost is such that for all $\alpha \in \R_+^\states$, there is $\imeasure \in \imeasures(\model)$ such that $\imeasure(\state, \state) = \alpha(\state)$. 
Moreover, just like for multi-armed bandit, the optimal gain is $\optgain(\model) = \max(\reward)$ regardless of the switching cost. 
Combining it all with \eqref{equation_switching_bandits_1}, we obtain \cref{corollary_switching_bernoulli}.

\begin{corollary}[\cite{agrawal_switching_1988}]
\label{corollary_switching_bernoulli}
    Fix $\lambda > 0$ and let $\models_\lambda$ the space of all Bernoulli bandits with switching costs $\lambda$ and arms $\states$.
    Let $\model \in \models_\lambda$ such that $\max(\reward) < 1$.
    Then:
    \begin{equation}
    \label{equation_bandits_switching_cost}
        \regretlb_{\jopt}(\model; \models_\lambda)
        =
        \sum_{\pair \in \pairs}
        \frac{\ogaps(\pair; \model)}{\kl(\reward(\pair)||\max(\reward))}
        .
    \end{equation}
\end{corollary}

\paragraph{Discussion.}
Remark that despite the fact that $\imeasures(\model) \ne \R_+^\pairs$, the navigation constraints do not appear in that reworked lower bound. 
This is because the information constraints do not constrain $\imeasure$ outside of $\set{(\state, \state): \state \in \states}$, hence the optimal exploration measure has zero mass on $\pairs \setminus \set{(\state, \state): \state \in \states}$ which is indeed where navigation is constrained. 
This is linked to the surprising fact that $\lambda$ is absent from $\regretlb_{\jopt}(\model; \models_\lambda)$, hence that the optimal asymptotic regret does not depend on switching costs. 
In other words, our main regret lower bound of \cref{theorem_lower_bound} claims that \strong{switching costs are negligible}.
It is not obvious at first sight that there are strongly consistent agents of which the asymptotic regrets are reaching $\regretlb_{\jopt}(\model; \models_\lambda)$, or independent of $\lambda$ at all --- we refer the interested reader to \cite{agrawal_switching_1988,brezzi_optimal_2002} for more insight on the subject.

\subsection{Example: Fixed kernel spaces}
\label{section_example_fixed_kernel_spaces}

Beyond optimally recurrent models, the lower bound may be impossible to decouple pair-wisely in general.
It may have no closed-form expression and navigation constraints may be non trivial.
One setting in which the lower bound cannot be decoupled, and which is slightly simpler than the general setting yet, is when the transition kernel is known or deterministic. 
This setting is the subject of a few works \cite{saber_logarithmic_2024,ortner_online_2010,tranos_regret_2021}. 
\cite{tranos_regret_2021} provides a lower bound for deterministic transition models and while \cite{saber_logarithmic_2024} provides model dependent regret analysis of their method in the more general setting where only the reward function has to be learned, no lower bound was known in this more general setting. 

\begin{definition}
    Fix a pair space $\pairs$.
    A model space $\models \in \modelss(\pairs)$ is called a \textbf{fixed kernel space} if all elements of $\models$ have the same transition kernel, i.e., $\forall \model, \model' \in \models$, $\kernel = \kernel'$. 
    We say that it is \textbf{deterministic kernel space} if, in addition, $\kernel(\state'|\state,\action) \in \set{0, 1}$ for every transition triplet $(\state, \action, \state')$.
\end{definition}

We start by relating our results to the lower bound of \cite{tranos_regret_2021}.
For deterministic kernel space, \cite[Theorem~1]{tranos_regret_2021} provided a model dependent lower bound and already make apparent the information and navigation constraints in a form that are similar to \cref{theorem_lower_bound}, although the objective function is written differently. 
In our set of notations, their result can be adapted as follows.

\begin{theorem}[\cite{tranos_regret_2021}]
    Let $\models$ a deterministic kernel space and fix $\model \in \models$.
    The regret of every strongly consistent learning agent satisfies $\Reg(T; \alg, \model) \gtrsim C(\model; \models) \log(T)$ where $C(\model; \models)$ is:
    \begin{equation}
        \inf \set*{
            \sum_{\pair \in \pairs}
            \imeasure(\pair)(\optgain - \reward(\pair))
            :
            \imeasure \in \imeasures(\model)
            \text{~and~}
            \inf_{\model^\dagger \in \confusing(\model; \models)}
            \sum_{\pair \in \pairs}
            \imeasure(\pair) \KL(\model(\pair)||\model^\dagger(\pair))
            \ge 1
        }
        .
    \end{equation}
\end{theorem}

This lower bound is the same as ours, with $\regretlb_{\jopt}(\model; \models) = C(\model; \models)$, because
\begin{align*}
    \sum_{\pair \in \pairs}
    \imeasure(\pair)(\optgain - \reward(\pair))
    & \overset{(*)}=
    \sum_{(\state, \action) \in \pairs}
    \imeasure(\state, \action) 
    \parens*{ 
        \ogaps(\state, \action)
        +
        (\kernel(\state, \action) - e_\state) \optbias
    }
    \\
    & =
    \sum_{\pair \in \pairs} \imeasure(\pair) \ogaps(\pair)
    + 
    \sum_{\state' \in \state} \optbias(\state') \sum_{(\state, \action)\in \pairs}
    \imeasure(\state, \action) \kernel(\state'|\state, \action)
    -
    \sum_{\state \in \states} \optbias(\state) 
    \sum_{\action \in \actions(\state)} \imeasure(\state, \action)
    \\
    & \overset{(\dagger)}=
    \sum_{\pair \in \pairs} \imeasure(\pair) \ogaps(\pair)
    + 
    \sum_{\state' \in \states} \optbias(\state') 
    \sum_{\action \in \actions(\state')} \imeasure(\state', \action)
    -
    \sum_{\state \in \states} \optbias(\state) 
    \sum_{\action \in \actions(\state)} \imeasure(\state, \action)
    \\
    & = 
    \sum_{\pair \in \pairs} \imeasure(\pair) \ogaps(\pair)
\end{align*}
where 
$(*)$ follows from the Poisson equation and
$(\dagger)$ uses that $\imeasure \in \imeasures(\model)$.
Therefore and in particular, our results generalizes \cite{tranos_regret_2021} to fixed kernel spaces. 

\paragraph{Policy decomposition and fixed kernel spaces.}
When a model $\model^\dagger$ is confusing, it means that there exists a policy which is optimal in $\model^\dagger$ and sub-optimal in $\model$.  
This is also true the other way around: Confusing models can be enumerated according to the sub-optimal policy that they make optimal, encouraging the introduction of confusing models \strong{related} to a policy, see \cref{definition_beneficial_set}.
\begin{definition}
    \label{definition_beneficial_set}
    Let $\model \in \models$ and fix $\policy \notin \optpolicies(\model)$.
    The \strong{beneficial set} related to $\policy$ is:
    \begin{equation}
        \beneficial(\policy, \model; \models)
        :=
        \set*{
            \model^\dagger \in \models
            :
            \gain_\policy(\model^\dagger)
            > 
            \sup_{\optpolicy \in \optpolicies(\model)}
            \gain_{\optpolicy}(\model^\dagger)
        }
        .
    \end{equation}
    We further denote $\beneficial_{\jopt}(\policy, \model; \models) := \beneficial(\policy, \model; \models) \cap \confusing(\model; \models)$.
\end{definition}
This decomposition of the confusing set is present in many works, including \cite{saber_logarithmic_2024,marjani_adaptive_2021,marjani_navigating_2021}. 
Now, fixed kernel spaces are a setting where a classical decomposition of the confusing set is especially pertinent. 
Indeed when the space $\mdps$ is a \strong{convex} fixed kernel space, then the set $\beneficial_{\jopt}(\policy, \model; \models)$ of both confusing and beneficial models related to a policy $\pi$ is a convex set.  
Further,  the infimum of $\sum_{\pair \in \pairs} \imeasure(\pair) \KL_\pair(\model||\model^\dagger)$ 
over $\model^\dagger \in \beneficial_{\jopt}(\policy, \model; \models)$ can be computed numerically as:
\begin{equation*}
    \inf \set*{
        \sum_{\pair \in \pairs}
        \imeasure(\pair)
        \KL(\reward(\pair)||\reward^\dagger(\pair))
        :
        \reward_{|\optpairs(\model)} = \reward^\dagger_{|\optpairs(\model)}
        \text{~and~}
        \gain_{\policy}((\reward^\dagger, \kernel)) > \optgain(\underbrace{(\reward, \kernel)}_{\model})
    }
\end{equation*}
This is a convex program with a strongly convex objective function and linear constraints, because $\reward^\dagger \mapsto \gain_{\policy}(\reward^\dagger, \kernel)$ is a linear function. 
So, while in general, the regret lower bound $\regretlb_{\jopt}(\model; \models)$ is the solution of an optimization program with a \strong{continuum} of convex constraints \eqref{equation_lower_bound}, it becomes the solution of an optimization program with \strong{exponentially} many convex constraints (one per policy) when $\models$ is a convex fixed kernel space \eqref{equation_lower_bound_policy}.
As constraints are indexed by policies, good heuristics on politics are crucial to get good estimates of $\regretlb_{\jopt}(\model; \models)$. 
As a matter of fact, using heuristics to select seemingly critical policies and information constraints is a key component of the algorithm of \cite{saber_logarithmic_2024}, even though \cite{saber_logarithmic_2024} does not try to approach the regret lower bound perfectly. 

\subsection{Policy-wise decompositions of the lower bound}
\label{section_example_policy_wise}

Following \cref{section_example_fixed_kernel_spaces}, fixed kernel spaces are a setting where a general decomposition of the confusing set is especially pertinent.  
We provide a few additional remarks about this policy-wise decomposition.

\subsubsection{Beneficial sets and fixed kernel spaces}

Remark that $\beneficial(\policy, \model; \models) \subseteq \alternative(\model)$.
We leave as an exercise to the reader that $\confusing(\model; \models) = \bigcup_{\policy \notin \optpolicies(\model)} \beneficial_{\jopt}(\policy, \model; \models)$.
From this fact, it immediately follows that $\regretlb_{\jopt}(\model;\models)$ can be written as:
\begin{equation}
\label{equation_lower_bound_policy}
\tag{LB${}_{\jopt}^\policy$}
\regretlb_{\jopt}(\mdp ; \mdps)=\!\inf\!\bigg\{\!\sum_{x\in\cX} \mu(x) \ogaps(x): \mu \in \imeasures(\mdp),  \forall \pi \notin \optpolicies(\mdp),  \bbU(\pi,\mu,\mdp) \!\geq \!1
\bigg\}\,.
\end{equation}
Here, we have introduced the \strong{unlikelihood of optimality} $\bbU(\pi,\mu,\mdp)$ of the policy $\policy$, borrowing the terminology from \cite{pesquerel2023information}, which is formally defined as
\begin{equation}
    \label{def:unlikelihood}
    \bbU(\pi,\mu,\mdp)
    :=
    \inf \set*{
        \sum_{x\in\cX}\mu(x)\KL_x(\mdp||\mdp^\dagger)
        :
        {\mdp^\dagger \in \beneficial_{\jopt}(\policy, \model; \models)}
    }
    \,.
\end{equation}
For general $\model$ and $\models$, $\beneficial_{\jopt}(\policy, \model; \models)$ is not necessarily convex and this expression cannot be argued to be simpler or more computationally friendly than the original \eqref{equation_lower_bound}.
However, when $\bbU(\pi,\mu,\mdp)$ can be computed efficiently for each policy, this rewriting becomes appealing as it involves finitely-many policy-wise constraints rather than a continuum of confusing models. 

\subsubsection{Policy-wise lower bounds, ergodic environments and  \cite{agrawal_asymptotically_1988}}

If the information constraints can be rewritten in a policy-wise fashion, as a last detour, we emphasize on the observation that the whole lower bound of \cref{theorem_lower_bound} can be rewritten \strong{policy}-wisely. 
This is actually how ergodic Markov decision processes have been first approached \cite{agrawal_asymptotically_1988,graves1997asymptotically}. 
In an ergodic environment, every policy can be played until regeneration, i.e., until coming back to the initial state. 
Doing so, the expected regret itself can be written in terms of policies, leading to the lower bound of \cite{agrawal_asymptotically_1988} that rewrites, when $\model$ is ergodic, as
\begin{equation}
\nonumber
    \regretlb(\model)
    = 
    \inf \set*{
        \sum_{\policy \in \policies} 
        \alpha(\policy) \parens*{
            \optgain(\model) \!-\! \gain_\policy(\model)
        }
        :
        \alpha \in \probabilities(\policies)
        \text{~and~}
        \sum_{\policy \in \policies}
        \alpha(\policy)
        \bbU(\pi,\mu_\pi,\mdp)
        \ge
        1
    }\,.
\end{equation}
The quantity $\alpha(\policy)$ accounts for the amount of time (in logarithmic scale) that the algorithm spends playing the policy $\policy$.
The above coincides with the lower bound of \cite{burnetas_optimal_1997}. 
However, it is not clear that it can be generalized beyond the scope of ergodic Markov decision processes. 
Indeed, the expected regret can only be expressed as $\sum_{\policy \in \policies} \alpha(\policy) \parens*{ \optgain(\model) - \gain_\policy(\model) }$ if the algorithm plays fixed policies until regeneration, hence heavily relying on the ergodic nature of the environment.
Moreover, algorithms play actions rather than policies and algorithms that explore by playing fixed policies for long periods of time are, although common, of a special kind.

    \section{An elementary construction of confusing MDPs}\label{sec:contruction}
    
We have seen in the previous section that the constraint on MDPs can be decomposed into constraints on policies, via $\confusing(\model; \models) = \bigcup_{\policy \notin \optpolicies(\model)} \beneficial_{\jopt}(\policy, \model; \models)$.
In this section, we aim to building confusing MDPs from an elementary constructive perspective.
To this aim, we examine specific scenarios where constraints on policies can be further replaced with constraints on individual state-action pairs. The most specific situation happens when modifying a \strong{single pair} $x$ in a Markov Decision Process (MDP) suffices to construct a confusing instance for a sub-optimal policy $\pi\notin \optpolicies(\model)$.
In this case, this means that the subset of MDPs $\mdps_1(x, \mdp)$, resulting from a \strong{single-pair} modification at $x \notin \optpairs(\mdp)$, satisfies $\beneficial_{\jopt}(\policy, \model; \models) \cap \mdps_1(x, \mdp) \neq \emptyset$.
Consequently, in this specific situation the unlikelihood of optimality of $\pi$ is controlled as $\bbU(\pi, \mu, \mdp)\leq \mu(x) \inf \big\{ \KL_x(\mdp, \mdp^\dagger) : \mdp^\dagger \in \beneficial_{\jopt}(\policy, \model; \models) \cap \mdps_1(x, \mdp) \big\}$,
which, together with the constraint $\bbU(\pi, \mu, \mdp)\geq 1$ yields a pair-wise constraint on $\mu(x)$.

To guide the construction of the general case, we need to scrutinize the structure policies and corresponding recurrent sets.
To this aim, we introduce the set $\cS_{\neq}(\pi, \pi^\dagger) := \{ s \in \cS : \pi(s) \neq \pi^\dagger(s) \}$
of states where two policies $\pi,\pi^\dagger$ differ, then characterize the proximity of policies by the size of this set:
\begin{definition}[$k$-neighborhoods]
	For a policy $\pi \in \policies$, the \strong{$k$-neighborhood} $\cV_{\pi}(k):= \{ \pi^\dagger \in \policies : |\cS_{\neq}(\pi, \pi^\dagger)| \leq k \}$ is defined as the set of policies differing from $\pi$ in at most $k$ states.
\end{definition}
Of particular interest is the set $\cV_{\jopt}(1) = \{ \pi : \exists \pi^\star \in \optpolicies(\mdp), \pi \in \cV_{\pi^\star}(1) \}$, representing policies that are immediate neighbors of optimal policies. These policies, denoted as $\pi = \pi^\star_x$, deviate from a specific optimal policy $\pi^\star$ only at the state $x$.  
Such policies are  especially appealing when  $\beneficial_{\jopt}(\policy^\star_x, \model; \models) \cap \mdps_1(x, \mdp)\neq\emptyset$,
since $x$ is also the pair where to modify the MDP to obtain a confusing one. However, note that this favorable situation may not happen in general, which means \emph{even single-pair modification of an optimal policy may require many pairs modifications} to build a corresponding confusing MDP.
Restricting in $\regretlb_{\jopt}(\mdp)$ the policy-wise constraints to the subset $\cV_{\jopt}(1)$ only may considerably reduce the set of constraints and yield computable approximations of this quantity.

To streamline the discussion, we assume below that the set of permissible rewards and transitions in the MDP is fully unstructured. In other words, no constraints link rewards and transitions across different pairs or within a single pair. This assumption isolates the generic mechanisms at play, free from the added complexity of structural dependencies, which may render some cases more or less challenging depending on the specific constraints.
Formally, we assume that the set of rewards and transitions in the MDP can be expressed as a product set, $\bigotimes\limits_{x \in \cX} (\cR_x \otimes \cP_x)$, where $\cR_x \subset \cP(\mathbb{R})$ represents the set of reward distributions and $\cP_x \subset \cP(\cS)$ denotes the set of transition distributions for each state-action pair $x$.
 Under this assumption, the subset of MDPs $\mdps_1(x, \mdp)$, resulting from a \strong{single-pair} modification at $x \notin \optpairs(\mdp)$, can be represented as $\cR_x \otimes \cP_x$. 
 
\subsection{Locally modifying an MDP to induce confusion}
\label{section_locally_modifying}

The construction of a confusing instance for a sub-optimal policy $\pi$ often involves selectively modifying an MDP $\mdp$ at recurrent state-action pairs under $\pi$. The objective is to either increase the rewards or adjust the transitions to amplify the visiting frequency of these pairs, provided the resulting MDP remains within the permissible set $\mdps$. While in some cases a single-pair modification suffices, in general, multiple pairs may need to be adjusted.

\subsubsection{Local modifications of an MDP}

Formally, for a configuration set $\mdps$ and a state-action pair $x = (s, a) \notin \pairs_{\jopt}(\mdp)$, we define two key sets:
\begin{align*}
	\text{(Local uncertainty set)} \quad \cC(x, \mdps) &:= \{ (\rho, q) \in \cP(\mathbb{R}) \times \cP(\cS) : \exists \mdp \in \mdps, {\bf r}_{\mdp}(x) = \rho, {\bf p}_{\mdp}(x) = q \}, \\
	\text{(Local optimistic set)} \quad \cC_+(x, \mdps) &:= \{ (\rho, q) \in \cC(x, \mdps) : \mathbb{E}_{\rho}[r] + (q \mathbf{b}_{\pi^\star})(s) > \mathbf{m}_{\pi^\star}(s) + (\mathbf{p}_{\pi^\star} \mathbf{b}_{\pi^\star})(s) \}.
\end{align*}

The locally modified MDP $\tmdp(x)$ matches $\mdp$ at all pairs except $x$, where the reward and transition are adjusted as follows:
\[
(\tilde{\mathbf{p}}(x), \tilde{\mathbf{r}}(x)) := \begin{cases}
	\argmin \{ \KL(\mathbf{r}(x), \rho) + \KL(\mathbf{p}(x), q) : (\rho, q) \in \cC_+(x, \mdp) \} & \text{if } \cC_+(x, \mdp) \neq \emptyset, \\
	\argmax \{ \mathbb{E}_{\rho}[r] + (q \mathbf{b}_{\pi^\star})(s) : (\rho, q) \in \cC(x, \mdp) \} & \text{otherwise}.
\end{cases}
\]
Intuitively, such elementary modifications aim to either increase the reward of this pair, enhancing the gain of a recurrent sub-optimal policy, or alter the transition probabilities to redistribute the invariant measure toward more rewarding pairs—or both. 
Now to ensure that stringent constraints on $\cC(x, \mdps)$ are appropriately handled, if $\cC_+(x, \mdp)$ is empty, this procedure selects the admissible modification that maximally increases the pair’s contribution to the gain.

\subsubsection{Specific Local Modification Sets}

Two subsets of $\cC_+(x, \mdp)$ merit special attention, each imposing distinct restrictions:
The first set $\cC^{\texttt{sw}}(x,\mdps)$ restricts to transitions that switch the behavior of action $a$ into that of an optimal policy $\pi^\star$, and is reminiscent of \cite{burnetas_optimal_1997} for ergodic MDPs.
The second set $\cC^{\texttt{r}}(x,\mdps)$ forces to keep the transition unchanged, hence only rewards can be modified.
Formally,
\beqan
\cC^{\texttt{sw}}(x,\mdps)&:=&\{(\rho,q) \in\cC(x,\mdps): q = (1-\epsilon) \bfp_{a}(s) + \epsilon \bfp_{\pi^\star}(s), \epsilon\in(0,1]\}\\
\cC^{\texttt{r}}(x,\mdps)&:=&\{(\rho,q) \in\cC(x,\mdps): q =\bfp_a(s)\}\,,
\eeqan
We define both $\cC_+^{\texttt{sw}}(x,\mdps)$, $\cC_+^{\texttt{r}}(x,\mdps)$ and $\tmdp^{\texttt{sw}}(x)$, $\tmdp^{\texttt{r}}(x)$ replacing $\cC(x,\mdp)$ with $\cC^+(x,\mdp)$ accordingly.
\begin{remark}[Switched Markov chain]
	In particular, when $\tilde \bfp(x)=q=\bfp_{\pi^\star}(s)$, the Markov chain induced by playing $x=(s,a)$ in $\tmdp^{\texttt{sw}}(x)$ coincides with 
	playing $\pi^\star$ in $\mdp$,  by construction.\\
	When considering a switch MDP $\mdp^\dagger$ such that $q=\bfp_{\mdp^\dagger}(x)=\bfp_{\pi^\star}(s)$ (so with $\epsilon=1$), then the condition in the definition of the local optimistic set simplifies into  $\Esp_{\rho}[r]> {\bf m}_{\pi^\star}(s)$. 
\end{remark}

\subsubsection{Choice of local modifications}
We choose either set in the computation of a local modification depending on the corresponding state-action pair $x=(s,\pi(s))$.
Note that we  locally modify an MDP  only when $\cS(\pi,\mdp) \cap \cS_{\neq}(\pi,\pi^\star)\neq\emptyset$, since otherwise we can show that
$\pi\in\optpolicies(\mdp)$. For the remaining cases, we proceed as follows
\begin{itemize}
	\setlength\itemsep{-0.5em}
\item When  $s \in  \cS(\pi,\mdp) \cap \cS_{\neq}(\pi,\pi^\star) \cap \cS(\pi^\star,\mdp)$, we build $\tmdp^{\texttt{sw}}(x)$  where $x=(s,\pi(s))$.

\item When  $s \in \cS(\pi,\mdp) \cap \cS_{\neq}(\pi,\pi^\star) \setminus \cS(\pi^\star,\mdp)$, we build $\tmdp^{\texttt{r}}(x)$  where $x=(s,\pi(s))$.

\item  When  $s\in \cS(\pi,\mdp)\setminus (\cS_{\neq}(\pi,\pi^\star)\cup\cS(\pi^\star,\mdp))$, we build $\tmdp^{\texttt{r}}(x)$ where $x=(s,\pi(s))$.
\end{itemize}
For convenience, we denote the union of all these states by $\states^{\texttt{ch}}(\pi,\mdp)$, their corresponding pairs by $\pairs^{\texttt{ch}}(\pi,\mdp)=\{(s,\pi(s)) : s\in \states^{\texttt{ch}}(\pi,\mdp)\}$, and the modified MDP $\tmdp^{\texttt{ch}}(x)$ (either
 $\tmdp^{\texttt{sw}}(x)$  or  $\tmdp^{\texttt{r}}(x)$) corresponding to the situation.
 Now let us recall that a confusing MDP $\mdp^\dagger$ should coincide with $\mdp$ on  $\optpairs(\mdp)$.
 In particular, provided $\mdp^\dagger$ does not modify the recurrent pairs of any policy in $\optpolicies(\mdp)$,
 then the gain and bias of any optimal policy $\pi^\star\in\optpolicies(\mdp)$ is the same in $\mdp$ and $\mdp^\dagger$. This is a key property that justifies the local optimist set defines valid confusing instances,
 and this is the reason why we only modify the transitions of the MDP in pairs where $\pi$ differs from $\pi^\star$. More formally:
 \begin{proposition}\label{prop:conf_keepopt}
 	For each $\mdp^\dagger \in \conf(\mdp)$, $\pi^\star\in\optpolicies(\mdp)$ such that $\pairs(\pi^\star,\mdp)=\pairs(\pi^\star,\mdp^\dagger)$, then  $\bfg_{\pi^\star}(\cdot,\mdp)=\bfg_{\pi^\star}(\cdot,\mdp^\dagger)$ and $\bfb_{\pi^\star}(\cdot,\mdp)=\bfb_{\pi^\star}(\cdot,\mdp^\dagger)$. 
 \end{proposition}

\begin{remark}[Changes outside of $\cS_{\neq}(\pi,\pi^\star)$]
	On  the states $\cS(\pi,\mdp)\setminus \cS_{\neq}(\pi,\pi^\star)$, one has $\pi=\pi^\star$. Hence modifying the transitions of $\mdp$ at pairs $(s,\pi(s))=(s,\pi^\star(s))$ may
	actually modify $\pairs(\pi^\star,\mdp)$. This is why we keep the transition unchanged in this case.	Further, when $s\in\cS(\pi^\star,\mdp)$ increasing the rewards also increases the gain of $\pi^\star$ in the modified MDP, which is prohibited. This is why we exclude the set $\cS(\pi^\star,\mdp)$ and only  increasing rewards on $\cS(\pi,\mdp)\setminus (\cS_{\neq}(\pi,\pi^\star)\cup\cS(\pi^\star,\mdp))$. 
\end{remark}

\paragraph{When a single pair modification is enough.}
To give some intuition, let us consider a policy $\pi$ such that $\cS_{\neq}(\pi,\pi^\star)=\{s\}$. 
In particular, $\pi\in\cV_{\jopt}(1)$ and this policy differs only at a single pair $x=(s,\pi(s))$ from an optimal policy. 
We say the \texttt{Switch} property holds if $\tmdp_{\texttt{sw}}(x) \in  \beneficial_{\jopt}(\policy, \model; \models)$, that is if a single-pair modification is enough to build a confusing MDP for policy $\pi$, when built from the local set $\cC_{\texttt{sw}}(x,\mdps)$. 
The following proposition justifies the choice of building  $\tmdp^{\texttt{sw}}(x)$ when  $s \in  \cS(\pi,\mdp) \cap \cS(\pi^\star,\mdp)$.

\begin{proposition}[\texttt{Switch} property]\label{prop:localsimplification}
	Let $x=(s,\pi(s))$ and assume that $\cC_+^{\texttt{sw}}(x)\neq \emptyset$.%
    \vspace{-0.33em} 
    \begin{enum}
        \item 
            If $\pi^\star$ is ergodic in $\mdp$,  then \texttt{Switch} 	holds at $x$.
        \item 
	        If $\pi^\star$ is unichain in $\mdp$, then 	\texttt{Switch}
	holds at $x$ except when  $s\notin \cS(\pi^\star,\mdp)$.
	    \item
	        If $\pi^\star$ is multichain in $\mdp$, then \texttt{Switch} holds at $x$ when $s\in \cS(\pi,\mdp) \cap \cS(\pi^\star,\mdp)$ and does not hold when $s \notin \cS(\pi^\star,\mdp)$.
    \end{enum}
\end{proposition}

In particular, when  restricting the policy-wise constraints in $\regretlb_{\jopt}(\mdp)$ to the subset $\cV_{\jopt}(1)$,
we obtain the following relaxation expressed  with pair-wise constraints:
\[
\regretlb_{\jopt}(\mdp) \geq \inf \bigg\{ \sum_{x \in \cX} \mu(x) \ogaps(x) : \mu \in \imeasures(\mdp), \forall x \in \pairs,
\begin{cases}
\mu(x) \KL_x(\mdp, \tmdp_{\texttt{sw}}(x) ) \geq 1  & \text{if } \texttt{Switch}(x)  \\
	\bbU(\pi^\star_x, \mu, \mdp) \geq 1 & \text{otherwise}.
\end{cases}
\bigg\}.
\]

\begin{remark}[Prior knowledge]
	Let us remark that if prior knowledge is assumed about 
	${\bf p}(x)$, say we know that ${\bf p}(x)\in\cP_x$, where $\cP_x$ is a strict subset of $\cP(\cS)$
	(meaning all MDP $\tmdp$ in $\mdps$ must satisfy $\tilde {\bf p}(x)\in\cP_x$),
	then $\cC_+^{\texttt{sw}}(x)$ may be empty: This happens e.g. if  $x=(s,a)$ and 	${\bf p}(s,\pi^\star(s))\notin \cP_{s,a}$.
	Hence the complexity of deriving the lower bound in the agnostic situation when $\cP_x=\cP(\cS)$ is possibly very different than 	when we have a more informative knowledge.
\end{remark}

\paragraph{Adapting to multi-pair modification.}
When no MDP can be found that is sufficiently optimistic in a single modification, we need to progressively modify the MDP at other pairs in order to increase the gain of policy $\pi$ and not only at single place.
We suggest using here a simple iterative procedure that successively modifies $\mdp$ at well-chosen pairs. 
We let $\mdp_0=\mdp$, then for each $i\geq 0$, we choose $x_i\in\pairs^{\texttt{ch}}(\pi,\mdp)\setminus\{x_{1},\dots,x_{i-1}\}$ in an appropriate way.
Then, we build
$\mdp_{i+1}=\tmdp_i^{\texttt{ch}}(x_i)$, that is we build a local modification of $\mdp_i$ at pair $x_i$.
We stop when ${\bf g}_{\pi}(\mdp_{i+1})\geq {\bf g}_{\pi^\star}(\mdp)$.
An appropriate choice of $x_i$ is one that minimizes the local contribution of this change to the unlikelihood of optimality,
that is 
\beqan
x_i = \argmin\Big\{ \Esp^{\model, \learner}[N_T(x)] \KL_{x}(\mdp,\tmdp^{\texttt{ch}}_i(x)) : x \in \pairs^{\texttt{ch}}(\pi,\mdp)\setminus\{x_{1},\dots,x_{i-1}\}\Big\}\,.
\eeqan

\subsection{Approaching confusing MDPs by local changes}
\label{section_construction_of_confusing}

We are now almost ready to state the algorithm that tries to compute a confusing MDP for a sub-optimal policy $\pi$.
It remains to handle a few special cases that enable to simplify the search procedure.

{\bf Special case.}
When 
$\cS(\pi,\mdp) \cap \cS_{\neq}(\pi,\pi^\star) = \emptyset$,
then $\pi$ coincides with an optimal policy $\pi^\star$ on
$\cS(\pi,\mdp)$. When this policy is unichain,
then it must be that  $\cS(\pi,\mdp)=\cS(\pi^\star,\mdp)$,
and thus the gain of $\pi$ is optimal in $\mdp$. In this special case, 
there is no need to build a confusing instance for $\pi$.

{\bf Saving computations.}
When  considering  a policy  $\pi\in\cV_{\pi^\star}(k)$, then $\cS(\pi,\mdp) \cap \cS_{\neq}(\pi,\pi^\star)$ consists of at most $k$ states.
On the other hand, the size $\cS(\pi,\mdp)\setminus (\cS_{\neq}(\pi,\pi^\star)\cup\cS(\pi^\star,\mdp))$ might be large, the order of $|\cS|$.
Hence, to save computations, one may prioritize to first compute 
$\tmdp^{\texttt{ch}}_i(x)$ for $x=(s,\pi(s))$ from $\cS(\pi,\mdp) \cap \cS_{\neq}(\pi,\pi^\star)$ only,
and scan the rest of the states only if ${\bf g}_{\pi}(\tmdp^{\texttt{ch}}_i(x))\geq {\bf g}_{\pi^\star}(\mdp)$ does not hold for these pairs.
This ``early stopping search" may yield a final MDP having larger KL cost, but possibly save computations. 

\begin{algorithm}[!hbtp]
	\caption{Construction of confusing MDP for policy $\pi$} \label{alg:constructiveconfusingMDP}
	\begin{algorithmic}[1]
		\REQUIRE Policy $\pi$, optimal policy $\pi^\star$ with gain $g^\star$, counts $(\kappa(x))_{x\in\cX}$.
		\STATE Let $\tilde \cX= \emptyset$ \hspace{5.5cm}$\triangleright\triangleright\triangleright$ \texttt{Set where }$\mdp$\texttt{ is modified}
		\IF{$ \cS_{\neq}(\pi,\pi^\star) \cap\cS(\pi,\mdp)=\emptyset$}
		\RETURN $\mdp,\tilde \cX$ \hspace{5.5cm}$\triangleright\triangleright\triangleright$  $\pi$ \texttt{is actually optimal}
		\ENDIF
		\STATE Set $\mdp_0=\mdp$, $i=0$,  $\cX_0= \cX^{\texttt{ch}}(\pi,\mdp)$.
	
		\WHILE{$\bfg_{\pi}(\mdp_{i})<g^\star$ or $\cX_i\neq\emptyset$}
		\STATE Compute $\cX'_i = \{ x\in \cX_i : x=(s,\pi(s)), s\in \cS(\pi,\mdp) \cap \cS_{\neq}(\pi,\pi^\star), {\bf g}_{\pi}(\tmdp^{\texttt{ch}}_i(x))\geq {\bf g}_{\pi^\star}(\mdp)\}$ 
		\IF{$\cX'_i\neq\emptyset$}	
		\STATE Compute 
		$
		x_i = \argmin\Big\{\kappa(x) \KL_{x}(\mdp,\tmdp^{\texttt{ch}}_i(x)) : x \in \cX'_i\Big\}\,.
		$
		\ELSE
			\STATE Compute 
		$
		x_i = \argmin\Big\{\kappa(x) \KL_{x}(\mdp,\tmdp^{\texttt{ch}}_i(x)) : x \in \cX_i\Big\}\,.
		$
		\ENDIF
		\STATE Let $\mdp_{i+1}=\tmdp_i^{\texttt{ch}}(x_i)$,  $\cX_{i+1}=\cX_i\setminus\{x_i\}$, $\tilde \cX = \tilde \cX \cup \{x_i\}$, $i=i+1$.
		\ENDWHILE
				\IF{$\cX_i=\emptyset$}
		\RETURN $\emptyset,\emptyset$ \hspace{5cm}\qquad$\triangleright\triangleright\triangleright$  $\pi$ \texttt{is unconfusingly suboptimal}
		\ELSE
		\RETURN $\mdp_i,\tilde \cX$
		\ENDIF
	\end{algorithmic}
\end{algorithm}

We summarize the procedure to build a confusing MDP in Algorithm~\ref{alg:constructiveconfusingMDP}.
This algorithm returns the confusing instance $\tmdp$ together with the set of pairs $\tilde \cX$ where the MDP differs from $\mdp$, and enables to control the unlikelihood of optimality as
\beqan
\bbU(\pi,\mu,\mdp) \leq \sum_{x\in\tilde \cX} \mu(x) \KL_{x}(\mdp,\tmdp)\,.
\eeqan

Provided $\tilde \cX\neq\emptyset$, this procedure outputs a confusing MDP, not necessarily the most confusing MDP  in the sense of achieving the minimal unlikelihood of optimality due to the considered relaxation.
Specifically, since for policies $\pi\in\cV_{\star}(k)$, performing switch modifications of 
$\mdp$ at $\cS_{\neq}(\pi,\pi^\star)$ (of size $\leq k$) enables to ensure $\pi$ induces the same Markov chain than an optimal policy in 
$\mdp^\dagger$, and such modifications are considered by the algorithm, this ensures the 
construction eventually finds a confusing MDP for such policies.
Indeed, since the algorithm proceed by maximizing the policy improvement at each non-final modification,
the local modification ensures a gain increase not below that of a single pair exchange between $\pi$ and $\pi^\star$, which in turn ensures that procedure eventually stops.

\paragraph{Complexity.} For each such policy $\pi$ with $|\cS_{\neq}(\pi,\pi^\star)|=j$ each construction requires at most $|\cS(\pi,\mdp)|\leq S$ local optimization steps, and possibly less than $j$  when $\cX_i'\neq \emptyset$.
Since there are ${S\choose j}A^j$ policies that differs from $\pi^\star$ at exactly $j$ pairs,
computing a confusing MDP for all policies e.g. in  $\cV_{\star}(k)$ requires at most $\sum\limits_{j=1}^k { S\choose j}A^j S$ local optimization steps, where the factor $S$ can be reduced to at most $j$. 
Of course when $k=S$, the procedure becomes infeasible, since it essentially involves $A^S$ many optimization problems. On the other hand, when $k$ is small with respect to $S$, say  $k=\OH(\log(S))$ this yields a fast procedure.  This motivates to consider scanning only a subset of all policies $\cV_{\star}(k)$.
 
\paragraph{Structure of policies.}
Another motivation for restricting to $\cV_{\star}(k)$ comes from \cite{saber_logarithmic_2024}. In this article, the authors consider the following structural assumption
\begin{property}[$k$-Local policy-improvement]\label{ass:localpi}
	$\forall \pi\notin\optpolicies(\mdp) \exists \pi' \in \cV_\pi(k), {\bf g}_{\pi'}(s)>{\bf g}_{\pi}(s)$.
\end{property}
\noindent
Let us recall that in a generic MDP, the classic policy iteration scheme ensures that at each step, there exists a state in which either the gain ${\bf g}_{\pi_{k+1}}$ or the bias ${\bf b}_{\pi_{k+1}}$ improves.
However, there may be in general several consecutive steps before  there exists a state  in which the gain is improved.
Hence this assumption ensures that a policy-improvement step is possible by searching in local neighborhood of any sub-optimal policy. It is a key step in the regret analysis of the IMED-KD strategy from \cite{saber_logarithmic_2024} to guarantee its correctness. In such situation, one can indeed scan only  $\cV_{\star}(k)$, that is, computing confusing MDPs for such policies only, and ensure a controlled regret. While somewhat different, it provides a complementary perspective on the choice of the set of policies to be considered.

\begin{remark}[Easy and Hard MDPs]
	Remark that all MDPs necessarily satisfy Property~\ref{ass:localpi} for $k=S$.
	Now, intuitively an MDP that satisfies $k$-Local policy-improvement with small value of $k$ (e.g. $1,2$)  is easier to solve than an MDP that only satisfies it for large $k$. That's because for larger $k$, one needs to search for a policy improvement in a larger neighborhood of a sub-optimal policy.
	Interestingly,  the policy-improvement lemma of \cite{puterman2014markov} implies that all ergodic MDPs satisfy the $1$-Local policy-improvement property. Hence, they are in a sense the simplest of all MDPs.
	On the other hand, it is interesting to note that the MDP built in Section~\ref{section_complexity} to prove NP-complete hardness only satisfies $k$-local policy-improvement property for $k=S$, hence is maximally complex in this sense.
\end{remark}

    \section{Conclusion, future work and conjectures}\label{sec:conclusion}
    
In this paper, we provide the regret lower bound in the model dependent setting for communicating Markov decision processes. 
The bound is found as the solution of an optimization problem that combines the mandatory exploration, co-exploration and second order navigational structure that every consistent learner must conform to. 
In general, the bound is $\Sigma_2^\textrm{P}$-complete, i.e., merely checking its value is coNP-complete; This is true when the considered class is discrete, and the question of the complexity remains open for convex classes.   
In many classical classes of Markov decision processes, such as bandits, switching bandits, ergodic classes or deterministic transition classes, our bound coincide with existing ones. 
Moreover, we have discussed a direction to approximate it in a constructive fashion, by exploiting the idea of local modifications of the underlying environment.

\paragraph{Tightness.}
One obvious and immediate concern is the tightness of our bound, that we indeed conjecture to be tight.
More precisely, we conjecture that there exists a consistent learning agent with theoretical regret guarantees matching the lower bound; Curious readers can already take a look at \cite[Chapter~10]{boone_thesis_2024} for a preliminary version. 
We further conjecture that despite the computational hardness of the lower bound, it is possible to design asymptotically optimal algorithms that run in reasonable time.
Once the bound will be proven tight, will raise the classical question of the compatibility of the model dependent setting with the model independent setting.
In multi-armed bandits, \cite{garivier_kl_ucb_switch_2022} design \texttt{KL-UCB-Switch}, reaching asymptotically optimal regret $\regretlb(\model) \log(T)$ and minimax optimal regret $\sqrt{\abs{\actions} T}$ simultaneously; This is the best-of-both-worlds of \cite{bubeck_best_2012}, that we conjecture to fail at Markov decision processes. 
In addition to the best-of-both-worlds, there is the question of how asymptotic the bound is.
For bandits again, \cite{garivier_explore_2018} pointed out that simple algorithms such as Thompson Sampling \cite{thompson_likelihood_1933} manage to beat the lower bound $\regretlb(\model) \log(T)$ even for large time horizons and it is nowadays known \cite{honda_non_asymptotic_2015,garivier_explore_2018} that the second order term is a \emph{negative} $\log \log(T)$ when the support of rewards is bounded.
We conjecture that even in simple cases, the second order term is a \emph{positive} $\log \log(T)$ for Markov decision processes and that the culprit should be co-exploration (\cref{proposition_forced_optimal_pairs}). 
All this is left for future work.

\paragraph{Extensions.}
Another immediate research direction is the generalization of our result to broader settings.
Sticking to finite Markov decision processes, there is the question of dropping the communicating assumption to a weakly communicating assumption or even considering general multi-chain models. 
We conjecture that the lower bound is essentially the same in the weakly communicating setting.
In the multi-chain setting however, the regret actually depends on the initial state and additional elements are required to properly generalize our results there. 
Beyond finite Markov decision processes, there is the question of countable state spaces such as in Queuing Theory, compact action spaces, compact state-action spaces or even parameterized environments such as linear MDPs or MDPs with function approximation. 

\paragraph{Local policy improvement.}
Last but not least, is the question of the approximation of the lower bound.
We find that Property~\ref{ass:localpi} plays an important role in providing a computationally efficient to approximate the bound. Especially, we remark that for ergodic MDPs, which satisfy a $k$-local policy improvement property with $k=1$, our procedure recovers the lower bounds for the ergodic case, while MDPs causing the NP-hardness only satisfy it with $k=S$. 
This motivates studying and characterizing the MDPs obeying an intermediate $k$ in greater depth. Up to our knowledge, very little is known about the intermediate regime of $1<k<S$,
and in particular the MDPs with $k<\log(S)$, which we find of special interest, as the complexity of our procedure directly follows from  Property~\ref{ass:localpi}. Characterizing such classes of MDPs and leveraging such structure is left for future work.

    \section*{Acknowledgments}
    This work has been supported by the French Ministry of Higher Education and Research, the Hauts-de-France region, Inria, the MEL, the I-Site ULNE regarding project R-PILOTE-19-004-APPRENF, and the PEPR FOUNDRY. O. Maillard acknowledges the Inria-Kyoto University Associate Team “RELIANT” and the Scool research group for its
    outstanding working environment.
    
    \bibliographystyle{apalike}
    \bibliography{biblio}

    \clearpage
    \appendix

    \section*{List of notations}\label{app:notations}
    
%
	
    \def\sp#1{\mathrm{sp}\parens{#1}}

    \begin{multicols}{2}
        \begin{description}[itemsep=-.25em]
            \item[$\action$] action
            \item[$\Action_t$] played action at time $t$
            \item[$\actions(\state)$] action space from $s$
            \item[$\alternative(\mdp)$] alternative set of $\model$

            \item[$\beneficial(\policy, \model)$] beneficial set of $\policy$ in $\model$
            \item[$\beneficial_{\jopt}(\policy, \model)$] beneficial-confusing set of $\policy$ in $\model$

            \item[$\confusing(\mdp)$] confusing set of $\model$

            \item[$\diameter(\mdp)$] diameter
            \item[$\ogaps$] Bellman gaps
            \item[$\mathrm{dmin}(-)$] definite minimum (minimal $>0$ entry)

            \item[$\E_{\state_0}^{\model, \learner}$] expectation under $\model$ and $\learner$ initialized at $\state_0$

            \item[$\gain_\policy$] gain vector of policy $\policy$
            \item[$\optimalgain$] optimal gain vector

            \item[$\bias_\policy$] bias vector of policy $\policy$
            \item[$\optbias$] optimal bias vector
            \item[$\History_t$] random history at time $t$
            \item[$\histories$] history space

            \item[$\imeasures(\mdp)$] invariant measures of $\mdp$


            \item[$\regretlb(\mdp)$] regret lower bound with contraction
            \item[$\regretlb_{\jopt}(\model)$] regret lower bound without contraction
            \item[$\KL(-||-)$] Kullback-Leibler div.
            \item[$\KL_\pair(\model||\model^\dagger)$] KL at pair $\pair$ between $\model$ and $\model^\dagger$
            \item[$\kl(-,-)$] KL for Bernoulli distributions

            \item[$L(\History_T; \model, \model^\dagger)$] log-likelihood ratio of observations

            \item[$\reward$] mean reward 
            \item[$\mdp$] reference hidden model
            \item[$\mdp^\dagger$] alternative/confusing model
            \item[$\mdp/\pairs_0$] minor of $\mdp$ by $\mathcal{X}_0$
            \item[$\mdps$] space of MDPs
            \item[$\imeasure$] invariant measure

            \item[$\visits, \visits_T(\state,\action)$] Visit vector

            \item[$\kerneld$] transition kernel 
            \item[$\mathcal{P}$] kernel space
            \item[$\policy, \optpolicy$] policy, optimal
            \item[$\alg$] learning agent
            \item[$\policies$] deterministic policies
            \item[$\rpolicies$] randomized policies
            \item[$\optpolicies(\model)$] gain optimal policies of $\model$
            \item[$\Pr_{\state_0}^{\model, \learner}$] probability under $\model$ and $\learner$ initialized at $\state_0$

            \item[$\rewardd$] reward distributions
            \item[$\Reward_t$] random reward at time $t$
            \item[$\mathcal{R}$] reward space
            \item[$\Reg(-)$] expected regret

            \item[$\state$] state
            \item[$\State_t$] random state at time $t$
            \item[${[}\state{]}$] contracted state
            \item[$\states$] state space
            \item[${[}\states{]}$] contracted state space
            \item[$\sp{-}$] span function

            \item[$t$] time instant
            \item[$T$] time horizon

            \item[$\mathbb{U}(\policy, \imeasure, \model)$] unlikelihood of optimality



            \item[$\pair$] state-action pair
            \item[$\Pair_t$] pair at time $t$
            \item[$\pairs$] state-action pairs
            \item[$\suboptimalpairs(\model)$] sub-optimal pairs $\model$
            \item[$\weakoptimalpairs(\model)$] weakly optimal pairs $\model$
            \item[$\optpairs(\model)$] optimal pairs of $\model$
        \end{description}
    \end{multicols}

    \clearpage

    \numberwithin{equation}{section}
    \numberwithin{definition}{section}
    \numberwithin{theorem}{section}

    \section{Supplementary proofs for the regret lower bound}\label{app:lowerbound}
    
\subsection{On self-normalized expected deviations of the empirical mean}

In this paragraph, we provide a technical lemma for the proof of \cref{proposition_expected_minor}.

\begin{lemma}
\label{lemma_self_normalized_expected_deviations}
    Consider $X_n \in [0, 1]$ a family of r.v.~with $\E[X_n|\mathcal{F}_n] = \mu$.
    Let $\hat{\mu}_n := \frac 1n \sum_{k=1}^n X_k$ their empirical mean.
    Let $N$ a random variable of support $\set*{0, 1, \ldots, T}$ where $T \ge 1$ is a fixed scalar.
    Then, for all $\epsilon > 0$, 
    \begin{equation*}
        \E\brackets*{N\parens*{\hat{\mu}_N - \mu}}
        \le
        \epsilon \parens*{
            \E[N] + \log(1+T)
        } + \sqrt{
            \E \brackets*{
                N \log \parens*{
                    2 \vee
                    \tfrac{3\sqrt{1+N}\log(1+T)}{2\epsilon^3\E[N]}
                }
            }
        } + 1.
    \end{equation*}%
\end{lemma}
\begin{proof}
    Let $\delta > 0$ that shall be tuned later.
    By Hoeffding's Lemma, $X_n$ is conditionally $\sigma$-subgaussian for $\sigma = \tfrac 12$. 
    By a time-uniform Azuma-Hoeffding's inequality \cite[Lemma~5]{bourel_tightening_2020}, for all $m \ge 1$, 
    \begin{equation}
        \label{equation_self_normalized_expected_deviations_1}
        \Pr \parens*{
            \exists m \le n \le T,
            \abs{\hat{\mu}_n - \mu}
            \ge
            2 \sigma \sqrt{\frac{\log(T\sqrt{1+T})}m}
        }
        \le
        \frac 1T
        .
    \end{equation}
    Setting $m = m_\epsilon := (\tfrac{2\sigma}\epsilon)^2 \log(T \sqrt{1 + T})$, we have $\Pr(\exists m_\epsilon \le n \le T, \abs{\hat{\mu}_n - \mu} \ge \epsilon) \le \frac 1T$.
    The target expectation is split into two.
    Denoting $f(N) := N(\hat{\mu}_N - \mu)$, we write:
    \begin{equation}
        \E[N(\hat{\mu}_N - \mu)]
        =
        \E[f(N) \indicator{N \ge m_\epsilon}] 
        +
        \E[f(N) \indicator{N < m_\epsilon}] 
        .
    \end{equation}
    We start by controlling $\E[f(N) \indicator{N \ge m_\epsilon}]$.
    By construction of $m_\epsilon$, we have:
    \begin{align*}
        \E[f(N) \indicator{N \ge m_\epsilon}] 
        & \le
        \E[f(N) \indicator{N \ge m_\epsilon} \indicator{\abs{\hat{\mu}_N - \mu} < \epsilon}] 
        +
        \E[f(N) \indicator{N \ge m_\epsilon} \indicator{\abs{\hat{\mu}_N - \mu} \ge \epsilon}] 
        \\
        & \le 
        \epsilon \E[N] + T \Pr \parens{
            \exists m \le n \le T,
            \abs{\hat{\mu}_n - \mu} \ge \epsilon
        }
        \\
        \text{(by \eqref{equation_self_normalized_expected_deviations_1})} & \le 
        \epsilon \E[N] + 1
        .
    \end{align*}
    We continue with $\E[f(N)\indicator{N < m_\epsilon}]$. 
    Denote $\event_\delta := (\forall n \ge 1, n (\hat{\mu}_n - \mu)^2 \le 4 \sigma^2 \log(\sqrt{1+n}/\delta))$. 
    By a time-uniform Azuma-Hoeffding's inequality again, this good event has probability at least $1 - \delta$. 
    We obtain:
    \begin{align*}
        \E[f(N) \indicator{N < m_\epsilon}]
        & =
        \E[f(N) \indicator{N < m_\epsilon} \indicator{\event_\delta}]
        +
        \E[f(N) \indicator{N < m_\epsilon} \indicator{\event_\delta^c}]
        \\
        & \le 
        2 \sigma \E \brackets*{
            \sqrt{N \log\parens*{\tfrac{\sqrt{1+N}}\delta}}
        }
        + 
        \delta m_\epsilon
        \\
        & \overset{(*)} \le
        2 \sigma \sqrt{\E \brackets*{
            N \log\parens*{\tfrac{\sqrt{1+N}}\delta}
        }}
        + 
        \delta m_\epsilon
        \\
        & \equiv 
        2 \sigma \sqrt{\E \brackets*{
            N \log\parens*{\tfrac{\sqrt{1+N}}\delta}
        }}
        + 
        \delta \parens{\tfrac{2\sigma}\epsilon}^2 \log\parens*{T \sqrt{1+T}}
    \end{align*}
    where $(*)$ follows from Jensen's inequality.
    Set $\delta := \frac{\epsilon^3}{6\sigma^2} \frac{\E[N]}{\log(1+T)}$ and plug everything together.
\end{proof}

Applied to sequential control, where $N_T$ is the number of triggers up to time $T$, we see that when $\E[N_T] + \E[\log(N_T)] = \Omega(\log(T))$, then $\E[N_T(\hat{\mu}_{N_T} - \mu)] = \oh(\E[N])$.

\subsection{Comments on less asymptotic regret lower bounds}

The techniques used to establish \cref{theorem_lower_bound} are all very asymptotic. 
This is no surprise, because the whole theory is built on a very asymptotic notion of consistency (\cref{definition_consistency}), that only asks that learners eventually converge to optimal play.
The asymptotic speed of that convergence $\oh(T^\eta)$ has nothing to do the possibly arbitrarily long burn-in time before the bound $T^\eta$ starts to barely holds. 
This allows for algorithms that are asymptotically optimal in theory, but impractical because of their immense burn-in times.
This is reproached by \cite{burnetas_optimal_1997} to the solution of \cite{agrawal_asymptotically_1988}, until their very solution was itself criticized for its impracticability by \cite{pesquerel_imed_rl_2022} two decades later. 
The lower bound of \cite{lai_asymptotically_1985} has also been pointed out for its over-asymptotic nature by \cite{garivier_explore_2018}, with Thompson Sampling \cite{thompson_likelihood_1933} beating the asymptotic lower bound $\regretlb(\model) \log(T)$ even for large time horizons. 

This kind of story is common and there have been many attempts at patching the asymptotic inclination of theory: the trend for finite time bounds, second order terms of asymptotic lower bounds, or straight up modifications of the learning framework. 

Properly addressing the asymptotic nature of our own \cref{theorem_lower_bound} goes way beyond the scope of this paper, but we thought important to leave a few elements behind. 
Following the direction of finite time bounds, we first introduce a notion of consistency that is less asymptotic than \cref{definition_consistency}. 

\begin{definition}[Timely-consistent learner]
    Let $\eta : \N \to [0, 1]$ a function satisfying $\eta(t) = \oh(1)$. 
    A learning agent $\learner$ is said \strong{$\eta$-timely-consistent} on $\models$, if for all $\model \in \models$, 
    \begin{equation*}
        \exists T_\model \in \N,
        \forall T \ge T_\model,
        \sup_{\pair : \ogaps(\pair: \model) > 0}
        \E_{\state_0}^{\model, \learner} \brackets*{
            \frac{\visits_{T+1}(\pair)}{T}
        }
        \le
        \eta(T)
        .
    \end{equation*}
\end{definition}
Observe that this definition is expressed in terms of visits rather than of regret. 
This is done so in order to streamline the argument behind the regret lower bound. 
Consider an $\eta$-timely-consistent learner $\learner$ and let us mimic the computations done for the simple example of discriminating random variable in \cref{section_rejecting_alternative}. 
Let $\model^\dagger$ such that $\wkoptpairs(\model^\dagger) \cap \wkoptpairs(\model) = \emptyset$ and let $U := \sum_{\pair \in \wkoptpairs(\model)} \visits_{T+1}(\pair)$. 
By assumption and provided that $T \ge \max\set{T_\model, T_{\model^\dagger}}$, we directly obtain
\begin{equation*}
    \E_{\state_0}^{\model, \learner} [\visits_{T+1}(\pair)] \ge (1 - \abs{\pairs} \eta(T)) T 
    \quad\text{and}\quad
    \E_{\state_0}^{\model^\dagger, \learner} [\visits_{T+1}(\pair)] \le \abs{\pairs} \eta(T) T 
    .
\end{equation*}
Setting $\event := (U \le \frac 12 T)$, we deduce that $\Pr_{\state_0}^{\model,\learner}(\event) \le 2 \abs{\pairs} \eta(T)$ and $\Pr_{\state_0}^{\model^\dagger,\learner}(\event) \ge 1 - 2 \abs{\pairs} \eta(T)$. 
For $\eta(T) \le \frac 1{2\abs{\pairs}}$ and invoking \cref{corollary_likelihood_event}, we conclude that:
\begin{equation*}
    \E_{\state_0}^{\model, \learner} \brackets*{
        \sum_{\pair \in \pairs}
        \visits_{T+1}(\pair) \KL_\pair(\model||\model^\dagger)
    }
    \ge
    \kl\parens[\Big]{2 \abs{\pairs} \eta(T), 1 - 2\abs{\pairs}\eta(T)}
    .
\end{equation*}
Following the recipe described in \cref{section_visit_rates_of_optimal} to convert the above into a regret lower bound, we normalize by $T$ and invoke \cref{proposition_pseudo_regret} to obtain the finite time bound below:

\begin{theorem}
    Let $\learner$ an $\eta$-timely-consistent learner. 
    For all $T \ge T_\model$, then $\Reg(T; \model, \learner, \state_0) + \mathrm{sp}(\optbias(\model))$ is lower-bounded by

    \medskip
    \noindent
    \resizebox{\linewidth}{!}{
    $
        \displaystyle
        T \inf_{\imeasure \in \R_+^\pairs} \set*{
            \sum_{\pair \in \pairs}
            \imeasure(\pair) \ogaps(\pair; \model)
            :
            \forall \model^\dagger \in \mathrm{Cnf}_{\wopt}(\model),
            \sum_{\pair \in \pairs} \imeasure(\pair) \KL_\pair(\model||\model^\dagger) 
            \ge
            \indicator{T_{\model^\dagger} < T}
            \frac{\kl(2\abs{\pairs}\eta(T), 1 - 2\abs{\pairs}\eta(T))}{T}
        }
    $}

    \medskip
    \noindent
    where $\mathrm{Cnf}_{\wopt}(\model) := \set[\big]{\model^\dagger \in \alternative(\model): \wkoptpairs(\model^\dagger) \cap \wkoptpairs(\model) = \emptyset}$. 
\end{theorem}

\subsection{Supplementary content for the \texttt{Switch} property}

In this paragraph, we provide a few proofs of the \cref{sec:contruction}, especially by providing a proof of \cref{prop:localsimplification}, of which the statement is recalled below. 

\bigskip
\par
\noindent
\strong{\cref{prop:localsimplification}.}
\textit{
	Let $x=(s,\pi(s))$ and assume that $\cC_+^{\texttt{sw}}(x)\neq \emptyset$.%
    \vspace{-0.33em} 
    \begin{enum}
        \item 
            If $\pi^\star$ is ergodic in $\mdp$,  then \texttt{Switch} 	holds at $x$.
        \item 
	        If $\pi^\star$ is unichain in $\mdp$, then 	\texttt{Switch}
	holds at $x$ except when  $s\notin \cS(\pi^\star,\mdp)$.
	    \item
	        If $\pi^\star$ is multichain in $\mdp$, then \texttt{Switch} holds at $x$ when $s\in \cS(\pi,\mdp) \cap \cS(\pi^\star,\mdp)$ and does not hold when $s \notin \cS(\pi^\star,\mdp)$.
    \end{enum}
}

\begin{proof}[Proof of \cref{prop:localsimplification}]
	Let us denote $\pi=\pi^\star_x$. 
	Before proceeding, with discuss the following few cases:
	
    \strong{Case a.}
    Assume that  $s \notin \cS(\pi,\mdp)$. 
	Let us remind that $\pi$ coincides with $\pi^\star$  on $\cS\setminus\{s\}$, hence in particular on $\cS(\pi,\mdp)$.
	This means that on the set of recurrent states $\cS(\pi,\mdp)$ of $\pi$, $\pi$ behaves as $\pi^\star$. Since $\cS(\pi,\mdp)$ is a closed set,
	this means that it is also a recurrent set according to $\pi^\star$, which shows $\cS(\pi,\mdp)\subset \cS(\pi^\star,\mdp)$.
	Now when $\pi^\star$ is unichain, $\cS(\pi,\mdp)= \cS(\pi^\star,\mdp)$. In particular $s \notin \cS(\pi^\star,\mdp)$, and since $\pi$ coincides with $\pi^\star$, they then have same gain hence $\pi$ is already optimal in $\mdp$, which contradicts assumption $\pi \notin \optpolicies(\mdp)$.
	
    \strong{Case b.}
    Now assume instead that  $s \in \cS(\pi,\mdp)$.  
	For a switch MDP $\tmdp\in\cC_x(\mdp)$ to be a confusing instance, the gain of $\pi$  must be larger than that of $\pi^\star$, without modifying the MDP on $\optpairs(\mdp)$.
	
    \strong{Case b.1}
    As a result, when $s \notin \cS(\pi^\star,\mdp)$, then switching the effect of action $a$ in $s$
	and increasing the reward of this pair cannot modify the gain in $\mdp^\dagger$ beyond the gain of $\pi^\star$, hence $\switch(x)$ cannot hold in such cases. 
	A special case is when $\cS(\pi^\star,\mdp)$ is further not reachable from $s$ under $\pi$. In particular, when $\pi$ is unichain, this means no state from $\cS(\pi,\mdp)$ communicates with $\cS(\pi^\star,\mdp)$ under $\pi$, which means $x$ is a bottleneck.
	
    \strong{Case b.2.}
    It remains to deal with the case when $s \in \cS(\pi,\mdp)\cap \cS(\pi^\star,\mdp)$. 
	In this case, switching the effect of $a$ in $\mdp^\dagger$ ensures the gain of the policy is not smaller than that of $\pi^\star$ in $\mdp$, and increasing the mean reward ensures it becomes larger. Indeed, for $\mdp^\dagger \in \cC_x(\mdp)$, since  the Markov chain induced by $\pi^\star_x$ in $\mdp^\dagger$ coincides with the Markov chain induced by $\pi^\star$ in $\mdp$, this means 
	choosing distribution ${\bf r}'$  with mean  $m'>{\bf m}(s, \pi^\star(s))$ at pair $x$ ensures that ${\bf g}^\dagger_{\pi^\star_x}(s_1)>{\bf g}^\dagger_{\pi_\star}(s_1)$ for all $s_1$. 
	Hence $\mdp^\dagger \in \conf(\mdp)\cap \cB(\pi^\star_x,\mdp)$	 and $\switch(x)$ holds. 
	
    For the case (1) in \cref{prop:localsimplification}, we note that by ergodicity of $\pi^\star$, $\cS(\optpairs(\mdp))=\cS$. 
    Now  $s\notin\cS(\pi,\mdp)$, would imply $\pi \notin \optpolicies(\mdp)$ by a) which is excluded by assumption, hence we are in the case \strong{b.2} and $\switch(x)$ holds.
    Regarding case (2) in \cref{prop:localsimplification}, then $s\notin\cS(\pi,\mdp)$ is excluded by a) hence we conclude by \strong{b.1} and \strong{b.2}.
    Regarding the multichain case, the only difference is that $s\notin\cS(\pi,\mdp)$ may not contradict that $\pi$ is sub-optimal. 
    But the conclusion regarding \strong{b.1} and \strong{b.2} still holds.
\end{proof}

    \section{Inevitable sets and details on informational constraints}
    \allowdisplaybreaks

\label{appendix_inevitable}

This section provides technical details on the proofs of informational constraints, especially \cref{proposition_rejecting_alternative}, \cref{proposition_forced_optimal_pairs} and \cref{corollary_rejecting_confusing}.

\subsection{Inevitable sets and the inevitability lemma}

A key idea, invoked at several places in the proofs of these statements, is the notion of \strong{inevitable set}.
Inevitable sets are sets of pairs that every policy of a reference set is forced to go through infinitely many times, i.e., linearly often and regardless of the initial state. 

\begin{definition}
\label{definition_inevitable_set}
    A set of pairs $\pairs_c \subseteq \pairs$ is said \strong{inevitable} relatively to $\optpolicies(\model)$ if, for all $\policy \in \optpolicies(\model)$ and all initial state $\state_0 \in \states$, we have $\Pr_{\state_0}^{\model, \policy}\set{\forall n, \exists m \ge n: \Pair_m \in \pairs_c} = 1$.
\end{definition}

Although inevitable sets could be described to different sets than $\optpolicies(\model)$, the generalization won't be required, and we will speak of \emph{inevitable sets} rather than \emph{inevitable sets relatively to $\optpolicies(\model)$} in the sequel.
By standard Markov chain theory \cite{levin_markov_2017}, recurrent and positive recurrent are equivalent when the number of states is finite, hence the condition ``$\forall \state_0 \in \states$, $\Pr_{\state_0}^{\model, \policy}\set{\forall n, \exists m \ge n: \Pair_m \in \pairs_c} = 1$'' can be changed to ``$\forall \state_0 \in \states$, $\E_{\state_0}^{\model, \policy}\brackets{\inf\set{t \ge 2: \Pair_t \in \pairs_c}} < \infty$'' or again to ``$\forall \state_0 \in \states$, $\E_{\state_0}^{\model, \policy}[\sum_{t=1}^T \indicator{\Pair_t \in \pairs_c}] = \Omega(T)$''.

Inevitability requires that, for every $\policy \in \optpolicies(\model)$, $\pairs_c$ is visited linearly often.
This point-wise property is converted to a more uniform one via the remarkable result below, actually stating that in probability, the number of visits of an inevitable set $\pairs_c$ satisfies $\sum_{\pair \in \pairs_c} \visits_{T+1}(\pair) = \Omega(T - \Reg(T; \model, \learner))$. 

\begin{lemma}[Inevitability lemma]
\label{lemma_inevitability}
    Let $\model \in \models$ and assume that $\pairs_c \subseteq \pairs$ is inevitable.
    There exist constants $\epsilon_c, \diameter_c > 0$ such that, whatever the learning agent $\alg$ and the initial state $\state_0$, we have:
    \begin{equation*}
        \forall u \ge 0,
        \quad
        \Pr_{\state_0}^{\model, \alg} \parens*{
            \sum_{x\in\pairs_c} N_{T+1}(\pair)
            +
            \diameter_c \sum_{x \in \suboptimalpairs(\model)} \visits_{T+1}(\pair)
            \le
            \epsilon_c T - u
        }
        \le 
        \exp \parens*{
            - \frac{2 u^2}{T \diameter_c^2}
        }
        .
    \end{equation*}
    Moreover, $\E_{\state_0}^{\model, \learner}[\sum_{\pair \in \pairs_c} \visits_{T+1}(\pair)] + \diameter_c \E_{\state_0}^{\model, \learner}[\sum_{\pair \in \suboptimalpairs(\model)} \visits_{T+1}(\pair)] \ge \epsilon_c T - \diameter_c$. 
\end{lemma}
\begin{proof}
    The idea is to construct the optimal policy of that visits $\pairs_c$ as rarely as possible. 
    Consider the reward function $f(\pair) := - \indicator{\pair \in \pairs_c}$ and consider $\model_f|_{\weakoptimalpairs(\model)}$, a copy of $\model$ restricted to $\weakoptimalpairs(\model)$ and with reward function $f$. 
    Let $\policy_f$ a bias optimal policy of $\model_f|_{\weakoptimalpairs(\model)}$; It is also a policy of $\model$ since $\states(\weakoptimalpairs(\model)) = \states$.
    Moreover, $\policy \in \optpolicies(\model)$ because $\policy$ only picks pairs such that $\ogaps(\pair; \model) = 0$, hence is guaranteed to have optimal gain via \cref{proposition_pseudo_regret}.
    Let $\gain_f, \bias_f$ and $\gaps_f$ its gain, bias and gap functions on $\model$ under the reward function $f$. 
    By construction of $\policy_f$, we have $\gaps_f (\pair) \ge 0$ for all $\pair \in \weakoptimalpairs(\model)$. 
    Let $\epsilon_c := - \max_{s \in \mathcal{S}} \gain_f(s)$, and denote $\diameter_c := \max\set{\vecspan{\bias_f}, \max_{\pair \in \pairs} \abs{\gaps_f(\pair)}} < \infty$.
    Since $\policy \in \optpolicies(\model)$ and $\pairs_c$ is inevitable, we have $\epsilon_c > 0$.
    Now,
    \begin{align*}
        \sum_{\pair \in \pairs_c} \visits_{T+1}(\pair)
        & = 
        - \sum_{t=1}^{T} f(\Pair_t)
        \\
        & \overset{(*)}= 
        - \sum_{t=1}^{T} \parens*{
            \gain_f(S_t) + \parens*{\unit_{S_t} - \kernel(\Pair_t)} \bias_f - \gaps_f(\Pair_t)
        }
        \\ 
        & \overset{(\S)}\ge
        \epsilon_c T 
        + \sum_{t=1}^{T} \parens*{\unit_{S_{t+1}} - \kernel(\Pair_t)} \bias_f
        - \diameter_c \parens*{1 + \sum_{\pair \in \suboptimalpairs(\model)} N_{T+1}(\pair)}
    \end{align*}
    where $(*)$ invokes the Poisson equation $\gain_f(s) + \bias_f(s) = f(s,a) + \kernel(s,a) \bias_f + \gaps_f(s,a)$,
    and $(\S)$ that $\gaps_f(\model) \ge 0$ for $\pair \in \weakoptimalpairs(\model)$.
    By Azuma-Hoeffding's inequality, the MDS term satisfies:
    \begin{equation*}
        \forall u \ge 0,
        \quad
        \Pr_{\state_0}^{\model, \alg} \parens*{
            \sum_{t=1}^T \parens*{\unit_{S_{t+1}} - \kernel(\Pair_t)} \bias_f
            \le - u
        }
        \le 
        \exp\parens*{
            - \frac{2u^2}{T\diameter_c^2}.
        }
    \end{equation*}
    This provides the result in probability. 
    The result in expectation is immediately obtained by using that $\E_{\state_0}^{\model, \learner}[\sum_{t=1}^T (\unit_{\State_{t+1}} - \kernel(\Pair_t)) \bias_f] = 0$.
\end{proof}

\paragraph{Discussion.}
In \cref{lemma_inevitability}, what is important is that $\epsilon_c \ne 0$. 
In a few scenarios, for instance when $\pairs_c = \optpairs(\model)$, one can check that $\epsilon_c = 1$. 
This is actually the idea behind \cref{lemma_time_outside_optimal_pairs}, that can be seen as a simplified version of \cref{lemma_inevitability} when $\pairs_c = \optpairs(\model)$. 
Overall, \cref{lemma_inevitability} is immensely useful in the proofs of \cref{proposition_rejecting_alternative} and \cref{proposition_forced_optimal_pairs}.

\subsection{Proof of \cref{lemma_discriminating}}

We start by providing a proof of \cref{lemma_discriminating} which is essential to \cref{proposition_rejecting_alternative}.
We recall its statement below.

\bigskip 
\par\noindent
\strong{\cref{lemma_discriminating}.}
\textit{
    Let $\model \in \models$, pick $\model^\dagger \in \alternative(\model)$
    and let $\pairs_c := \set{\pair \in \optpairs(\model^\dagger) : \ogaps(\pair; \model) > 0}$. 
    There exist constants $\epsilon^\dagger_c, \diameter^\dagger_c > 0$ such that, whatever the learning agent $\learner$ and the initial state $\state_0$, we have:
    \begin{equation*}
        \forall u \ge 0,
        \quad
        \Pr_{\state_0}^{\model^\dagger, \learner} \parens*{
            \sum_{x\in\pairs_c} N_{T+1}(\pair)
            +
            \diameter^\dagger_c \sum_{x \in \suboptimalpairs(\model)} \visits_{T+1}(\pair)
            \le
            \epsilon^\dagger_c T - u
        }
        \le 
        \exp \parens*{
            - \frac{2 u^2}{T \diameter^{\dagger 2}_c}
        }
        .
    \end{equation*}
}
\begin{proof}
    We show that $\pairs_c$ is inevitable in $\model^\dagger$. 
    The conclusion will follow by \cref{lemma_inevitability}.

    Let $\policy^\dagger \in \optpolicies(\model^\dagger)$.

    We say that a state $\state_\opt$ is \emph{reachable} from $\state$ in $\model' \in \set{\model, \model^\dagger}$ if $\Pr_{\state}^{\model',\policy^\dagger}\set{\exists n: \State_n = \state_\opt} > 0$. 

    Assume, by contradiction, that there is $\state \in \states$ such that $\Pr_{\state}^{\model^\dagger, \policy^\dagger}\braces{\forall n, \exists m \ge n: \Pair_m \in \pairs_c} < 1$; There must exist $\state^\dagger \in \states$ reachable from $\state$ in $\model^\dagger$ such that $\Pr_{\state^\dagger}^{\model^\dagger,\policy^\dagger}\set{\forall n, \exists m: \Pair_m \in \pairs_c} = 0$. 
    Let $\states^\dagger$ the collection of all states that are reachable from $\state^\dagger$ in $\model^\dagger$. 
    Remark that every state of $\states^\dagger$ is reachable (under $\policy^\dagger$) is reachable from any other of $\states^\dagger$ in $\model^\dagger$. 
    We see that for all $\state \in \states^\dagger$ and all $\action \in \actions(\state)$ with $\policy^\dagger(\action|\state) > 0$, we have $\ogaps(\state, \action|\model) = 0$. 

    From there, we construct $\policy \in \optpolicies(\model^\dagger) \cap \optpolicies(\model)$, as the policy given by:
    \begin{equation*}
        \policy(\action|\state)
        :=
        \begin{cases}
            \policy^\dagger(\action|\state)
            & \text{if~} \state \in \states^\dagger;
            \\
            \frac 1{\abs{\actions(\state)}}
            & \text{if~} \state \notin \states^\dagger.
        \end{cases}
    \end{equation*}
    Because it is uniform outside of $\states^\dagger$, $\states^\dagger$ is reachable under $\policy$ from every state both in $\model$ and $\model^\dagger$. 
    Now, observe that for all $\state \in \states^\dagger$ and $\action \in \actions(\state)$ with $\policy(\action|\state) > 0$, we have $\supp(\kernel^\dagger(\state,\action)) \subseteq \states^\dagger$. 
    Since $\kernel \ll \kernel^\dagger$, we similarly have $\supp(\kernel(\state,\action)) \subseteq \states^\dagger$. 
    It follows that the recurrent states of $\policy$ satisfy $\states_{\policy}(\model^\dagger) = \states^\dagger$ and $\states_{\policy}(\model) \subseteq \states^\dagger$. 
    In other words, both on $\model$ and $\model^\dagger$, the iterates of $\policy$ are eventually confined within $\states^\dagger$, whatever the initial state.
    Since $\states_{\policy}(\model^\dagger) = \states^\dagger$ which is a component of recurrent states of $\policy^\dagger$, and that $\policy^\dagger$ satisfies $\gain_{\policy^\dagger}(\model^\dagger) = \optgain(\model^\dagger)$, it follows that $\gain_{\policy}(\model^\dagger) = \optgain(\model^\dagger)$.
    Since $\states_{\policy}(\model) \subseteq \states^\dagger$ and that every pair $\pair$ that $\policy$ plays on $\states^\dagger$ satisfies $\ogaps(\policy; \model) = 0$, it follows by \cref{proposition_pseudo_regret} that $\gain_{\policy}(\model) = \optgain(\model)$.
    So $\policy \in \optpolicies(\model^\dagger) \cap \optpolicies(\model)$; A contradiction with $\model^\dagger \in \alternative(\model)$. 
\end{proof}

\subsection{Proof of \cref{proposition_forced_optimal_pairs}}

To conclude this section, we provide a proof of \cref{proposition_forced_optimal_pairs}.

\bigskip 
\par\noindent
\strong{\cref{proposition_forced_optimal_pairs}.}
\textit{
    Let $\model \in \models$ and fix $\pair_\opt \in \optpairs(\model)$. 
    Assume that, for all $\epsilon > 0$, there exists $\model' \in \models$ such that (1) $\model'$ and $\model$ only differ with $\rewardd(\pair_\opt) \ne \rewardd'(\pair_\opt)$, (2) $\reward(\pair_\opt) < \reward'(\pair_\opt)$ and (3) $\KL(\rewardd(\pair_\opt)||\rewardd'(\pair_\opt)) < \epsilon$.
    Then, for every consistent learner $\learner$ and regardless of the initial state $\state_0 \in \states$, 
    \begin{equation*}
        \E_{\state_0}^{\model, \learner}\brackets*{\visits_{T+1}(\pair_\opt)}
        =
        \omega \parens*{\log(T)}
        .
    \end{equation*}
}

\begin{proof}
    Let $\epsilon > 0$ small enough (to be chosen later) and pick $\model' \in \models$ as the above.
    The idea of the proof is to show that $\pairs_c := \set{\pair_\opt}$ is an inevitable set in $\model'$, i.e., that every optimal policy of $\model'$ plays $\pair_\opt$ infinitely often regardless of the initial state. 
    Let $\delta := \reward'(\pair_{\opt}) - \reward(\pair_{\opt}) > 0$. 
    Given a policy $\policy \in \policies$, we denote $\imeasure_\policy(\pair|\state_0) := \lim_{T \to \infty} \frac 1T \E_{\state_0}^{\model, \policy}[\sum_{t=1}^T \indicator{\Pair_t = \pair}]$ the asymptotic average number of plays of $\pair$ under $\policy$ in $\model$ starting from $\state_0$.
    Since $\model$ and $\model'$ have the same transition kernel, we obtain:
    \begin{equation*}
        \gain_{\policy}(\state_0; \model') = \gain_{\policy}(\state_0; \model) + \imeasure_\policy(\pair_\opt|\state_0) \delta.
    \end{equation*}
    Provided that $\delta < \gaingap(\model) := \min \set{\norm{\optgain(\model) - \gain_\policy(\model)}_\infty : \policy \notin \optpolicies(\model)}$, it follows that optimal policies of $\model'$ are necesseraly optimal in $\model$.
    Then, optimal policies of $\model'$ are found as the optimal policies of $\model$ that maximize $\imeasure_\policy(\pair_{\opt}|\state_0)$, i.e., such that $\imeasure_\policy(\pair_{\opt}|\state_0) = \max_{\policy' \in \optpolicies(\model)} \imeasure_{\policy'}(\pair_{\opt}|\state_0)$. 
    Because $\pair_{\opt} \in \optpairs(\model)$, it follows that this maximum is positive and that $\pair_{\opt}$ is visited linearly often under every optimal policy of $\model'$; Hence, that $\pairs_c$ is inevitable. 
    This all is provided that $\delta < \gaingap(\model)$.
    This assumption is met as soon as $\KL(\rewardd(\pair_\opt)||\rewardd'(\pair_\opt)) \to 0$ implies $\reward'(\pair_{\opt}) \to \reward(\pair_{\opt})$, which is true in particular when rewards have support within $[0, 1]$. 

    So, $\set{\pair_{\opt}}$ is inevitable in $\model'$. 
    By \cref{lemma_inevitability}, there exists $\epsilon'_c, \diameter'_c > 0$ such that:
    \begin{equation*}
        \forall u \ge 0,
        \quad
        \Pr_{\state_0}^{\model', \learner} \parens*{
            \visits_{T+1}(\pair_{\opt})
            + \diameter'_c \sum_{\pair \in \suboptimalpairs(\model)} \visits_{T+1}(\pair)
            \le
            \epsilon'_c T - u
        }
        \le
        \exp\parens*{-\frac{2u^2}{T \diameter'^2_c}}
        .
    \end{equation*}
    The remaining of the proof is essentially similar to the one of \cref{proposition_rejecting_alternative}.
    Let $\alpha := \vecspan{\optbias(\model')}$, introduce $\varphi(T) := \sqrt{(\Reg(T; \model', \learner, \state_0) + \alpha)/T}$ and $\psi(T) := \diameter'_c \dmin(\ogaps(\model'))^{-1} \sqrt{T (\Reg(T; \model', \learner, \state_0) + \alpha)}$.
    Combining the above inequality with Markov's inequality, we find:
    \begin{equation*}
        \forall u \ge 0,
        \quad
        \Pr_{\state_0}^{\model', \learner} \parens*{
            \visits_{T+1}(\pair_{\opt})
            \le
            \epsilon'_c T - \psi(T) - u
        }
        \le
        \varphi(T) + 
        \exp\parens*{-\frac{2u^2}{T \diameter'^2_c}}
        .
    \end{equation*}

    To conclude, we aim at invoking \cref{corollary_likelihood_jensen,corollary_likelihood_jensen} with $U := \frac 1{1 + \visits_{T+1}(\pair_{\opt})}$.
    We start by upper-bounding $\E_{\state_0}^{\model', \learner}[U]$.
    We have:
    \begin{equation*}
    \begin{aligned}
        \E_{\state_0}^{\model', \learner}\brackets*{
            U
        }
        \equiv
        \E_{\state_0}^{\model', \learner}\brackets*{
            \frac1{1 + \visits_{T+1}(\pair_{\opt})}
        }
        & \le
        \inf_{v \ge 0} \braces*{
            \frac 1{1+v}
            +
            \Pr_{\state_0}^{\model', \learner} \parens*{\visits_{T+1}(\pair_{\opt}) \le v}
        }
        \\
        & \le
        \inf_{v \ge 0} \braces*{
            \frac 1{1+v}
            +
            \varphi(T) + 
            \exp\parens*{-\frac{
                2\parens*{\epsilon'_cT - \psi(T) - v}^2
            }{T \diameter'^2_c}}
        }
        \\ 
        & \overset{(\dagger)}\le
        \frac 1{1+\frac 12 \epsilon'_c T}
        +
        \varphi(T) + 
        \exp\parens*{-\frac{
            2\parens*{\frac 12\epsilon'_cT - \psi(T)}^2
        }{T \diameter'^2_c}}
        \\
        & \le
        \frac 1{1+\frac 12 \epsilon'_c T}
        +
        \varphi(T) + 
        \exp\parens*{
            -\frac {\epsilon'^2_c T}{2 \diameter'^2_c}
            + \frac{2 \epsilon'_c \psi(T)}{\diameter'^2_c}
        }
    \end{aligned}
    \end{equation*}
    where $(\dagger)$ follows by setting $v := \frac 12 \epsilon'_c T$. 
    By strong consistency, $\psi(T) = \oh(T)$ and $\log \varphi(T) = - \frac 12 \log(T) + \oh(\log(T))$. 
    We conclude that $\log(\E_{\state_0}^{\model', \learner}[U]) \le - \frac 12 \log(T) + \oh(\log(T))$.

    So, applying \cref{corollary_likelihood_jensen}, we obtain:
    \begin{equation*}
    \begin{aligned}
        \epsilon ~ \E_{\state_0}^{\model, \learner}[\visits_{T+1}(\pair_\opt)]
        & \ge
        \E_{\state_0}^{\model, \learner}[\log(U)]
        + 
        \log \parens*{
            \frac 1{\E_{\state_0}^{\model', \learner}[U]}
        }
        \\
        & =
        - \E_{\state_0}^{\model, \learner}[\log(1 + \visits_{T+1}(\pair_{\opt}))]
        - \log \parens*{\E_{\state_0}^{\model', \learner}\brackets*{\frac 1{1 + \visits_{T+1}(\pair_\opt)}}}
        \\
        & \overset{(\dagger)}\ge
        - \log\parens*{\E_{\state_0}^{\model, \learner}\brackets*{1 + \visits_{T+1}(\pair_\opt)}}
        + \frac 12 \log(T) + \oh\parens*{\log(T)}
    \end{aligned}
    \end{equation*}
    where $(\dagger)$ follows by Jensen's inequality, 
    Since $\log(\E_{\state_0}^{\model, \learner}[1 + \visits_{T+1}(\pair_{\opt})])$ is negligible in front of $\E_{\state_0}^{\model, \learner}[\visits_{T+1}(\pair_\opt)]$.
    We conclude that $\E_{\state_0}^{\model, \learner}[\visits_{T+1}(\pair_\opt)] \ge \frac 1{2\epsilon} \log(T) + \oh(\log(T))$. 
    As this holds for $\epsilon > 0$ arbitrarily small, we conclude accordingly. 
\end{proof}

    \section{Background from information theory}\label{app:informationtheory}
    
In this appendix and for the sake of self-containedness, we provide a proof of \eqref{equation_change_of_measure} and \eqref{equation_change_of_measure_log} that are both key to derive the regret lower bound of \cref{theorem_lower_bound}.
The curious reader may also read \cite{maillard2019mathematics,kaufmann2014analyse,kaufmann_complexity_2016}.

\subsection{Change of measures and log-likelihood ratios}

We begin with an instance of the celebrated Radon-Nikodym' theorem, specialized for the stochastic process underlying a Markov decision process. 

\begin{theorem}[Change of measure]
    \label{theorem_change_measure}
    Let $\model \ll \model'$ two Markov decision processes and fix an arbitrary learning agent $\learner$ and $\state \in \states$. 
    If $U \ge 0$ is a $\sigma(\History_T)$-measurable random variable, then:
    \begin{equation*}
        \E^{\model', \learner}_\state
        \brackets*{
            U
        }
        \ge 
        \E^{\model, \learner}_\state
        \brackets*{
            U 
            \exp\braces*{-L(\History_t)}
        }.
    \end{equation*}
\end{theorem}
\begin{proof}
    Let $\nu_T, \nu'_T$ the distributions of $\History_T$ induced by running $\learner$ for $T$ steps from $\state$ on $\model, \model'$ respectively.
    We have $\nu_T \ll \nu'_T$, so by Radon-Nikodym's theorem, for every $\sigma(\History_T)$-measurable random variable $V$, we have:
    \begin{equation*}
        \E_{\state}^{\model, \learner}[V]
        \equiv
        \int V(\history_T) ~ \dd \nu_T(\history_T)
        =
        \int V(\history_T) ~ \frac{\dd \nu_T}{\dd \nu'_T}(\history_T) ~ \dd \nu'_T(\history_T)
        \equiv
        \E_{\state}^{\model', \learner}\brackets*{
            V \cdot \frac{\dd\nu_T}{\dd\nu'_T}(\History_T)
        }
    \end{equation*}
    where $\frac{\dd\nu_T}{\dd\nu'_T}$ is the Radon Nikodym of $\nu_T$ with respect to $\nu'_T$. 
    By Markov's property, the Radon-Nikodym derivative can be written as:
    \begin{equation*}
    \begin{aligned}
        \frac{\dd\nu_T}{\dd\nu'_T}(\History_T)
        & =
        \frac{
            \prod_{t=1}^{T-1} \kerneld(\State_{t+1}|\State_t, \Action_t) ~ \rewardd(\Reward_t|\State_t,\Action_t) ~ \learner(\Action_t|\State_1, \Action_1, \Reward_1, \ldots, \State_t)
        }{
            \prod_{t=1}^{T-1} \kerneld'(\State_{t+1}|\State_t, \Action_t) ~\rewardd'(\Reward_t|\State_t,\Action_t) ~ \learner(\Action_t|\State_1, \Action_1, \Reward_1, \ldots, \State_t)
        }
        \\
        & =
        \prod_{t=1}^{T-1} 
        \frac{\kerneld(\State_{t+1}|\State_t,\Action_t)}{\kerneld'(\State_{t+1}|\State_t,\Action_t)}
        \frac{\rewardd(\State_{t+1}|\State_t,\Action_t)}{\rewardd'(\State_{t+1}|\State_t,\Action_t)}
        \\
        & =
        \exp\braces*{
            \sum_{t=1}^{T-1}
            \parens*{
                \log \parens*{
                    \frac{\kerneld(\State_{t+1}|\State_t,\Action_t)}{\kerneld'(\State_{t+1}|\State_t,\Action_t)}
                }
                +
                \log \parens*{
                    \frac{\rewardd(\State_{t+1}|\State_t,\Action_t)}{\rewardd'(\State_{t+1}|\State_t,\Action_t)}
                }
            }
        }
        \\
        & =: \exp\braces{L(\History_T)}
    \end{aligned}
    \end{equation*}
    with the convention that $\log(0) = - \infty$. 
    Set $V := U \indicator{L(\History_T) > - \infty} \exp\set{-L(\History_T)}$, with the convention $0 \cdot \infty = 0$. 
    We have:
    \begin{equation*}
    \begin{aligned}
        \E_{\state}^{\model', \learner}[U]
        & \overset{(\dagger)}\ge
        \E_{\state}^{\model', \learner}[V \cdot \exp\set{L(\History_T)}]
        \\
        & \overset{(\ddagger)}=
        \E_{\state}^{\model, \learner}[
            U \cdot \indicator{L(\History_T) > - \infty} \exp\braces{- L (\History_T)}
        ]
        \overset{(\S)}=
        \E_{\state}^{\model, \learner}[U]
    \end{aligned}
    \end{equation*}
    where $(\dagger)$ uses that $U \ge V$, $(\ddagger)$ invokes Radon-Nikodym's theorem and $(\S)$ uses that, under $\model, \learner$, we have $L(\History_T) > - \infty$ almost surely. 
\end{proof}

\begin{lemma}
    \label{lemma_log_likelihood}
    The log-likelihood ratio has expected value:
    \begin{equation*}
        \E_{\state}^{\model, \learner}\brackets*{
            \sum_{t=1}^{T-1}
            \parens*{
                \log \parens*{
                    \frac{\kerneld(\State_{t+1}|\State_t,\Action_t)}{\kerneld'(\State_{t+1}|\State_t,\Action_t)}
                }
                +
                \log \parens*{
                    \frac{\rewardd(\State_{t+1}|\State_t,\Action_t)}{\rewardd'(\State_{t+1}|\State_t,\Action_t)}
                }
            }
        }
        =
        \E_{\state}^{\model, \learner} \brackets*{
            \sum_{\pair \in \pairs}
            \visits_T(\pair)
            \KL_\pair(\model||\model')
        }
        .
    \end{equation*}
\end{lemma}
\begin{proof}
    Direct application of the tower property.
\end{proof}

\subsection{Simpler forms of the change of measure's inequality}

While the inequality of \cref{theorem_change_measure} can be used as is, we provide below a few alternative (weaker) form that are often more easy to work with. 

\begin{corollary}
\label{corollary_likelihood_jensen}
    Let $\model \ll \model'$ two Markov decision processes and fix an arbitrary learning agent $\learner$ and $\state \in \states$. 
    If $U \ge 0$ is a $\sigma(\History_T)$-measurable random variable, then:
    \begin{equation*}
        \log \E_{\state}^{\model', \learner}[U]
        + 
        \E_{\state}^{\model, \learner} \brackets*{
            \sum_{\pair \in \pairs}
            \visits_T(\pair)
            \KL_\pair(\model||\model')
        }
        \ge
        \E_{\state}^{\model, \learner} [\log(U)].
    \end{equation*}
\end{corollary}
\begin{proof}
    We write:
    \begin{align*}
        \E_{\state}^{\model', \learner}[U]
        & \overset{(\dagger)}\ge 
        \E_{\state}^{\model, \learner}\brackets*{
            U
            \exp\braces*{- L (\History_T)}
        }
        \\
        & =
        \E_{\state}^{\model, \learner}\brackets*{
            \exp\braces*{
                - L (\History_T) + \log(U)
            }
        }
        \\
        & \overset{(\ddagger)}\ge
        \exp \braces*{
            \E_{\state}^{\model, \learner} \brackets*{
                - L (\History_T) + \log(U)
            }
        }
        \\
        & \overset{(\S)}=
        \exp \braces*{
            - \E_{\state}^{\model, \learner} \brackets*{
                \sum_{\pair \in \pairs}
                \visits_T(\pair) \KL_\pair(\model||\model')
            }
            + 
            \E_{\state}^{\model, \learner} [\log(U)]
        }
    \end{align*}
    where $(\dagger)$ follows by \cref{theorem_change_measure}, $(\ddagger)$ is a use of Jensen's inequality and $(\S)$ follows by \cref{lemma_log_likelihood}.
    Conclude by taking the log and rearranging terms.
\end{proof}

\begin{corollary}
\label{corollary_likelihood_event}
    Let $\model \ll \model'$ two Markov decision processes and fix an arbitrary learning agent $\learner$ and $\state \in \states$. 
    If $\event$ is a $\sigma(\History_T)$-measurable event, then:
    \begin{equation*}
        \E^{\model, \learner}_{\state}\brackets*{
            \sum_{\pair \in \pairs}
            \visits_T(\pair)
            \KL_\pair(\model||\model')
        }
        \ge
        \kl\parens*{\Pr_{\state}^{\model, \learner}(\event), \Pr_{\state}^{\model', \learner}(\event)}
        .
    \end{equation*}
\end{corollary}
\begin{proof}
    For conciseness, we write $\E[-], \E'[-]$ rather $\E_{\state}^{\model, \learner}[-], \E_{\state}^{\model', \learner}[-]$, and $\rho := \Pr(\event)$ and $\rho' := \Pr'(\event)$.

    If $\rho = \rho'$, there is nothing to prove.
    Up to considering $\event^\complement$, we can assume that $\rho > \rho'$. 
    We have:
    \begin{equation*}
    \begin{aligned}
        \E \brackets*{
            \sum_{\pair \in \pairs}
            \visits_T(\pair)
            \KL_\pair(\model||\model')
        }
        & \ge 
        \sup_{\lambda \ge 0} \braces[\bigg]{
            \E\brackets*{\log(\lambda + \indicator{\event})}
            + \log \E'\parens*{\lambda + \indicator{\event}}
        }
        \\
        & =
        \sup_{\lambda > 0} \braces[\bigg]{
            \rho \log(\lambda + 1) + (1 - \rho) \log(\lambda) - \log(\lambda + \rho')
        }
        =: \sup_{\lambda > 0} \varphi(\lambda)
    \end{aligned}
    \end{equation*}
    The supremum of $\varphi$ is reached for $\lambda := \frac{\rho'(1-\rho)}{\rho - \rho'}$, and with a bit of algebra, we find that $\sup_{\lambda > 0} \varphi(\lambda) = \kl(\rho, \rho')$. 
    This concludes the proof.
\end{proof}

    \section{Proofs of complexity results}\label{appendix_complexity}
    
In this appendix, we provide the complete proofs of \cref{theorem_critical,theorem_conp_hard}.

\subsection{Proof of \cref{theorem_critical}}

    Proving that it is NP is immediate, because the optimal gain of a MDP is the solution of a linear program \cite{puterman2014markov} hence given $\model^\dagger$, checking that $\sum_\pair \imeasure(\pair) \KL_\pair(\model||\model^\dagger) < \alpha$ and $\optgain(\model^\dagger) > \beta$ is done in polynomial time.
    The point is to show the NP-hardness.

    \STEP{1}
    To prove that the problem is NP-hard, it is reduced from the Knapsack Problem (\texttt{KP}). 
    Recall that an instance of \texttt{KP} is given by a collection of $n$ items of integer values $\set{v_1, \ldots, v_n}$ and integer weights $\set{w_1, \ldots w_n}$, as well as a capacity $W$ and a value threshold $V$, both integers.
    The problem is to determine whether there exists ${\mathcal{K}} \subseteq [n]$ such that $\sum_{k \in {\mathcal{K}}} w_k \le W$ and $\sum_{k \in {\mathcal{K}}} v_k \ge V$. 

    Fix $\epsilon, \sigma, \delta > 0$ to be tuned later on.
    Given an instance of \texttt{KP}, consider the MDP $\models$ whose structure is given by $n$ ({\sc choose $k$}) $3$-state widgets connected in a ring fashion. 

    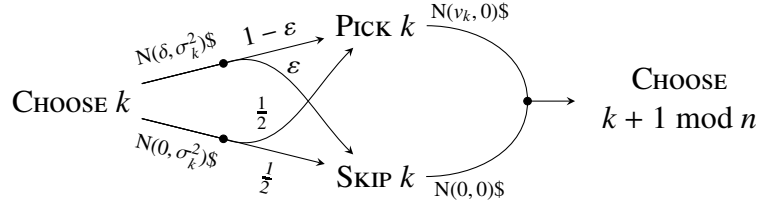
\begin{figure}[h]
        \centering
        \begin{tikzpicture}
            \node (0) at (0, 0) {\sc Choose $k$};
            \node (1) at (4, 1) {\sc Pick $k$};
            \node (2) at (4, -1) {\sc Skip $k$};
            \node (3) at (8, 0) {\begin{tabular}{c}\sc Choose\\ $k + 1 \textrm{ mod } n$\end{tabular}};

                \draw[fill=black] (2, .5) circle(.5mm);
                \draw (0) to node[midway, above, sloped] {\scriptsize $\mathrm{N}(\delta, \sigma_k^2)\$$} (2, .5);
                \draw[->,>=stealth] (0) to (2, .5) to node[midway, above, sloped] {\footnotesize $1-\epsilon$} (1);
                \draw[->,>=stealth] (0) to (2, .5) to[out=20,in=180-45] node[pos=.5, above] {\footnotesize $\epsilon$} (2);

                \draw[fill=black] (2,-.5) circle(.5mm);
                \draw (0) to node[midway, below, sloped] {\scriptsize $\mathrm{N}(0, \sigma_k^2)\$$} (2,-.5);
                \draw[->,>=stealth] (0) to (2,-.5) to node[midway, below, sloped] {\footnotesize $\frac 12$} (2);
                \draw[->,>=stealth] (0) to (2,-.5) to[out=-20,in=180+45] node[pos=.25, above] {\footnotesize $\frac 12$} (1);

                \draw[fill=black] (6, 0) circle(.5mm);
                \draw (1) to[in=90,out=0] node[pos=.3, above] {\scriptsize $\mathrm{N}(v_k, 0)\$$} (6, 0); 
                \draw (2) to[in=-90,out=0] node[pos=.3, below] {\scriptsize $\mathrm{N}(0, 0)\$$} (6, 0); 
                \draw[->,>=stealth] (6, 0) to (3);
        \end{tikzpicture}
        \caption{
            \label{figure_widget}
            The ({\sc Choose $k$}) widget, where
            $\sigma_k^2 := \frac{\sigma^2}{w_k}$.
        }
    \end{figure}

    From the state ({\sc Choose $k$}) are two actions: The top action that is likely to go to ({\sc Pick $k$}) that shall be referred to as action {\sc Pick}, and the bottom action called {\sc Skip}. 
    From every over state, there is a single action that denoted $*$. 
    A (deterministic) policy of $\models$ is analogue to a subset ${\mathcal{K}} \subseteq \set{1, ..., n}$, written $\policy_{\mathcal{K}}$, which is given by $\policy_{{\mathcal{K}}}(\textsc{Pick}|\textsc{Choose}~k) := \indicator{k \in {\mathcal{K}}}$. 
    We get:
    \begin{equation}
        \gain(\policy_{\mathcal{K}}) 
        = \frac 1{2n} \parens*{
            \frac 12 \sum_{k=1}^n v_k
            +
            \sum_{k \in {\mathcal{K}}} \parens*{\parens*{\tfrac 12-\epsilon} v_k + \delta}
        }
        =
        \frac{\norm{v}_1}{4n}
        + 
        \frac 1{2n}
        \sum_{k \in {\mathcal{K}}} \parens*{\parens*{\tfrac 12-\epsilon} v_k + \delta}.
    \end{equation}

    \STEP{2}
    Every policy $\policy_{{\mathcal{K}}}$ can equivalently be seen as a \emph{single-action} Markov decision process $\model_{{\mathcal{K}}}$, i.e., the model of a policy over the state-space 
    \begin{equation}
    \nonumber
        \states :=
        \set{(\textsc{Choose }k), (\textsc{Pick }k), (\textsc{Skip }k) : k=1, \ldots, n}.
    \end{equation}
    The choice of an action is equivalently the choice of a kernel distribution. 
    The set of stationary deterministic policies of $\models$, denoted $\policies^\text{SD}(\models)$, can therefore be seen as the set of Markov reward processes $\models^{\text{SD}} := \set{\model_{\mathcal{K}} : {\mathcal{K}} \subseteq \set{1, \ldots, n}}$. 
    Provided that the parameters $\epsilon, \sigma, \delta$ are polynomial in $n, v, w$, this (structured) set of mdps is described in polynomial size in $n, v, w$.

    Consider $\model_\varnothing \in \models^\text{SD}$. 
    Because $\models^\text{SD}$ is a space of single-action MDPs, we don't make any distinction between a state and a pair of $\model_{\mathcal{K}} \in \models^\text{SD}$. 
    Now, we see that $\gain(\model_\varnothing) = \frac{\norm{v}_1}{4n}$ and the invariant measure $\imeasure$ of the unique policy of $\model_\varnothing$ is:
    \begin{equation}
    \nonumber
        \imeasure(\textsc{Choose $k$}) = \tfrac 1{2n}
        \text{~and~}
        \imeasure(\textsc{Pick $k$}) = \imeasure(\textsc{Skip $k$})= \tfrac 1{4n}.
    \end{equation}
    Moreover, check that for ${\mathcal{K}} \subseteq \set{1, \ldots, n}$, the only states such that $\KL_\state({\model_\varnothing\|\model_{\mathcal{K}}}) \ne 0$ are (\textsc{Choose $k$}) states, with:
    \begin{equation}
    \nonumber
        \KL_{(\textsc{Choose $k$})}({\model_\varnothing\|\model_{\mathcal{K}}})
        =
        \indicator{k \in {\mathcal{K}}} \parens*{
            \log\parens*{\tfrac 1{4\epsilon(1-\epsilon)}}
            +
            w_k \parens*{\tfrac \delta\sigma}^2
        }.
    \end{equation}
    Hence:
    \begin{equation}
    \nonumber
        \sum_\pair \imeasure(\pair) 
        \KL_\pair({\model_\varnothing\|\model_{\mathcal{K}}})
        =
        \frac 1{2n}
        \sum_{k \in {\mathcal{K}}}
        \parens*{
            \log\parens*{\tfrac 1{4\epsilon(1-\epsilon)}}
            +
            w_k \parens*{\tfrac \delta\sigma}^2
        }.
    \end{equation}

    \STEP{3}
    We want (1) to be able to retrieve the value of $\sum_{k \in {\mathcal{K}}} v_k$ from $\gain(\model_{\mathcal{K}})$; (2) to be able to retrieve the value of $\sum_{k \in {\mathcal{K}}} w_k$ from $\sum_\pair \imeasure(\pair) \KL_{\pair}({\model_\varnothing\|\model_k})$. 
    For simplicity and because it will eventually work with it, fix $\epsilon \equiv \frac 14$. 

    The condition (1) holds when $\delta = \frac 1{16n}$. 
    Indeed, then:
    \begin{align*}
        \gain(\model_{\mathcal{K}}) 
        & = 
        \frac{\norm{v}_1}{4n}
        + \frac 1{8n} \sum_{k \in {\mathcal{K}}}\parens*{v_k + 4\delta}
        =
        \frac 1{8n} \parens*{
            2\norm{v}_1 
            + \sum_{k \in {\mathcal{K}}} v_k 
            \pm \frac 14
        }
    \end{align*}
    where $\pm \frac 14$ denotes an arbitrary quantity in the range of $[-\frac 14, \frac 14]$.
    Rearranging, we get $\sum_{k\in {\mathcal{K}}} v_k = 8n \gain(\model_{\mathcal{K}}) - 2\norm{v}_1 \pm \frac 14 = [8n\gain(\model_{\mathcal{K}}) - 2\norm{v}_1]$ where $[\lambda]$ denotes the rounding operation (nearest integer). 

    The condition (2) is satisfied when 
    \begin{equation}
    \nonumber
        \sigma^2 
        = \frac{\delta^2}{4n\log\parens*{\frac 1{4\epsilon(1-\epsilon)}}} 
        \equiv \frac{1}{1024n^3\log\parens*{\frac 43}}.
    \end{equation}
    Indeed, then we have
    \begin{align*}
        \sum_\pair \imeasure(\pair)
        \KL_\pair({\model_\varnothing\|\model_{\mathcal{K}}})
        & = 
        \frac 1{2n} \sum_{k \in {\mathcal{K}}}
        \parens*{
            \log\parens*{\tfrac 1{4\epsilon(1-\epsilon)}}
            +
            w_k \parens*{\tfrac \delta\sigma}^2
        }
        \\
        & = 
        \frac {\delta^2}{2n\sigma^2}
        \sum_{k \in {\mathcal{K}}} \parens*{
            w_k + \frac 1{4n}
        } 
        =
        \frac{\delta^2}{2n \sigma^2} \parens*{
            \sum_{k \in {\mathcal{K}}} w_k
            \pm \frac 14
        }.
    \end{align*}
    Rearranging, we find $\sum_{k \in {\mathcal{K}}} w_k = \round*{\frac{2n\sigma^2}{\delta^2} \sum_\pair \imeasure(\pair) \KL_\pair({\model_\varnothing\|\model_{\mathcal{K}}})}$ where $\round{\lambda}$ also denotes the rounding operation. 

    \STEP{4}
    Remark that this choice of $\epsilon, \sigma, \delta$ is polynomial in the size of $n$. 
    Following this remark, it should be clear that $\models^\text{SD}$ can be encoded in polynomial size.
    Finally set $\alpha = 2 \log(4/3)(W + \frac 13)$ and $\beta = \frac 1{8n}(2 \norm{v} + V)$. 
    Then, we claim that there is $\model^\dagger \in \models^\text{SD}$ such that 
    \begin{equation}
    \label{equation_criticalmodel_1}
        \sum_\pair \imeasure(\pair) \KL_\pair(\model||\model^\dagger) \le \alpha ,
        \text{~and~}
        \optgain(\model^\dagger) \ge \beta
    \end{equation}
    if, and only if the \texttt{KP} instance $(v, w, V, W)$ has a solution. 

    This is just a commodity to check using the formulas established so far. 
    If the \texttt{KP} instance has solution ${\mathcal{K}}$, then $\model_{\mathcal{K}}$ is by construction a solution of \eqref{equation_criticalmodel_1}, because 
    \begin{equation}
    \nonumber
        \sum_\pair \imeasure(\pair) \KL_\pair(\model||\model^\dagger) 
        =
        \frac{\delta^2}{2n\sigma^2} \sum_{k \in {\mathcal{K}}} \parens*{w_k + \frac 1{4n}}
        \le
        2 \log\parens*{\tfrac 43} \parens*{W + \tfrac{\abs{{\mathcal{K}}}}{4n}} 
        < \alpha
    \end{equation}
    and 
    \begin{equation}
    \nonumber
        \gain(\model_{\mathcal{K}})
        = 
        \frac 1{8n} \parens*{
            2\norm{v}_1 
            + \sum_{k \in {\mathcal{K}}} \parens*{v_k + \tfrac 1{4n}}
        }
        >
        \frac 1{8n} \parens*{
            2\norm{v}_1 
            + V
        } = \beta.
    \end{equation}

    Conversely, if $\model_k$ is a solution of \eqref{equation_criticalmodel_1}, then we have
    \begin{equation}
    \nonumber
        \alpha = 2\log\parens*{\tfrac 43}\parens*{W + \tfrac 13}
        \ge 
        \frac{\delta^2}{2n\sigma^2} \sum_{k \in {\mathcal{K}}} \parens*{w_k + \frac 1{4n}}
        \ge 
        2\log\parens*{\tfrac 43}\sum_{k \in {\mathcal{K}}} w_k
    \end{equation}
    hence $\sum_{k\in {\mathcal{K}}} w_k \le W + \tfrac 13$, so $\sum_{k \in {\mathcal{K}}} w_k \le W$; and similarly
    \begin{equation}
    \nonumber
        \beta = \frac 1{8n}\parens*{2\norm{v} + V}
        \le 
        \frac{1}{8n}\parens*{
            2\norm{v} + \frac14 + \sum_{k \in {\mathcal{K}}} v_k
        }
    \end{equation}
    so $\sum_{k \in {\mathcal{K}}} v_k \ge V - \frac 14$, so $\sum_{k \in {\mathcal{K}}} v_k \ge V$. 
    \hfill\qed

\subsection{Proof of \cref{theorem_conp_hard}}

    We provide a reduction from the co-knapsack problem (co-\texttt{KP}), which is coNP-complete because \texttt{KP} is NP-complete.
    An instance of co-\texttt{KP} is given by a collection of $n$ items of integer values $\set{v_1, \ldots, v_n}$ and integer weights $\set{w_1,\ldots, w_n}$, as well as a capacity and a value threshold $V$, both integers.
    The problem is to determine if, for all ${\mathcal{K}} \subseteq [n]$, we either have $\sum_{k \in {\mathcal{K}}} w_k \ge W$ or $\sum_{k \in {\mathcal{K}}} v_k \le V$.

\begin{figure}[h]
        \centering
        \resizebox{\linewidth}{!}{
        \begin{tikzpicture}
            \draw (-2, 0) arc (180:180+360:2);
            \node[fill=white, draw, circle] (0) at (180:2) {$0$};
            \node[fill=white, draw, dashed] (1) at (140:2) {$1$};
            \node[fill=white, draw, dashed] (2) at (100:2) {$2$};
            \node[fill=white, draw, dashed] (n) at (220:2) {$n$};
            \node[fill=white, draw, dashed] (k) at (0:2) {$k$};

            \draw[->, >=stealth, dashed] (k) to (3.6, 0);
            \draw[->, >=stealth] (0) to[in=180+45,out=180-45,looseness=3] node[left,pos=0.5] {$\mathrm{N}(\theta, 0)$} (0);
            \node at (160:2.75) {$\mathrm{N}(0, 0)$};

            \begin{scope}[shift={(5, 0)}]
                \draw[dashed, rounded corners] (-1.4, -2) rectangle (9.5, 2);
                \node (0) at (0, 0) {\sc Choose $k$};
                \node (1) at (4, 1) {\sc Pick $k$};
                \node (2) at (4, -1) {\sc Skip $k$};
                \node (3) at (8, 0) {\begin{tabular}{c}\sc Choose\\ $k + 1 \mod n$\end{tabular}};

                    \draw[fill=black] (2, .5) circle(.5mm);
                    \draw (0) to node[midway, above, sloped] {\scriptsize $\mathrm{N}(\delta, \sigma_k^2)$} (2, .5);
                    \draw[->,>=stealth] (0) to (2, .5) to node[midway, above, sloped] {\footnotesize $1-\epsilon$} (1);
                    \draw[->,>=stealth] (0) to (2, .5) to[out=20,in=180-45] node[pos=.5, above] {\footnotesize $\epsilon$} (2);

                    \draw[fill=black] (2,-.5) circle(.5mm);
                    \draw (0) to node[midway, below, sloped] {\scriptsize $\mathrm{N}(0, \sigma_k^2)$} (2,-.5);
                    \draw[->,>=stealth] (0) to (2,-.5) to node[midway, below, sloped] {\footnotesize $\frac 12$} (2);
                    \draw[->,>=stealth] (0) to (2,-.5) to[out=-20,in=180+45] node[pos=.25, above] {\footnotesize $\frac 12$} (1);

                    \draw[fill=black] (6, 0) circle(.5mm);
                    \draw (1) to[in=90,out=0] node[pos=.3, above] {\scriptsize $\mathrm{N}(v_k, 0)$} (6, 0); 
                    \draw (2) to[in=-90,out=0] node[pos=.3, below] {\scriptsize $\mathrm{N}(0, 0)$} (6, 0); 
                    \draw[->,>=stealth] (6, 0) to (3);
            \end{scope}
        \end{tikzpicture}}
        \caption{
        \label{figure_knapsack_widget}
            Embedding a knapsack problem in a Markov decision process.
        }
    \end{figure}
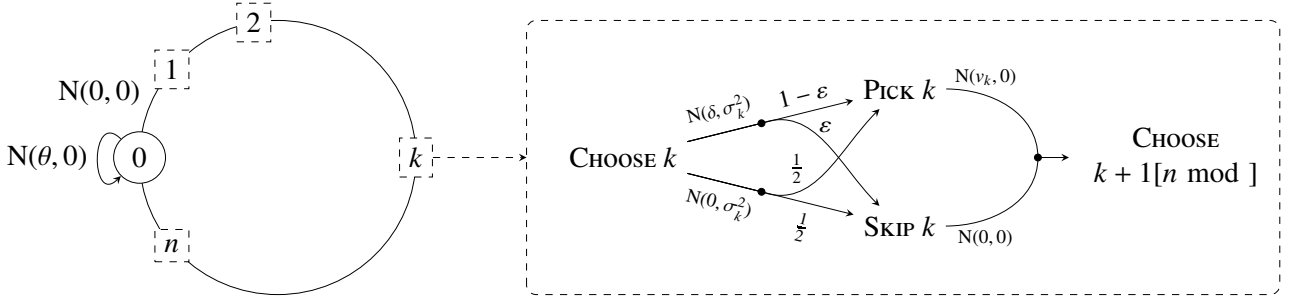

    The reduction is very similar to \texttt{CONFUSING-MODEL}'s. 
    Fix $\epsilon, \sigma, \delta, \theta$ to be tuned later on and consider an instance of co-\texttt{KP}.
    Consider the MDP $\models$ whose structure is as given by \cref{figure_knapsack_widget}.
    
    The change regarding the reduction of \texttt{CONFUSING-MODEL} is the state $(0)$, in between ({\sc Choose $n$}) and ({\sc Choose $1$}).
    From $(0)$ you can either loop with the action \textsc{Loop} scoring $\theta$, or go to ({\sc Choose $1$}) with the action {\sc Cycle} scoring $0$, hence entering the big cycle. 
    The state $(0)$ is a special state.
    From the state ({\sc Choose $k$}) are two actions: The top action that is likely to go to ({\sc Pick $k$}) that we shall refer to as action {\sc Pick}, and the bottom action that we shall call {\sc Skip}. 
    From every over state, there is a single action that denoted $*$. 
    The special policy looping on $(0)$ is denoted $\policy^\opt$ and will model the optimal policy later on. 
    The other (deterministic) policies of $\models$ are analogue to a subset ${\mathcal{K}} \subseteq \set{1, ..., n}$, written $\policy_{\mathcal{K}}$, and are given $\policy_{{\mathcal{K}}}(\textsc{Pick}|\textsc{Choose}~k) := \indicator{k \in {\mathcal{K}}}$ with $\policy(\textsc{Cycle}|0) = 1$. 
    We get:
    \begin{equation}
        \gain(\policy_{\mathcal{K}}) 
        = \frac 1{2(n+1)} \parens*{
            \frac 12 \sum_{k=1}^n v_k
            +
            \sum_{k \in {\mathcal{K}}} \parens*{\parens*{\tfrac 12-\epsilon} v_k + \delta}
        }
        =
        \frac{\norm{v}_1}{4(n+1)}
        + 
        \frac 1{2(n+1)}
        \sum_{k \in {\mathcal{K}}} \parens*{\parens*{\tfrac 12-\epsilon} v_k + \delta}.
    \end{equation}

    Every policy $\policy_{\mathcal{K}}$ can equivalently be seen as a single-action Markov decision process $\model_{\mathcal{K}}$. 
    The choice of an action is equivalently the choice of a kernel distribution. 
    The set of stationary deterministic policies of $\models$, denoted $\policies^\text{SD}(\models)$, can therefore be seen as the set of Markov reward processes $\models^\text{SD} := \set{\model_{\mathcal{K}} : {\mathcal{K}}\subseteq \set{1, \ldots, n}}$.
    Now, we see that $\gain(\model_\varnothing) = \frac{\norm{v}_1}{4(n+1)}$ and the invariant measure $\imeasure_\varnothing$ of the unique policy of $\model_\varnothing$ is:
    \begin{equation}
    \nonumber
        \imeasure_\varnothing(0) = \tfrac 1{n+1},~
        \imeasure_\varnothing(\textsc{Choose $k$}) = \tfrac 1{2(n+1)},
        \text{~and~}
        \imeasure_\varnothing(\textsc{Pick $k$}) = \imeasure_\varnothing(\textsc{Skip $k$})= \tfrac 1{4(n+1)}.
    \end{equation}
    Moreover, check that:
    \begin{equation}
    \nonumber
        \sum_\pair \imeasure_\varnothing(\pair) \KL_\pair({\model_\varnothing\|\model_{\mathcal{K}}})
        =
        \frac 1{2(n+1)} 
        \sum_{k \in {\mathcal{K}}}
        \parens*{
            \log\parens*{\tfrac 1{4\epsilon(1-\epsilon)}} + w_k \parens*{\tfrac{\delta}{\sigma}}^2
        }.
    \end{equation}

    We find the values:
    \begin{equation}
    \nonumber
        \epsilon = \frac 14,~ 
        \delta = \frac 1{16n}, ~
        \sigma^2 = \frac{\delta^2}{4(n+1)\log\parens*{\frac 43}},~
        \theta = \frac {2\norm{v}_1 + V}{8(n+1)},
        \text{~and~}
        \rho = \frac{V}{16\log\parens*{\frac 43} W}.
    \end{equation}
    Consider $\models_*^\text{SD}$ the copy of $\model^\text{SD}$ with each element augmented with the action \textsc{Loop} at $0$, scoring $\mathrm{N}(\theta, 0)$ and pick the reference model $\model_\varnothing \in \models^\text{SD}$ (In abuse of notations, we write the elements of $\models^\text{SD}$ and $\models^\text{SD}_*$ similarly because the two sets are obviously isomorphic, so $\model_\varnothing$ contains the policies $\policy_\varnothing$ and $\policy_*$). 
    We show that the initial co-\texttt{KP} problem is reduced to the \texttt{REGRET} instance $(\models^\text{SD}_*, \model_\varnothing, \rho)$.

    First, remark that $\policy^\opt$ is the optimal policy of $\model_\varnothing$.
    Then, We show that given $\imeasure \in \imeasures(\model_\varnothing/\optpairs)$ such that $\sum_\pair \imeasure(\pair) \Delta^\opt(\pair; \model_\varnothing) \le \rho$, we have 
    \begin{itemize}
        \item[(1)]
            $\sum_{k \in {\mathcal{K}}} v_k \le V$ if, and only if $\gain(\model_{\mathcal{K}}) > \theta$, i.e., $\model_{\mathcal{K}} \in \confusing(\model_\varnothing; \models^\text{SD}_*)$; and
        \item[(2)]
            $\sum_{k \in {\mathcal{K}}} w_k \ge W$ if, and only if $\sum_\pair \imeasure(\pair) \KL_\pair({\model_\varnothing\|\model_{\mathcal{K}}}) \ge 1$. 
    \end{itemize}
    We start with (1). 
    If $\sum_{k \in {\mathcal{K}}} v_k \le V$, then 
    \begin{equation}
    \nonumber
        \gain(\policy_{\mathcal{K}})
        >
        \frac{\norm{v}_1}{4(n+1)}
        + \frac{V}{8(n+1)}
        = \theta. 
    \end{equation}
    Conversely, if $\gain(\policy_{\mathcal{K}}) > \theta$, then 
    \begin{align*}
        \frac{2\norm{v}_1 + V}{8(n+1)}
        & < 
        \frac 1{8(n+1)}
        \parens*{
            2\norm{v}_1 
            + 
            \sum_{k \in {\mathcal{K}}} \parens*{v_k + \frac 1{4n}}
        }
        \le
        \frac{2\norm{v}_1 + \frac 14 + \sum_{k \in {\mathcal{K}}} v_k}{8(n+1)},
    \end{align*}
    so $\sum_{k \in {\mathcal{K}}} v_k \ge V - \frac 14$, so $\sum_{k \in {\mathcal{K}}} v_k \ge V$. 

    For (2), first remark that the only positive Bellman-gap of $\model_\varnothing$ is at the state-action pair $(0, \textsc{Cycle})$ with $\Delta^\opt((0, \textsc{Cycle}); \model_\varnothing) = \frac V8$. 
    Moreover, every element of $\imeasures(\model_\varnothing/\optpairs)$ is of the form $c \imeasure_\varnothing$ where $c > 0$.
    So, having $\sum_\pair \imeasure(\pair) \Delta^\opt(\pair; \model_\varnothing) \le \rho$ means that $\imeasure = c \imeasure_\varnothing$ with $c \le \frac{8\rho}{V} = (2 \log(\frac 43) W)^{-1}$.
    With this in mind, if $\sum_{k \in {\mathcal{K}}} w_k \ge W$, then
    \begin{equation}
    \nonumber
        \sum_\pair \imeasure(\pair) \KL_\pair({\model_\varnothing\|\model_{\mathcal{K}}})
        = 
        \frac{c\delta^2}{2(n+1)\sigma^2}\sum_{k \in {\mathcal{K}}} \parens*{w_k + \frac 1{4n}}
        \ge 
        \frac{2\log\parens*{\frac 43} W}{2\log\parens*{\frac 43} W}
        \ge 1.
    \end{equation}
    Conversely, if $\sum_\pair \imeasure(\pair) \KL_\pair({\model_\varnothing\|\model_{\mathcal{K}}}) \ge 1$, then 
    \begin{equation}
    \nonumber
        1 \le 
        \frac{c\delta^2}{2(n+1)\sigma^2}\sum_{k \in {\mathcal{K}}} \parens*{w_k + \frac 1{4n}}
        \le 
        \frac{2\log\parens*{\frac 43}\parens*{\sum_{k \in {\mathcal{K}}} w_k + \frac{|{\mathcal{K}}|}{4n}}}{2\log\parens*{\frac 43} W}
    \end{equation}
    so $\sum_{k \in {\mathcal{K}}} w_k \ge W - \frac 14$, so $\sum_{k \in {\mathcal{K}}} w_k \ge W$.

    We readily obtain that: ``every ${\mathcal{K}} \subseteq [n]$ satisfies either $\sum_{k \in {\mathcal{K}}} w_k \ge W$ or $\sum_{k \in {\mathcal{K}}} v_k \le V$'' is equivalent to $\imeasure \equiv (2 W \log(\frac 43))^{-1} \imeasure_\varnothing$ satisfying:
    \begin{equation}
    \nonumber
        \forall \model^\dagger \in \confusing(\model_\varnothing; \models^\text{SD}_*),
        ~
        \sum_\pair \imeasure(\pair) \KL_\pair({\model_\varnothing\|\model_{\mathcal{K}}}) \ge 1,
    \end{equation}
    and this $\imeasure$ is the unique $\imeasure \in \imeasures(\model_\varnothing/\optpairs)$ such that $\sum_\pair \imeasure(\pair) \Delta^\opt(\pair; \model_\varnothing) = \rho$. 
    \hfill\qed

\end{document}